%% file: main.tex
 \documentclass[10pt, a4paper, twocolumn, showabstract]{naverlabseurope} % no full width teaser figure

\usepackage{lipsum}
\usepackage{multicol}
\usepackage{tikz,tkz-kiviat}

\input{macros}

\usepackage{multirow}
\usepackage{bm}
\usepackage{graphicx}
\usepackage{subfig}
\usepackage{amsmath}  % for \mathds
\usepackage{dsfont}   % for \mathds

\title{What could go wrong? Discovering and describing failure modes in computer vision}
\correspondingauthor{Gabriela.Csurka@naverlabs.com}

\authors{Gabriela Csurka \authsep Tyler~L.~Hayes \authsep Diane Larlus \authsep Riccardo Volpi}
\affiliations{NAVER LABS Europe}
\contributions{}
\website{https://europe.naverlabs.com}
\websiteref{\href{https://europe.naverlabs.com}}
\teaserfile{}
\teasercaption{}
\date{}

\begin{abstract}

Deep learning models are effective, yet brittle. Even if carefully trained, their behavior tends to be hard to predict when confronted with out-of-distribution samples. In this work, our goal is to propose a simple yet  effective solution to \emph{predict} and \emph{describe}  via natural language potential  failure modes of computer vision  models.  Given a pretrained model and a set of samples, our aim is to find sentences that accurately describe the visual conditions in which the model underperforms. In order to study this important topic and foster future research on it, we
formalize the problem of Language-Based Error Explainability (\task{}) and propose a set of metrics to evaluate and compare different  methods for this task. We propose solutions that operate in a joint vision-and-language embedding space, and can characterize through language descriptions model failures caused, \eg, by  objects unseen during training  or adverse visual conditions.   We experiment with different tasks, such as classification under the presence of dataset bias
and semantic segmentation in unseen environments, and show that the proposed methodology isolates nontrivial sentences associated with specific error causes. We hope our work will help practitioners better understand the behavior of models, increasing their overall safety and interpretability. 

\end{abstract}

\begin{document}

\maketitle

% %%%%%%%%%%%%%%%%%%%%%%%%%%%%%%%%%%%%%%%%%%%%%%%%%%%%%%%%%%%%%%%%%
\input{1_intro}

\input{2_related_work}

\input{3_prob_form}
\input{4_method}

\input{5_metrics}

\input{6_results}
\input{7_conclusion}
% %%%%%%%%%%%%%%%%%%%%%%%%%%%%%%%%%%%%%%%%%%%%%%%%%%%%%%%%%%%%%%%%%

{
    \small
    \bibliographystyle{ieeenat_fullname}
 %   \bibliography{lbe}

\input{main.bbl}
}

%\clearpage
\appendix
\vspace{3cm}
\textbf{\Large Appendix }

\input{8_Appendix}

\end{document}

%% file: macros.tex
\newlength{\smallimage}
        \definecolor{rel}{rgb}{.1,.6,.2}
        \definecolor{nrl}{rgb}{1,1,1}
        \definecolor{qim}{rgb}{1,1,1}

\def\egs{\emph{e.g.\,}}
\def\eg{\emph{e.g.}}

\def\ies{\emph{i.e.\,}}
\def\ie{\emph{i.e.}}

\def\etc{\emph{etc.\,}}
\def\vs{\emph{versus\,\,}}

\def\etal{\emph{et al.\,}}

\definecolor{lightgray}{gray}{0.93}
\newcommand{\step}[1]{\colorbox{lightgray}{#1}}
\def\be{\begin{equation}}
\def\ee{\end{equation}}
\def\bea{\begin{eqnarray}}
\def\eea{\end{eqnarray}}
\def\ben{\begin{eqnarray*}}
\def\een{\end{eqnarray*}}

\def\bi{\begin{itemize}}
\def\ei{\end{itemize}}

\newcommand{\btab}[1]{\begin{tabular}{#1}}
\newcommand{\etab}{\end{tabular}}
\newcommand{\ba}[1]{\begin{array}{#1}}
\newcommand{\ea}{\end{array}}

\def\<{\langle}
\def\>{\rangle}

\newcommand{\bfc}{{\bf c}}
\newcommand{\bfd}{{\bf d}}

\newcommand{\bfs}{{\bf s}}

\newcommand{\bfv}{{\bf v}}

\newcommand{\bfx}{{\bf x}}

\newcommand{\bfz}{{\bf z}}

\newcommand{\calC}{{\mathcal C}}

\newcommand{\calE}{{\mathcal E}}

\newcommand{\calM}{{\mathcal M}}

\newcommand{\calR}{{\mathcal R}}
\newcommand{\calS}{{\mathcal S}}

\newcommand{\calX}{{\mathcal X}}

\newcommand{\myparagraph}[1]{\vspace{0.1cm}\noindent\textbf{#1.}}

\makeatletter\newcommand*{\at}{@}\makeatother

\newcommand{\nth}{$n^\textrm{th}$\,}

\definecolor{DarkCoral}{rgb}{0.8, 0.36, 0.27}

\newcommand{\redx}{{\color{red}$\times$}}

\newcommand{\rplus}{{\color{red}\textbf{+}}}
\definecolor{DarkGreen}{rgb}{0.5, 0.9, 0.5}
\newcommand{\gcheckmark}{{\color{ForestGreen}\checkmark}}
\newcommand{\gplus}{{\color{ForestGreen}\textbf{+}}}

\newcommand{\biased}{{\textit{Biased}}}
\newcommand{\unbiased}{{\textit{Unbiased}}}

\newcommand{\npp}{{\textbf{NICO${++}$\,}}}
\newcommand{\nppl}{{\textbf{NICO$_{++}^{75}$\,}}}
\newcommand{\nppm}{{\textbf{NICO$_{++}^{85}$\,}}}
\newcommand{\npph}{{\textbf{NICO$_{++}^{95}$\,}}}

\renewcommand{\npp}{{NICO${++}$\,}}
\renewcommand{\nppl}{{NICO$_{++}^{75}$\,}}
\renewcommand{\nppm}{{NICO$_{++}^{85}$\,}}
\renewcommand{\npph}{{NICO$_{++}^{95}$\,}}

\newcommand{\fpdiff}{{\textbf{FPDiff}\,}}
\newcommand{\pdiff}{{\textbf{PDiff}}\,}
\newcommand{\setdiff}{{\textbf{SetDiff}\,}}
\newcommand{\tops}{{\textbf{TopS}\,}}

\newcommand{\patk}{{\textbf{P\at K}\,}}
\newcommand{\task}{{LBEE}}

\newcommand{\jointSpace}{$\boldsymbol{\mathcal{F}}$\,}

\usepackage[capitalize]{cleveref}
\creflabelformat{table}{#2\textcolor{red}{#1}#3}
\creflabelformat{figure}{#2\textcolor{red}{#1}#3}
\creflabelformat{section}{#2\textcolor{red}{#1}#3}
\creflabelformat{appendix}{#2\textcolor{red}{#1}#3}
\creflabelformat{algorithm}{#2\textcolor{red}{#1}#3}

\Crefname{equation}{Eq.}{Eqs.}
\Crefname{figure}{Fig.}{Figs.}
\Crefname{tabular}{Tab.}{Tabs.}
\Crefname{table}{Tab.}{Tabs.}
\Crefname{section}{Sec.}{Secs.}

\crefname{equation}{Eq.}{Eqs.}
\crefname{figure}{Fig.}{Figs.}
\crefname{tabular}{Tab.}{Tabs.}
\crefname{table}{Tab.}{Tabs.}
\crefname{section}{Sec.}{Secs.}

%% file: 1_intro.tex
\section{Introduction}
\label{sec:intro}

The sharp contrast between the ideal conditions found in standard benchmarks and the unpredictable nature of the real world majorly hinders the deployment of computer vision systems in the wild. Despite the most meticulous efforts, samples used to train and validate visual models will only represent a fraction of the diversity that these models will face once deployed.
It is thus critical to 
detect model vulnerabilities and bring them to the user's attention
in an interpretable way~\cite{DoshiVelezX17TowardsRigorousScienceInterpretableML}. 

Among research lines that focus on advancing model interpretability, a prominent one
is that of  \textit{explaining errors made by computer vision models via natural language descriptions}. Different works
have addressed the problem of  \textbf{L}anguage-\textbf{B}ased \textbf{E}rror \textbf{E}xplainability (\task{})
before~\cite{EyubogluICLR22DominoDiscoveringSystematicErrors,YenamandraICCV23FACTS,JainICLR23DistillingModelFailuresAsDirectionsInLatentSpace,WilesNIPSW22DiscoveringBugsInVisionModels,RezaeiICLR24PRIME}. Yet, none of them proposed ways to quantitatively assess the  predicted  descriptions, mainly confining them to a qualitative inspection of the model's error modes. We assert that the \task{} problem requires  appropriate metrics, in order
to rank methods and track progress in this field.  In this paper, we take different steps in this direction through the following contributions: \textit{i) }we provide a rigorous formalization of the \task{} problem, \textit{ii) }we propose a family of methods to tackle it, and \textit{iii) }we introduce  different metrics to evaluate performance.

\begin{figure*}[t!]
\centering
\includegraphics[width=\linewidth]{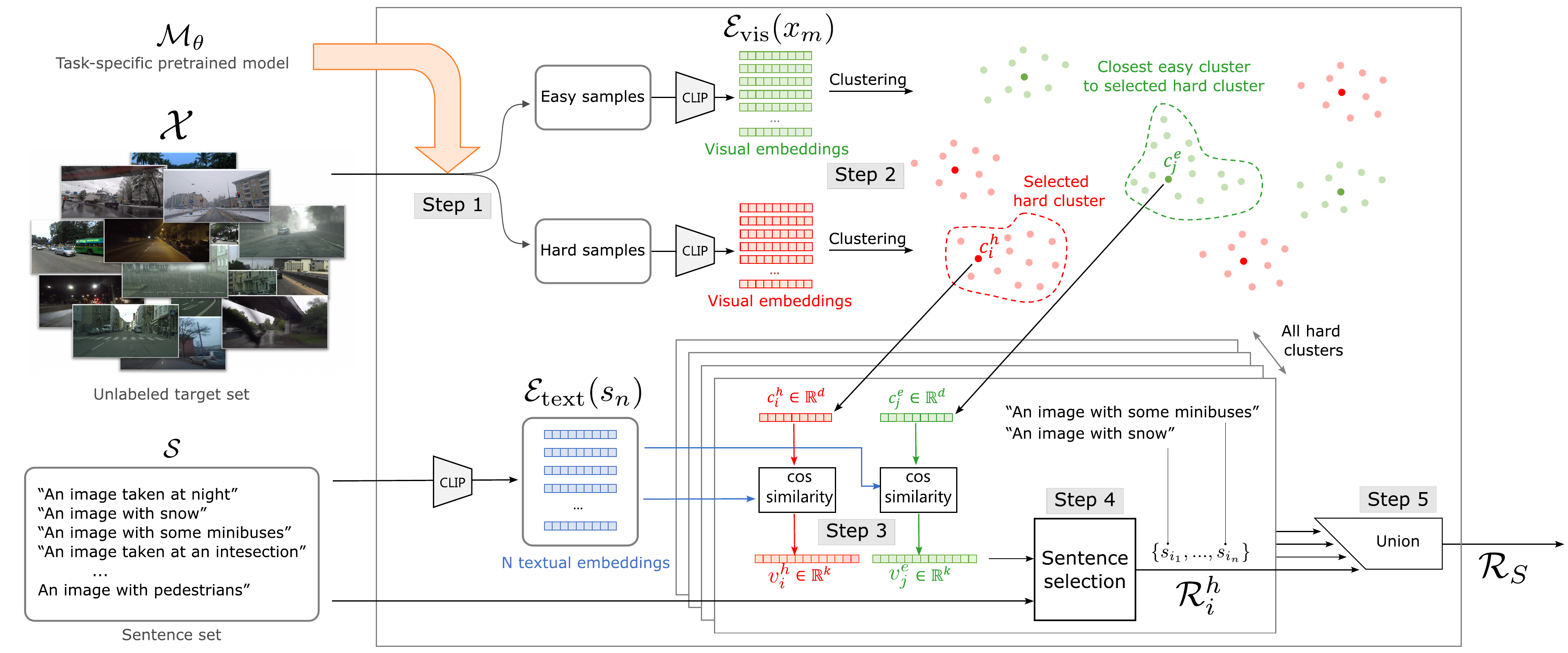}
    \caption{
    \textbf{Overview}. Provided with a pretrained model $\calM_\theta$, a target image set $\calX$, and a set of sentences $\calS$, the family of solutions we propose for \task{} follows the following steps. In \step{Step 1}, images from the target set $\calX$ are split into easy and hard samples, based on the model's confidence. In \step{Step 2}, samples are embedded using Open-CLIP and the visual embeddings of each set are clustered (independently). In \step{Step 3}, for each hard and easy cluster we compute the cosine similarities between the textual embeddings of  candidate sentences and the cluster prototypes. We also assign the closest easy prototype to each hard prototype. \step{Step 4} performs sentence selection for that hard cluster, based on these sentence-prototype similarities. 
    \step{Step 5} aggregates the cluster-specific sentence sets to produce the output.}
    \label{fig:method}
\end{figure*}

In short, the problem at hand is the following: Given a set of 
images, denoted by $\mathcal{X}$,
a task-specific pre-trained model $\mathcal{M}_{\theta}$, 
and a large sentence set $\calS$,
our desired output is a subset of sentences
$\calS^*\subset \calS$
that describes in plain text the samples from $\mathcal{X}$
on which the model $\mathcal{M}_{\theta}$ underperforms.
To tackle  this problem, we propose  
a simple, yet powerful family of approaches
that summarize the visual conditions deemed challenging for a given model via natural language. 
Our methods rely on uncertainty estimation to distinguish between samples for which
the model is more likely to perform well or poorly, and on clustering in a joint vision-and-language embedding space
to group together challenging samples associated with specific visual conditions. Operating in this joint 
representation space allows us to associate sentences to each identified 
cluster of samples (see illustration in \cref{fig:method}). In order to isolate sentences that are
unique to each  cluster composed of under-performing samples,  we propose different ways to contrast descriptions of the clusters on which the model struggles with the ones on which it performs performs well
(denoted as \textit{hard} and \textit{easy} clusters, respectively).

%%%%%%%%%%%%%%%%%%%%%%%%%%%%%%%%%%%%%%%%
%%%%%%%%%%%%%%%%%%%%%%%%%%%%%%%%%%%%%%%%

Our proposed family of approaches is unsupervised and task-agnostic, departing from previous methods which 
have been specifically designed  for classification~\cite{EyubogluICLR22DominoDiscoveringSystematicErrors,YenamandraICCV23FACTS,JainICLR23DistillingModelFailuresAsDirectionsInLatentSpace}, 
and often require access to ground-truth or privileged information, such as the types of 
errors to be discovered~\cite{EyubogluICLR22DominoDiscoveringSystematicErrors,YenamandraICCV23FACTS}.
To assess how well these  methods  find suitable
failure explanations, we  introduce metrics specifically designed for \task{} capturing whether or not the sentences satisfactorily explain reasons for failure, and 
how well the produced sentences describe the images within each error mode.  

While  the need to provide the sentence set $\calS$ is a practical limitation of our methods,  it plays a crucial role as it  allows
1) defining a more generic ground-truth (GT) for evaluation that goes
beyond human-in-the-loop qualitative assessment,  
2)  finding more subtle explanations than clear test set's biases (\eg
``waterbirds in land''~\cite{LiuICML21JustTrainTwiceImprovingGroupRobustnessWithoutTrainingGroupInformation},  ``blonde males'' in CelebA~\cite{LiuICCV15DeepLearningFaceAttributesWild} or 
human-implanted biases in \npp~\cite{EyubogluICLR22DominoDiscoveringSystematicErrors,YenamandraICCV23FACTS}), 
3) performing  a fair and  quantitative  comparisons of different methods, and
4) avoiding the need to involve the assessment with a large language model (\eg, as in ~\cite{DunlapCVPR24Visdiff}).

In summary, the contributions of our paper are: \textit{i) } formalizing the 
\task{} problem (\Cref{sec:prob-form});
\textit{ii) } proposing a family of methods to tackle it (\Cref{sec:method}); \textit{iii) } introducing  
different metrics to evaluate performance  (\Cref{sec:LBEquality}) \textit{iv)}; showing the effectiveness of our proposed solutions through extensive experiments, focusing on both classification and semantic segmentation tasks (\Cref{sec:res}).

%% file: 2_related_work.tex
\section{Related work}
\label{sec:RW}

\myparagraph{Automated error discovery}
Recently, methods to assess errors made by machine learning models
have shifted from human-in-the-loop approaches~\cite{WongICML21LeveragingSparseLinearLayers4DebuggableDeepNetworks,SinglaCVPR21UnderstandingFailuresDeepNetworksViaRobustFeatureExtraction,LiICLR22MissingnessBiasInModelDebugging,dEonFACCT22TheSpotlightDiscoveringSystematicErrors,GaoICCV23AdaptiveTestingOfCVModels} to automatic detection of consistent model failure modes~\cite{EyubogluICLR22DominoDiscoveringSystematicErrors,YenamandraICCV23FACTS,JainICLR23DistillingModelFailuresAsDirectionsInLatentSpace,WilesNIPSW22DiscoveringBugsInVisionModels,RezaeiICLR24PRIME}. 

In particular, Eyuboglu \etal\cite{EyubogluICLR22DominoDiscoveringSystematicErrors,YenamandraICCV23FACTS} propose to  learn a mixture model to partition the data according to classification errors related to spurious correlations. 
Jain~\etal\cite{JainICLR23DistillingModelFailuresAsDirectionsInLatentSpace} characterize model failures as directions in a latent space learned by SVMs. 
Kim~\etal\cite{KimCVPR24Bias2Text} identify biases  by analyzing captions from misclassified images with a Vision-Language Model (VLM) and rank the keywords from the captions using CLIP~\cite{RadfordICML21CLIP}. 
Wiles~\etal\cite{WilesNIPSW22DiscoveringBugsInVisionModels} use conditional text-to-image diffusion models to generate  synthetic images that are grouped based on how the model misclassifies them. 
Rezaei~\etal\cite{RezaeiICLR24PRIME} extract human-understandable concepts (tags) with VLMs and examine the model's behavior conditioning on the presence or absence of the combination of those tags.
Also Metzen~\etal\cite{MetzenICCV23IdentificationOfSystematicErrorsImgClassifRareSubgroups} rely on text-to-image models: by encoding specific information in the prompt, they synthesize images from the combinatorially large set of data sub-populations, and run tests to find out on which ones the model underperforms.
Finally, a recent work proposes methods to describe differences between two image sets~\cite{DunlapCVPR24Visdiff}. While this method  could be applied to \task{},  the proposed evaluation  based on GPT-4 can only assess  whether the difference between the sets is correct but not whether this difference is related or not to model failure.

In contrast with the above works, 
our method only require encoding the images and the textual explanations in a shared visual-textual space such as CLIP, where we rely on simple arithmetic operations among clusters to characterize them via natural language.
Our solutions seamlessly apply to arbitrary image task, and indeed we port this line of work to semantic segmentation for the first time.  Differently from previous work on this topic,
we also provide a rigorous formalization of the \task{} problem, with ad hoc metrics.

%%%%%%%%%%%%%%%%%%%%%%%%%%%%%%%%%%%%%%%%%%%%%%%%%%%%%%%%%%%%%%
\myparagraph{Explainability}
There is a growing body of work connecting explainable AI and CV. Several  focus on producing visual explanations 
visualizing image regions that contribute most to model failures~\cite{RibeiroACM16WhyShouldITrustYou,ZhangECCV18TopDownNeuralAttentionByExcitationBackprop,SelvarajuICCV17GradCAM,TursunX23TowardsSelfExplainabilityOfDNNWthHeatmapCaptioning}. 
A few works aim at generating counterfactual explanations
\cite{GoyalICML19CounterfatualVisualExplanations,HendricksICMLW18GeneratingCounterfactualExplanationsWithNLP,JacobECCV22STEEXSteeringCounterfactualExplanationsWithSemantics},
identifying small image edits that cause the model to fail.  More recently, 
diffusion models have been used to this end~\cite{JeanneretACCV22DiffusionModels4CounterfactualExplanations,AugustinNIPS22DiffusionVisualCounterfactualExplanations,VendrowICMLWS24DatasetInterfacesDiagnosingModelFailuresCounterfactualGeneration}. In contrast to these methods which only explain failures on a particular image, our aim is to provide explanations of the model's overall behavior on an entire image set, in order to uncover consistent error patterns and major weaknesses under particular conditions. Furthermore, we rely on natural language as a more interpretable interface.

Another body of work tackles interpretability by providing short text descriptions to describe why a model has produced some output~\cite{HendricksECCV16GeneratingVisualExplanations,KimECCV18TextualExplanationsForSelfDrivingVehicles,HendricksECCV18GroundingVisualExplanations,SammaniCVPR22NLXGPT}. Such methods are related to our work, in that we are
also interested in generating language descriptions for model behaviors; yet, we are
interested in \textit{failure modes}, not in the general rationale behind a prediction.

%% file: 3_prob_form.tex
\section{Language-Based Error Explainability}
\label{sec:prob-form}

We can define the \task{} problem as
automatically finding and describing potential errors from a model based on its performance on 
a given dataset.  Formally, let $\calM_{\theta}$ be a computer vision model parameterized by $\theta$,  for an arbitrary computer vision task, such as classification or segmentation.
Let $\calX =\{\bfx_i\}^M_{i=1}$  be a set of images sampled from an arbitrary target distribution,  and let $\omega^{\text{avg}}_{\theta}$ be the average performance of the model on such data.
Finally, let $\calS = \{s_n\}^N_{n=1}$ be a 
broad and generic  set of sentences  that  describe various visual elements and conditions, which can be either manually defined or generated with a large language model (LLM)---see \cref{sec:sentences}.

\myparagraph{Problem formulation}
Given a target set $\calX$, our goal is to find a set of sentences~$\calR_{\calS} \subset \calS$
describing any  likely failure causes for the model $\calM_{\theta}$ 
with respect to images from $\calX$---such as  
unseen objects, visual conditions never encountered during training, or errors of any other nature.
Intuitively, each selected sentence in $\calR_{\calS}$ should  describe elements present
in images on which the models fails and not in 
those in which the model succeeds.

Letting $\omega^{s_n}_{\theta}$ be the average performance 
of the model on images for which the sentence  $s_n$ is relevant,  we determine that such 
sentence is \textit{describing a failure mode} if the difference between the global average $\omega^{\text{avg}}_{\theta}$  (average over all images in $\calX$) 
and $\omega^{s_n}_{\theta}$ is larger than a threshold $\beta$, 
namely a predefined margin. 
The desired output $\calS^{*}_{\beta}$, used to evaluate the performance of the methods, is the collection of all of these sentences
\begin{equation}\label{eq:prob-form}
\calS^{*}_{\beta} = \{s_n \in \calS | \,  \omega_{\theta}^{s_n} <  \omega_{\theta}^\text{avg} - \beta  \} \enspace , 
\end{equation}
Note that $\beta$ allows the user to set an arbitrary severity level for the error descriptions, \ie, a higher $\beta$ restricts the desired set $\calS^{*}_{\beta}$ to more challenging sentences. 
We provide more details in \cref{sec:GTSentenceSet}.

\myparagraph{Relation with prior art}
To conclude this section, we position the formulation introduced above with the ones tackled
by related works~\cite{EyubogluICLR22DominoDiscoveringSystematicErrors,YenamandraICCV23FACTS,JainICLR23DistillingModelFailuresAsDirectionsInLatentSpace}. 

One core difference is that these methods limit their problem formulation to the classification task, while we do not make any assumption related to the nature of the task at hand. 
Furthermore, in \cite{JainICLR23DistillingModelFailuresAsDirectionsInLatentSpace} 
they analyze errors \textit{per class}, and
in \cite{EyubogluICLR22DominoDiscoveringSystematicErrors,YenamandraICCV23FACTS} 
assume access to ground truth or prior knowledge about the error types to look for (\eg, bias-conflicting sub-populations or context).  
In contrast to the above works, we do not use any knowledge about errors prior to  evaluation and neither assume, in general, access to class information. 
Finally, while in prior works  the language-based descriptions are  part of the 
proposed methodology, they are not part of the problem formulation and they are not quantitatively evaluated. Instead,  we treat the language-based
description of errors made by computer vision models as the problem itself, providing means to evaluate them. 

%% file: 4_method.tex
\section{A family of  approaches to solve \task{}}
\label{sec:method}

We propose a family of simple, unsupervised and task agnostic
approaches to tackle the \task{} problem. 
Our solutions rely on two key elements: \textit{i)} a joint vision-and-language
space  \jointSpace,  for which 
we use Open-CLIP~\cite{openclip21}, and \textit{ii)} the 
sentence set $\calS$ introduced in the previous section,
assumed to be large enough to describe images from the target set and to contain potentially relevant reasons for model failure (see \cref{sec:sentences}). 

 We recall that 
the problem at hand is the following: Given a set of  images, denoted by $\mathcal{X}$,
a task-specific pretrained model $\mathcal{M}_{\theta}$,  and a large sentence set $\calS$,
we want to select the subset from  $\calS$ that describes  the samples from $\mathcal{X}$
on which the model $\mathcal{M}_{\theta}$ underperforms. Our proposed solution
is illustrated in \cref{fig:method} and follows the steps detailed below. 

\myparagraph{Step 1: Splitting the target set into easy and hard subsets}
Given the model $\calM_{\theta}$ and a confidence measure $\varphi_\theta$,
we define two thresholds $t_{\varphi}^h$ and $t_{\varphi}^e$ and split the target set $\calX$ into three sets: an easy one 
$\calX^e = \{\bfx_i \in \calX | \varphi_\theta(\bfx_i) > t_{\varphi}^e\}$, 
a hard one $\calX^h = \{\bfx_i \in \calX | \varphi_\theta(\bfx_i) < t_{\varphi}^h\}$, and, 
if $t_{\varphi}^e > t_{\varphi}^h$, a neutral set with the remaining images. If %GT 
ground truth for the target set is available, we can instead use the model's performance 
to split the data.

\myparagraph{Step 2: Clustering easy and hard subsets}
Let us denote by $\calE_{\text{vis}}$ and $\calE_{\text{txt}}$ the visual and text encoders mapping images and sentences into the joint space \jointSpace, respectively. 
Let $\Phi = \{\calE_{\text{vis}}(\bfx_m)\}^M_{m=1}$  
be the set of visual representations of images in $\calX$ and
let $\Phi^h$
and $\Phi^e$ be the set of representations
associated with hard and easy samples, respectively. We cluster the samples in
$\Phi^h$ and $\Phi^e$ independently, using an arbitrary clustering method\footnote{We use Agglomerative Clustering with Ward-linkage~\cite{WardAMS63HierarchicalGrouping2OptimizeObjFunction} in our experiments.}
and  represent each cluster by its centroid, referred also as its \textit{prototype}.
We compute it as the average of the embeddings of the samples in the cluster.  Let $\calC^{h} = \{\bfc^h_i\}_{i=1}^{|\calC^h|}$ and $\calC^{e} = \{\bfc^e_j\}_{j=1}^{|\calC^e|}$ be the set of prototypes associated with hard and easy samples, respectively. In the remainder of the paper, we will use  $\bfc^h_i$ and $\bfc^e_j$ to  indistinctly refer to the clusters or to their prototypes.

\myparagraph{Step 3. 
Matching sentences with prototypes}
Let us denote by $\bfs_n=\calE_{\text{text}}(s_n)$ the textual embeddings of the sentences $s_n$. These 
embeddings lie in the same joint embedding space as the images and, in turn, as the prototypes $\bfc^e_i$ and $\bfc^h_i$. Therefore, for each prototype $\bfc$ (easy or hard) we can compute the cosine similarity (\ie, the dot product between L2-normalized representations) to each textual embedding $\bfs_n$. 
We can store these similarities into a vector $\bfv \in \mathbb{R}^{N}$, where the \nth element of this vector is $<\bfc, \bfs_n>$.  This allows us to characterize each cluster in terms of its semantic similarity with the sentences in $\calS$. 

\myparagraph{Step 4: Retrieving sentences for hard prototypes}
Here, we look for
a set of peculiar sentences for each hard cluster. 
To  select relevant sentences for a given cluster (easy or hard), we   simply rank the sentences by their similarity to the cluster center (by
ranking the elements of the vectors $\bfv^e_j$ and $\bfv^h_i$). Then we either retain the top  ranked sentences or all sentences with a corresponding value in $\bfv$ above a predefined threshold $\tau$.  We denote the set of retained sentences  for  $\bfc^h_i$ and 
$\bfc^e_j$ by $\calS^h_i$ and $\calS^e_j$,  respectively.  
While $\calS^h_i$ effectively describes images within $\bfc^h_i$, not all sentences will point to
failure reasons. We need to find sentences describing the visual features that make the cluster \textit{hard}.
We propose therefore to contrast the hard cluster with its closest easy cluster:
its proximity in \jointSpace implies content similarity, hence, contrasting 
allows isolating the attributes that characterize the hard cluster specifically, and not the easy one. 

Let  $\bfc_j^e \in \calC^e$ be the closest easy prototype  to $\bfc_i^h \in \calC^h$ based on their distance in \jointSpace. In the following, we propose three different methods to
isolate  sentences that are peculiar to the hard cluster $\bfc_i^h$.  For a fair and more straightforward comparison between different methods,  we fix 
the number of  sentences retained for each cluster and for each method to a predefined $K$,  
\ies  $\lvert \calR_i^h \rvert = K$, where $\calR_i^h$ is the sentence set retained by a method for the hard cluster $\bfc_i^h$.

\myparagraph{SetDiff: Sentence set differences} 
Given a pair of clusters ($\bfc^h_i$,$\bfc^e_j$), 
we first select relevant sentences for both---$\calS_i^h$ and $\calS_j^e$---and then remove from $\calS_i^h$  the sentences that are present in $\calS^e_j$, yielding $\calR^h_i=\calS^h_i \setminus \calS_j^e$. Since this approach does not guarantee that $\lvert \calR_i^h \rvert = K$,
we fill $\calS_j^e$ with  sentences corresponding to values in  $\bfv^e_j$ above a threshold\footnote{We set $\tau=0.25$ based on the observation that the average similarity between our sentence sets and image sets is around 0.17, due to the well-known modality gap~\cite{LiangNIPS22MindTheGapUnderstandingModalityGap}.} $\tau$ and 
build $\calR^h_i$  with the $K$ most similar sentences to $\bfc^h_i$ that are not in $\calS^e_j$.

\myparagraph{PDiff: Prototype difference}
Assuming that higher scores in the  element-wise difference between the similarity vectors
$\bfd^h_i =\bfv^h_i-\bfv^e_j$  better describe the hard clusters than the easy ones\footnote{This, up to some normalization, is equivalent to first consider the prototype differences
and then compute the similarity between each sentence $s_n$  and $\bfd^h_i$.},  we can 
rank $\bfd^h_i$  and retain the sentences corresponding to the top $K$ values.
We denote this set by $\calS^d_{i}$  and define $\calR^h_i=\calS^d_{i}$.

\myparagraph{FPDiff: Filtered PDiff}
A drawback of \pdiff is
that it does not  guarantee that the selected sentences accurately describe the hard cluster. A remedy to this is only retaining sentences that have high similarity to the prototype 
$\bfc^h_i$,  namely
sentences in $\calS^d_{i}$ that are also in $\calS^h_i$. We define $\calR^h_i=\calS^h_i \cap \calS_i^d$. Also here, to guarantee that $\lvert \calR_i^h \rvert = K$, we fill $\calS_i^h$ with  sentences corresponding to values in  $\bfv^h_i$ above a threshold $\tau$ and 
build $\calR^h_i$  with the top $K$ sentences having the highest values in $\bfd^h_i$  that are also in $\calS^h_i$. 

\myparagraph{TopS} We also include a baseline called \tops,  describing the  hard clusters without any contrasting, 
by selecting
the top  $K$  ranked sentences from $\calS^h_i$ ($\calR^h_i= \calS^h_i$). 

\myparagraph{Step 5: Producing the final output} The final set of sentences that describe potential reasons for failure for the whole dataset is the union of all the sentence sets retained for all the hard clusters, namely  
\begin{equation}
  \calR_{\calS} = \bigcup\nolimits_{i=1}^{|C^h|}
  \calR^h_i \enspace. 
  \label{eq:unionRS}
\end{equation}

We conclude by positioning these methods with respect to  prior works. 

\myparagraph{Relations with prior art}
As mentioned in~\cref{sec:prob-form}, prior art~\cite{EyubogluICLR22DominoDiscoveringSystematicErrors,YenamandraICCV23FACTS}
describes a problem formulation that does not perfectly align with ours. 
Yet, their approaches proposed to assign
explanations to predicted error modes can be related to some of the methods we propose above. DOMINO~\cite{EyubogluICLR22DominoDiscoveringSystematicErrors}, which associates sentences based on 
average CLIP embedding from which the average class representative is extracted, is closely related to \pdiff:  when we analyze the errors per class (see \cref{sec:res}), 
this version of our method can be interpreted as a variant of DOMINO in which the class prototype is replaced with the closest easy prototype.
FACTS~\cite{YenamandraICCV23FACTS} relies on a captioning  tool~\cite{MokadyX21ClipCapCLIPPrefix4ImageCaptioning} to assign a single tag to each partition without contrasting,  hence, it is mostly related to \textbf{TopS}.

%% file: 5_metrics.tex
\section{Evaluation metrics for \task{}}
\label{sec:LBEquality}

Our third contribution is a set of metrics to evaluate predicted language explanations by \task{} methods, which can retrieve error-related sentences without any supervision for an arbitrary visual task. 
Concretely, given the output of a method, namely
the set of  sentences 
$\calR_\calS \subset \calS$, we need to assess
\textit{i)} \textit{whether these sentences 
actually
point to reasons for model failure}, 
\textit{ii)} \textit{how well the retrieved sentences characterize the image set  assigned to the cluster}, 
and   \textit{iii)} \textit{how well the 
predicted sentence set  $\calR_\calS$
covers the set of potential explanations given by 
$\calS^{*}_{\beta}$}.
We propose metrics that allow measuring a method's performance in these dimensions.

\myparagraph{\textit{i)} Average Hardness Ratio (AHR)}
Our main goal is retrieving sentences $s_n$ that \textit{point to reasons for model failure}, \ie,
$s_n \in \calS^{*}_{\beta}$. This means that the hardness score of the sentence $\omega_{\theta}^{s_n}$---formally defined in \cref{eq:hardness},  \cref{sec:GTSentenceSet}--- satisfies the 
condition $\omega_{\theta}^{s_n} < \omega_{\theta}^{avg} - \beta$. 
The lower this value is, the more the sentence is correlated with model failure. 

We define the  Hardness Ratio (HR) for a given cluster as the ratio of sentences among the retrieved ones in $\calR_i^h$ (see~\cref{sec:method}) for which  the above condition holds
\begin{equation}
\label{eq:AHR}
\text{HR}_i=\frac{\lvert  \calR_i^h
\cap \calS^{*}_{\beta} \rvert}{\lvert  \calR_i^h \rvert} \enspace , 
\end{equation}

and we define the AHR as the average across all  clusters

\begin{equation*}
\text{AHR}=\frac{1}{|C^h|} \sum_{i=1}^{|C^h|} \text{HR}_i  \enspace .
\end{equation*}

\myparagraph{\textit{ii)} Average Coverage Ratio (ACR)}
Next, we introduce a metric to assess
the \textit{coverage} of  the
sentences associated with the hard clusters, \ies
the ratio of images in the cluster for which the retrieved
sentence holds. 
Intuitively 
this shows  how well a sentence $s_n \in \calR^h_i$  characterizes a hard cluster  
$\bfc^h_i$. $\text{CR}_i$ is  the mean of these ratios taken over the sentences retained in  $\calR^h_i$. Formally,
\begin{equation}\label{eq:ACR}
\text{CR}_i = \frac{1}{\lvert \calR_i^h \rvert} \sum_{s_n \in  \calR_i^h} 
\frac{1}{\lvert \bfc^h_i \rvert} \sum_{\bfx_m \in  \bfc^h_i} \Gamma (\bfx_m,s_n) \enspace ,
\end{equation}
where 
$\Gamma(.,.)$ is
a binary operator that, taking in input an image and a sentence, outputs 1 if the sentence 
$s_n$ is relevant for the image $\bfx_m$ and zero otherwise (see~\cref{sec:GTSentenceSet} for details). 
To obtain the final ACR value, we simply average $\text{CR}_i$ across all clusters, 
namely $$\text{ACR}=\frac{1}{|C^h|} \sum_{i=1}^{|C^h|} \text{CR}_i  \enspace .$$

\myparagraph{\textit{iii)}
$\calR_\calS$ \textit{vs.} $\calS^{*}_{\beta}$}
To assess if the union of the retained sentences 
$\calR_\calS$ (defined in \cref{eq:unionRS})
accurately covers
the potential errors represented by 
$\calS^{*}_{\beta} \subset\calS$, we compute the True Positive Rate (TPR) of the retrieved sentences 
and the Jaccard Index (JI) between the two sets 
\begin{equation}
\text{TPR}=\frac{\lvert  \calR_\calS \cap \calS^{*}_{\beta} \rvert}{\lvert  \calS^{*}_{\beta} \rvert} \,\,\, \text{and} \,\,\,  \text{JI}=\frac{\lvert  \calR_\calS \cap \calS^{*}_{\beta} \rvert}{\lvert  \calR_\calS \cup \calS^{*}_{\beta} \rvert} \enspace .
\end{equation} 
TPR indicates how well a method covers the  ground-truth explanations,
while JI measures the overall coverage, also taking into account false positives.

%% file: 6_results.tex
\begin{figure*}[!ttt]
    \centering
    \includegraphics[width=0.5\linewidth]{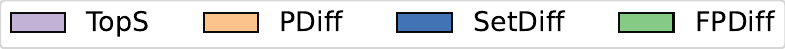}\\
\includegraphics[width=0.24\linewidth]{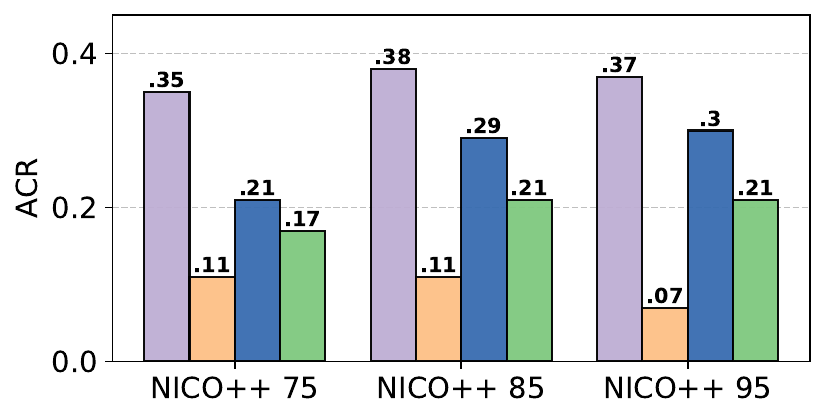}
\includegraphics[width=0.24\linewidth]{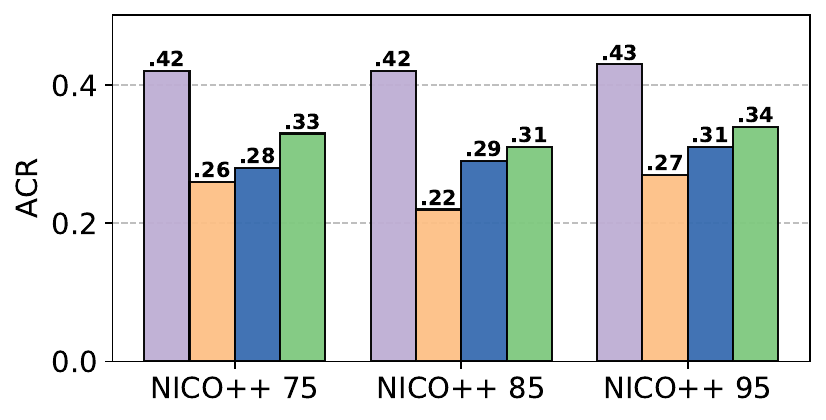}
\includegraphics[width=0.24\linewidth]{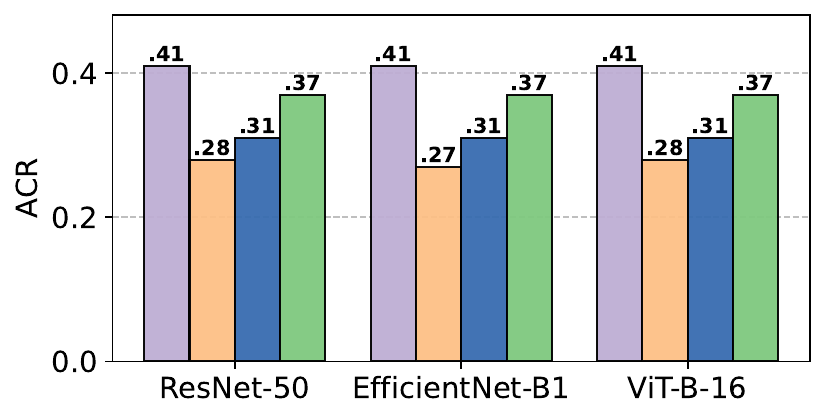}
\includegraphics[width=0.24\linewidth] {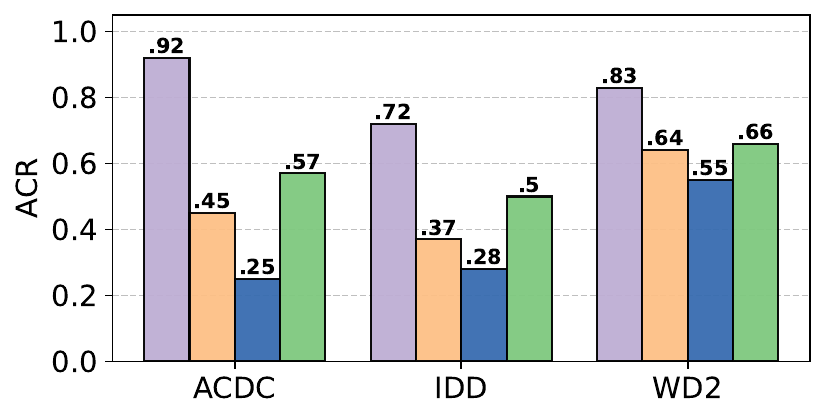}
 \\
\includegraphics[width=0.24\linewidth]{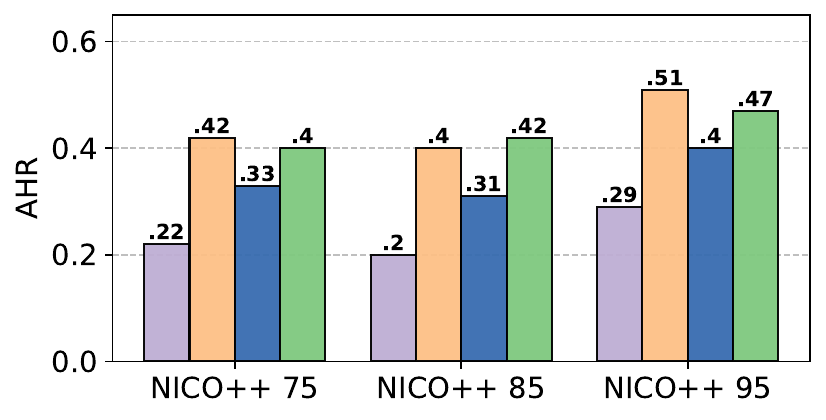}
\includegraphics[width=0.24\linewidth]
{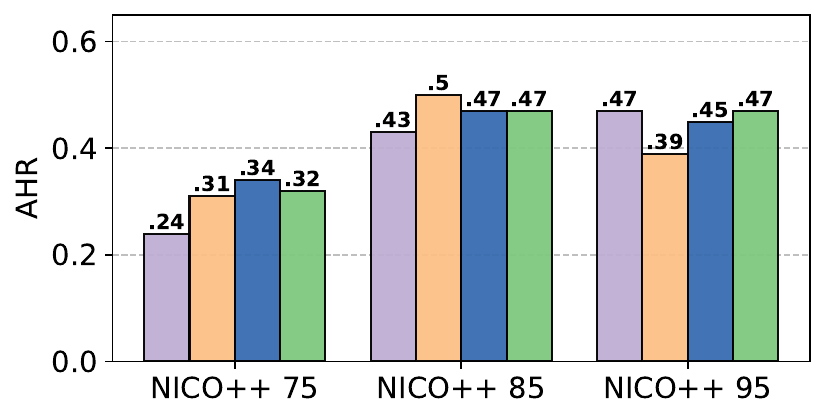}
\includegraphics[width=0.24\linewidth]{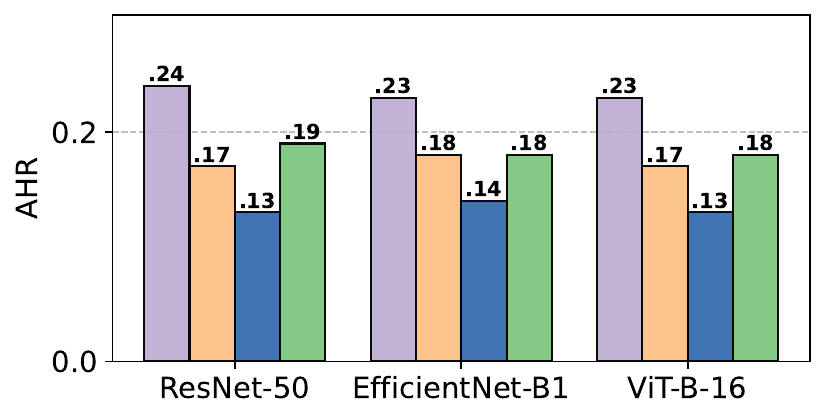}
\includegraphics[width=0.24\linewidth]{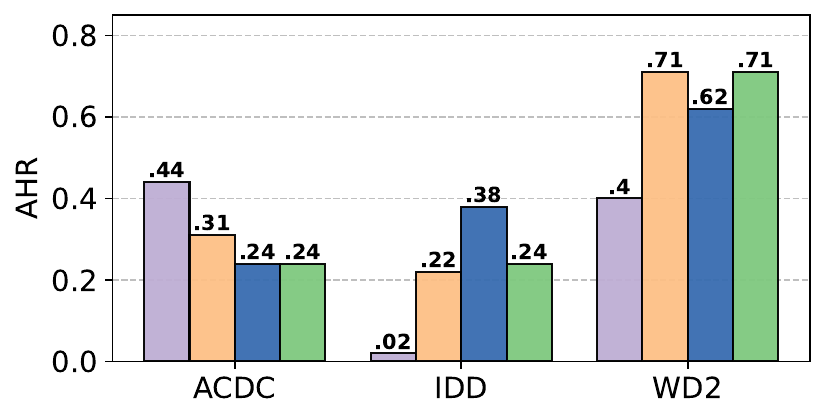}\\
\includegraphics[width=0.24\linewidth]{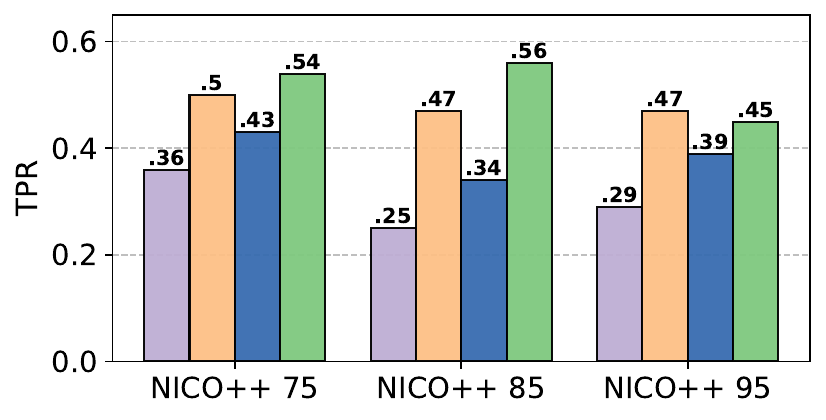}
\includegraphics[width=0.24\linewidth]{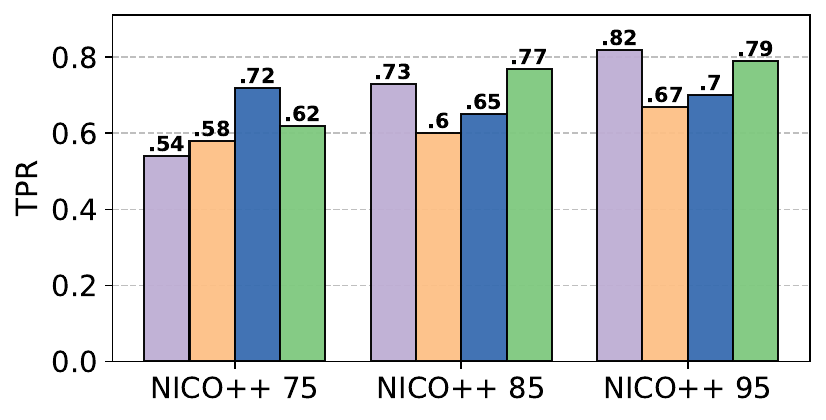}
 \includegraphics[width=0.24\linewidth]{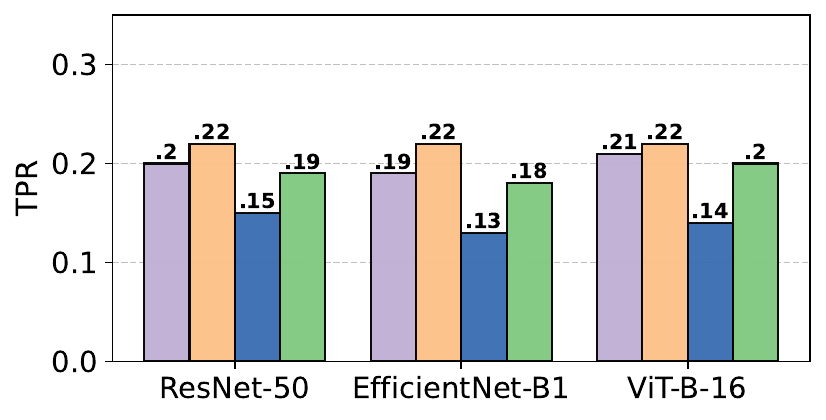} 
 \includegraphics[width=0.24\linewidth]{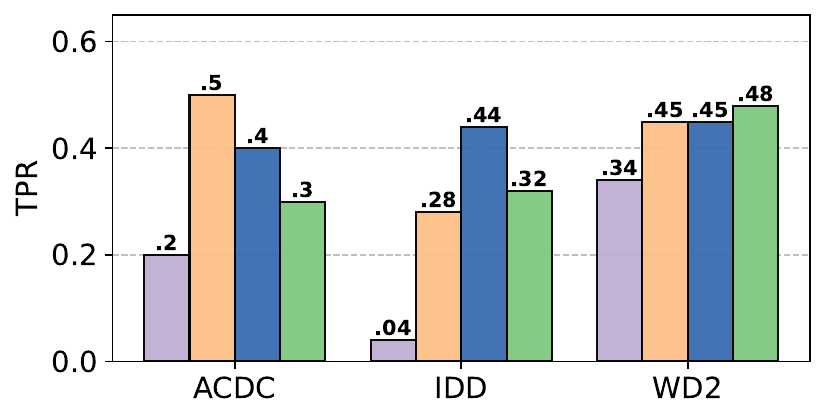}\\

\includegraphics[width=0.24\linewidth]{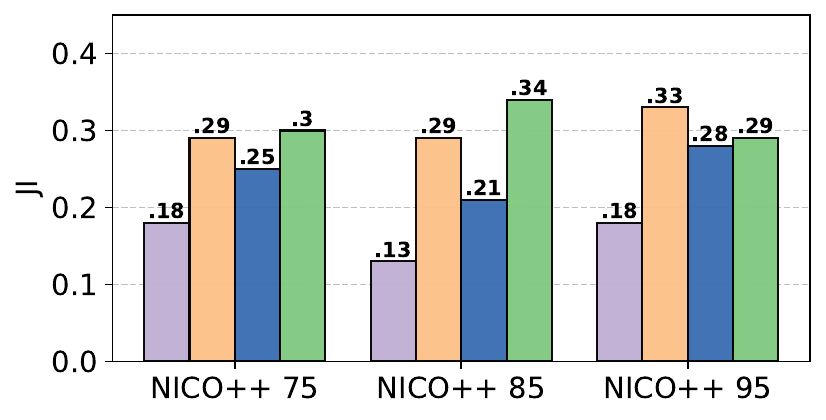}
\includegraphics[width=0.24\linewidth]{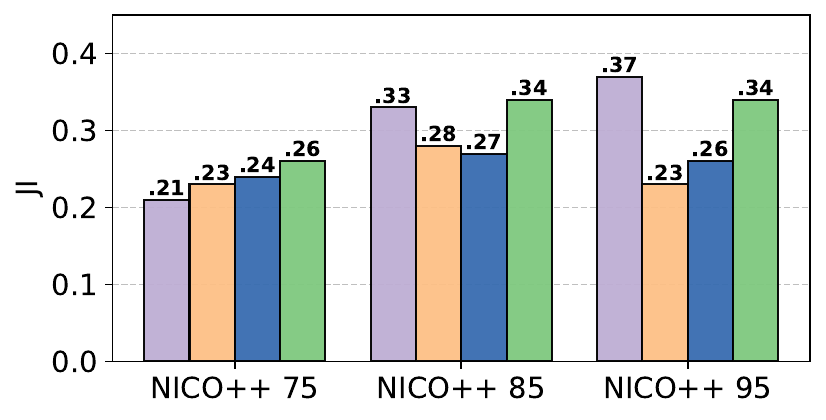}
\includegraphics[width=0.24\linewidth]{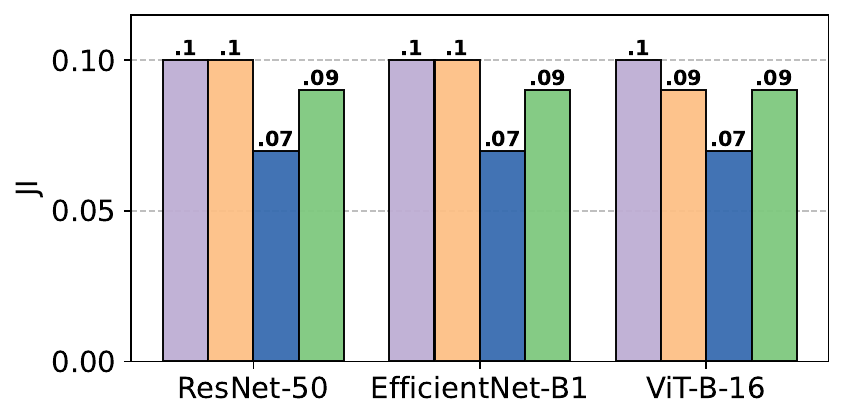} 
\includegraphics[width=0.24\linewidth]{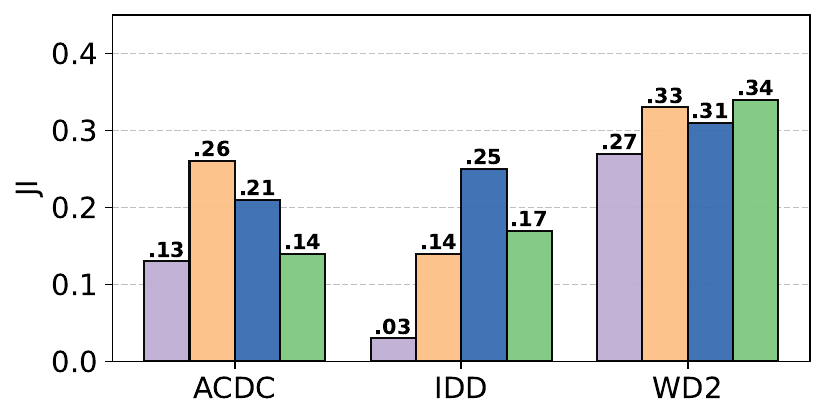}\\
 \hspace{0.4cm}\textbf{\small Unsupervised \npp } 
 \hspace{+0.9cm}  \textbf{\small Per-class   \npp}  
 \hspace{+1.8cm} \textbf{\small ImageNet}   \hspace{+1.8cm}\textbf{\small Urban segmentation} \\
    \caption{\textbf{Numerical results (all datasets).} From top  to bottom: ACR, AHR, TPR and JI   scores  on \npp{} unsupervised  and  supervised per-class (first and second column, respectively),   ImageNet per-class  (third column) and Urban Scene Segmentation  (last column).
     } 
    \label{fig:all_res}
\end{figure*}

\section{Experiments}
\label{sec:res}

\input{TPFPUrbanSegmSmall}

\myparagraph{Experimental setup}
We tackle three tasks: semantic segmentation of urban scenes, classification under the presence of spuriously 
correlated data, and large-scale ImageNet  classification. 
For the first, we consider a  ConvNeXt~\cite{LiuCVPR22AConvNet} segmentation model
trained on Cityscapes~\cite{CordtsCVPR16CityscapesDataset} and test on three challenging datasets:
WD2~\cite{ZendelECCV18WildDashCreatingHazardAwareBenchmarks}, IDD~\cite{VarmaWACV19IDDDatasetExploringADUnconstrainedEnvironments} and ACDC~\cite{SakaridisICCV21ACDCAdverseConditionsDatasetSIS}. For the second task we train
ResNet50 image classification models  on different spuriously correlated data derived from \npp
and demonstrate how the proposed approaches can be used for understanding model failures due to such subtle biases~\cite{zhao2017men,hendricks2018women}.  We consider two cases, an unsupervised case where  the full set is  split  with class prediction entropy and a supervised one,  where following  \cite{EyubogluICLR22DominoDiscoveringSystematicErrors,YenamandraICCV23FACTS} we analyze the performance on  each class separately.  In the latter case  we use the class prediction to split the data into easy and hard sets and, in addition to evaluating LBEE, we compare  our simple split-and-cluster partitioning  to the more complex 
partitioning methods from \cite{EyubogluICLR22DominoDiscoveringSystematicErrors,YenamandraICCV23FACTS}, using their evaluation protocol (see in \cref{sec:npp}). 
Finally,  to showcase the scalability of our approaches, we analyze different pre-trained architectures 
on the ImageNet 1K validation set, focusing on the top-1 classification performance and seeking for explanations for each class individually. We provide details about each task, related models and the datasets 
in \cref{sec:datasets}.

\myparagraph{Default design choices} 
We use Open-CLIP~\cite{openclip21} 
as vision-and-language embedding space. 
$\calS$ is a User-defined sentence set in the case of  urban scene segmentation and ImageNet and GT-based for \npp (see  details in \cref{sec:construcS}).
To generate the  ground-truth sentence set $\calS^{*}_{\beta}$ containing the sentences  with  valid \textit{hardness score} (see \cref{sec:GTSentenceSet}),
we set $\beta = .2 *\omega_{\theta}^{\text{std}}$ with $\omega_{\theta}^{\text{std}}$ being the standard deviation of $\omega_{\theta}$ computed over $\calX$.  
We set the number of hard clusters  to $C=15$ for large datasets  and $C=5$ for the per-class data analyses. We use the same number of clusters for the easy set ($|\calC^h|=|\calC^e|=C$). To split the data we use the output entropy in the unsupervised cases and class probability or ranking in the per-class case (see details in  \cref{sec:param_select}). We retain three sentences for each cluster ($K=3$).

In the following, we show numerical results and illustrate them with some qualitative examples.
We report in-depth analyses for the different datasets, also concerning the sensitivity to crucial
hyper-parameters.

\begin{figure*}[ttt]%
{\footnotesize
    \centering
     {{\includegraphics[width=.9\linewidth]{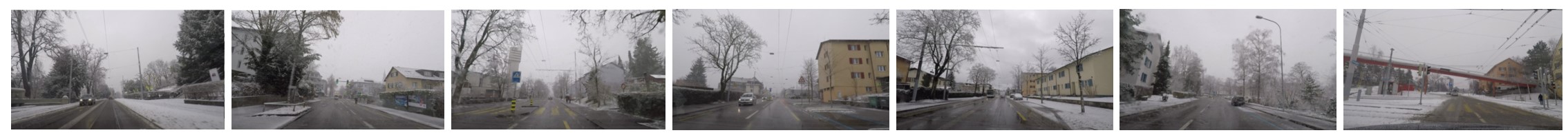} }} \\
     \begin{tabular}{cccc}
    \tops  & \setdiff    & \pdiff    & \fpdiff \\
 "\textit{taken in a snowy weather}"   \,\, & \,\,  "\textit{with underexposure}"   \,\, & \,\,  "\textit{taken in a foggy weather}"  & "\textit{taken in a foggy weather}"\\
"\textit{taken in a cloudy weather}"   \,\, & \,\,  "\textit{showing a city scene}"    \,\, & \,\,  "\textit{taken in a windy weather}"  \,\, & \,\,  "\textit{taken in a windy weather}"\\
"\textit{taken in a dull weather}"    \,\, & \,\, "\textit{with low contrast}"  \,\, & \,\,"\textit{taken in a dull weather}"  \,\, & \,\, "\textit{taken in a dull weather}"
    \end{tabular} \\ 
    \vspace{0.3cm}
    \includegraphics[width=0.9\linewidth]{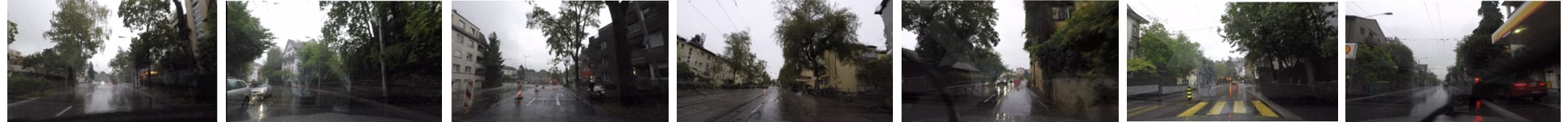}\\
     \begin{tabular}{cccc}
    \tops \,\, & \,\, \setdiff  \,\, & \,\, \pdiff  \,\, & \,\, \fpdiff \\
 ``\textit{taken in a rainy weather}''  \,\, & \,\,  ``\textit{shadows on the road}''  \,\, & \,\,  ``\textit{water on the road}'' \,\, & \,\,  ``\textit{water on the road}'' \\
``\textit{taken in a dull weather}''  \,\, & \,\, ``\textit{people on the road}''   \,\, & \,\,  ``\textit{taken in a rainy weather}'' \,\, & \,\,  ``\textit{taken in a rainy weather}''\\
``\textit{taken in a cloudy weather}''    \,\, & \,\, ``\textit{of a sidewalk}'' \,\, & \,\, ``\textit{rickshaw on the road}''  \,\, & \,\, ``\textit{taken in a stormy weather}''
    \end{tabular}\\
    }
    \caption{ Two hard clusters in ACDC explained  by different methods. We illustrates a few example images from the cluster and below the three sentences retained  by different methods for it.
    }    
    \vspace{0.5cm}
\label{fig:qualityACDC}
\end{figure*}

\begin{figure*}[!ttt]
    \centering
    \includegraphics[width=0.5\linewidth]{legend_ADSeg_TPR_bar.pdf}\\
\includegraphics[width=0.24\linewidth] {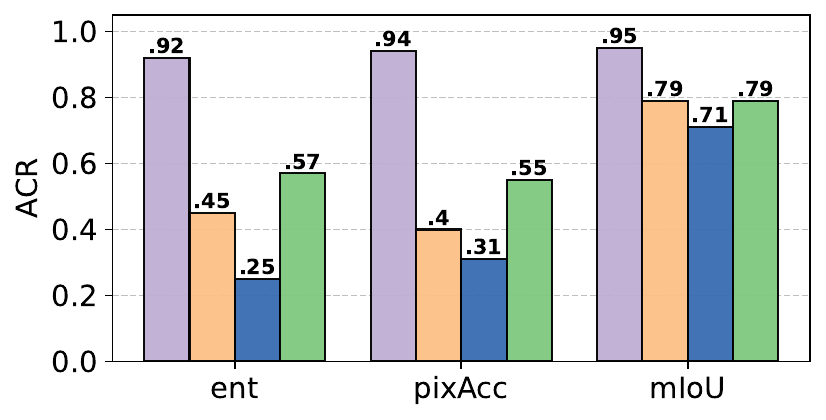}
\includegraphics[width=0.24\linewidth]{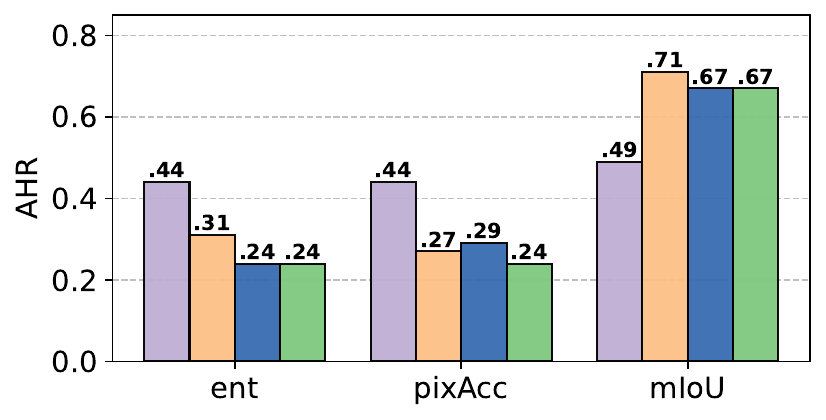}
\includegraphics[width=0.24\linewidth]{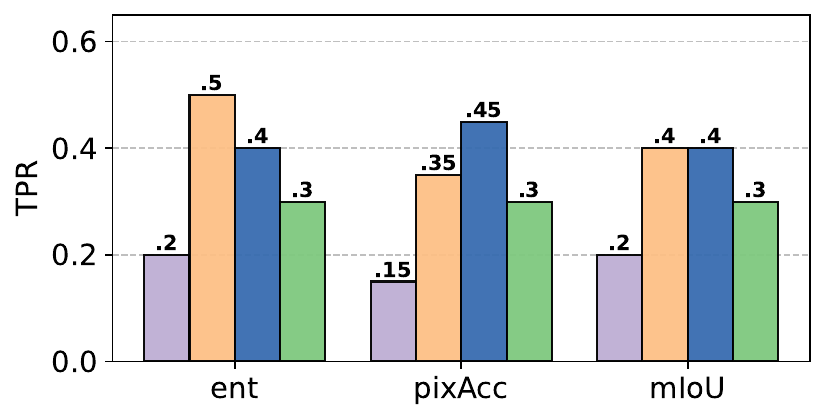}
\includegraphics[width=0.24\linewidth]{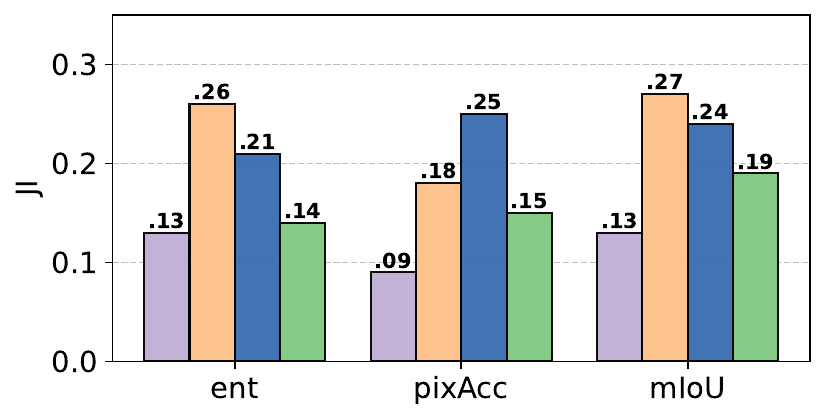}
\\
    \caption{\textbf{Unsupervised and supervised splitting (ACDC).} Results of different metrics 
 (from left to right  ACR, AHR, TPR  and JI) on the ACDC dataset  when  the split is done with entropy, pixel accuracy  and the  mIoU.  } 
    \label{fig:ent_vs_pa_miou}
\end{figure*}

\subsection{Results}
\label{sec:quantitativeMain}

\begin{figure*}[ttt]%
{\footnotesize
    \centering
    {{\includegraphics[width=.9\linewidth]{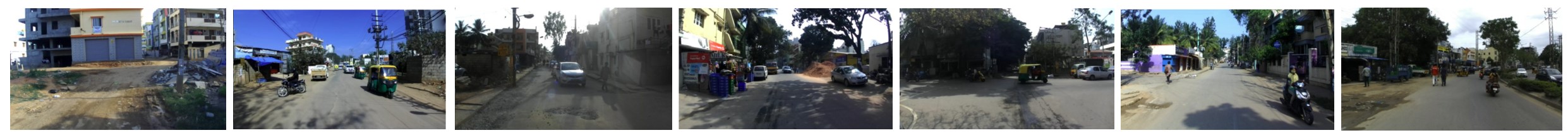} }}\\
     \begin{tabular}{cccc}
    \tops & \setdiff  & \pdiff  & \fpdiff \\
 "\textit{rickshaw on the road}"  \,\,\,\, & \,\,\,\,  "\textit{shadows on the road}"  \,\,\,\, & \,\,\,\, "\textit{dirt on the road}"  \,\,\,\, & \,\,\,\, "\textit{dirt on the road}"\\
"\textit{construction on the road}"  \,\,\,\, & \,\,\,\, "\textit{residential scene}"  \,\,\,\, & \,\,\,\, "\textit{countryside scene}"    \,\,\,\, & \,\,\,\, "\textit{construction on the road}"\\
"\textit{sub-urban scene}"    \,\,\,\, & \,\,\,\, "\textit{debris on the road}"  \,\,\,\, & \,\,\,\,"\textit{mud on the road}" \,\,\,\, & \,\,\,\,"\textit{debris on the road
}"
    \end{tabular} \\
     \vspace{0.5cm}
    \includegraphics[width=0.9\linewidth]{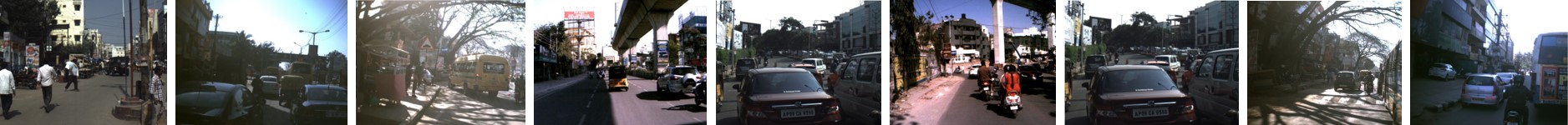}\\
    \begin{tabular}{cccc}
    \tops & \setdiff    & \pdiff  & \fpdiff \\
   ``\textit{traffic}'' \,\, & \,\,  ``\textit{shadows on the road}'' \,\, & \,\,  ``\textit{highway scene}'' \,\, & \,\,  ``\textit{highway scene}''\\
   ``\textit{rickshaw on the road}'' \,\, & \,\,
    ``\textit{crowded background} \,\, & \,\,  ``\textit{traffic jam}'' \,\, & \,\,  ``\textit{traffic jam}''\\
   ``\textit{showing an sub-urban scene}'' \,\, & \,\,  ``\textit{light reflections on the road}''  \,\, & \,\,  ``\textit{advertising board}'' \,\, & \,\,  ``\textit{traffic}''
    \end{tabular}\\
    }
    \caption{
     Two hard clusters  from IDD explained  by different methods. We illustrates a few example images from the cluster and below the three sentences retained  by different methods for it.}    
     \vspace{0.5cm}
\label{fig:qualityIDD}
\end{figure*}

\begin{figure*}[ttt]
{\footnotesize
    \centering
    {{\includegraphics[width=.85\linewidth]{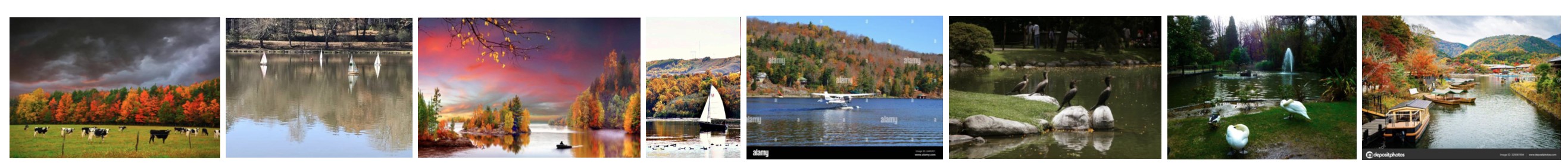} }}\\
     \begin{tabular}{cccc}
    \tops & \setdiff    & \pdiff  & \fpdiff\\
 ``\textit{taken in autumn}''  \,\, \,\, & \,\, \,\, ``\textit{taken in autumn}''  \,\, \,\, & \,\, \,\, ``\textit{sailboat in the water}'' 
\,\, & \,\,  ``\textit{sailboat in the water}'' \\
``\textit{with waterways}''  \,\, & \,\,``\textit{with waterways}''   \,\, \,\, & \,\, \,\, ``\textit{sailboat in the rocks}''  \,\, \,\, & \,\, \,\, 
 ``\textit{sailboat}''\\
``\textit{plants in the water}''    \,\, & \,\,``\textit{plants in the water}'' \,\, & \,\,``\textit{goose in the water}'' \,\, & \,\,``\textit{birds in the water}''
    \end{tabular} \\
            {{\includegraphics[width=.85\linewidth]{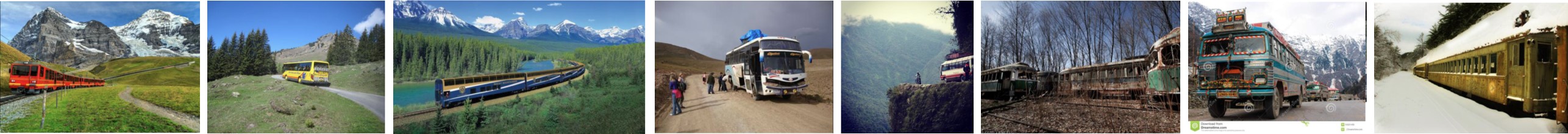} }} \\
     \begin{tabular}{cccc}
    \tops \,\, & \,\,\setdiff   \,\, & \,\,\pdiff   \,\, & \,\,\fpdiff \\
 ``\textit{train}''  \,\, \,\, & \,\, \,\, ``\textit{train}''  \,\, \,\, & \,\, \,\, ``\textit{ train in the rocks}'' \,\, \,\, & \,\, \,\, ``\textit{ train in the rocks}''\\
``\textit{train in the rocks}''  \,\, & \,\,``\textit{train in the rocks}''   \,\, \,\, & \,\, \,\, ``\textit{bus in the rocks}'' \,\, \,\, & \,\, \,\, ``\textit{bus in the rocks}''\\
``\textit{bus in the rocks}''    \,\, & \,\,``\textit{bus in the rocks}'' \,\, & \,\,``\textit{train in the grass}'' \,\, & \,\,``\textit{train in the grass}''
    \end{tabular} \\
{{\includegraphics[width=.85\linewidth]{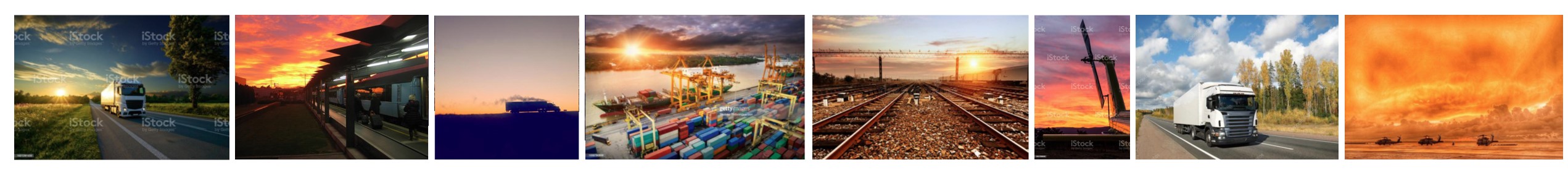} }}\\
  \begin{tabular}{cccc}
    \tops \,\, & \,\,\setdiff   \,\, & \,\,\pdiff   \,\, & \,\,\fpdiff \\
 ``\textit{truck taken at sunset}''  \,\, \,\, & \,\, \,\, ``\textit{truck taken at sunset}''  \,\, \,\, & \,\, \,\, ``\textit{truck taken at sunset}''  \,\, \,\, & \,\, \,\, ``\textit{truck taken at sunset}''\\
``\textit{bus taken at sunset}''  \,\, & \,\,``\textit{bus taken at sunset}''   \,\, \,\, & \,\, \,\, ``\textit{bus taken at sunset}'' \,\, \,\, & \,\, \,\, ``\textit{bus taken at sunset}''\\
``\textit{train taken at sunset}''    \,\, & \,\,``\textit{train taken at sunset}'' \,\, & \,\,``\textit{truck in the water}''  \,\, & \,\,``\textit{truck in the water}''
    \end{tabular} \\
    }
    \caption{
   Three out of 15 hard clusters from \nppm  (unsupervised case)   explained by different methods.
    }    
\label{fig:qualitativeNPPmain}
\end{figure*}

In~\cref{fig:qualityACDC,fig:qualityIDD,fig:qualitativeWD2Sent,fig:qualitativeNPPmain,fig:qualityEffNet} we report qualitative results, showing the sentences selected by our methods for various hard clusters. 
In all cases, each row represents a hard cluster, and at the bottom we report the three sentences provided by the different methods. 

Focusing, \eg, on~\cref{fig:qualityACDC,fig:qualityIDD,fig:qualitativeWD2Sent}  (urban segmentation on ACDC, IDD and WD2, respectively) we can see that our methods
can successfully isolate the  main 
reasons for the clusters to be \textit{hard}, for example
difficult visual conditions  
(\textit{rainy/foggy weather} or \textit{image taken at night}) or the presence of elements unseen during training  such as \textit{tunnels, mud} or \textit{light reflections}). Indeed, such elements can represent
real challenges for a model trained on Cityscapes, as observed also, \eg, in~\cite{deJorgeCVPR23ReliabilityInSemanticSegmentation}. 

With our methods, the user can easily asses the potential failures at a glance, instead of inspecting  the hundreds of images contained in the above clusters. They could easily rank the images in the cluster for each selected sentence for the visual inspection based on the corresponding Open-CLIP similarity scores.
Such information can be further used for different down-stream tasks, for example helping the user to select which new samples to annotate~\cite{settles2009active,ren2021survey}.

In~\cref{fig:all_res}, we provide numerical results 
for the different methods, tasks and datasets, using
the metrics introduced in~\cref{sec:LBEquality} and the default setting and hyper-parameters detailed above. 
First,  we observe that \tops{} is the best performing 
method in terms of ACR scores. This is not surprising, since sentences  
closest to the prototype have obviously the highest coverage. Yet, these sentences are not necessarily pointing to error explanations---indeed, see the low AHR scores in some cases, with IDD being the most dramatic one (right-most column, third panel).
Concerning the metrics related to the failure explanations (AHR/TPR/JI),
in both \npp{} scenarios, where we have ground truth relevance scores, and on WD2 \fpdiff performs best or close to it.
It performs worse on ACDC and IDD: there is no clear winner among the proposed methods in these two datasets, since the higher performance of \pdiff{} and \setdiff{} in terms of TPR and JI comes at the price of a lower ACR.

In the following, we provide more in-depth analysis focusing on the different datasets.

\begin{figure*}[!ttt]
    \centering
   \includegraphics[width=0.5\linewidth]{legend_ADSeg_TPR_bar.pdf}\\
\includegraphics[width=0.24\linewidth]{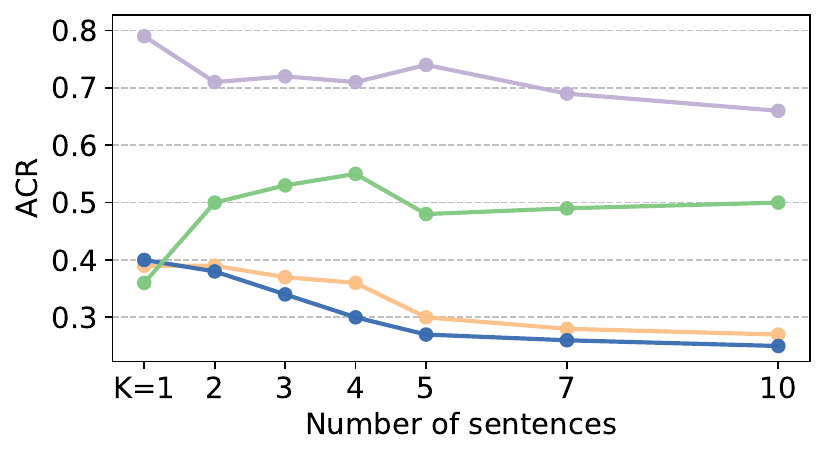}
\includegraphics[width=0.24\linewidth]{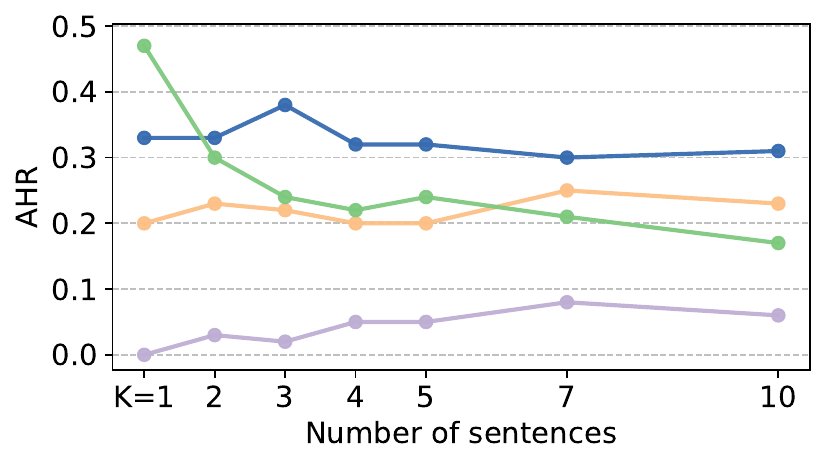}
\includegraphics[width=0.24\linewidth]{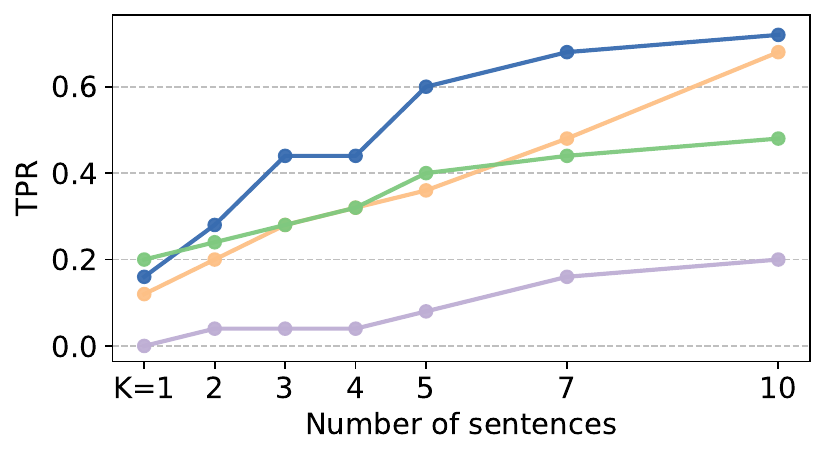}
\includegraphics[width=0.24\linewidth]{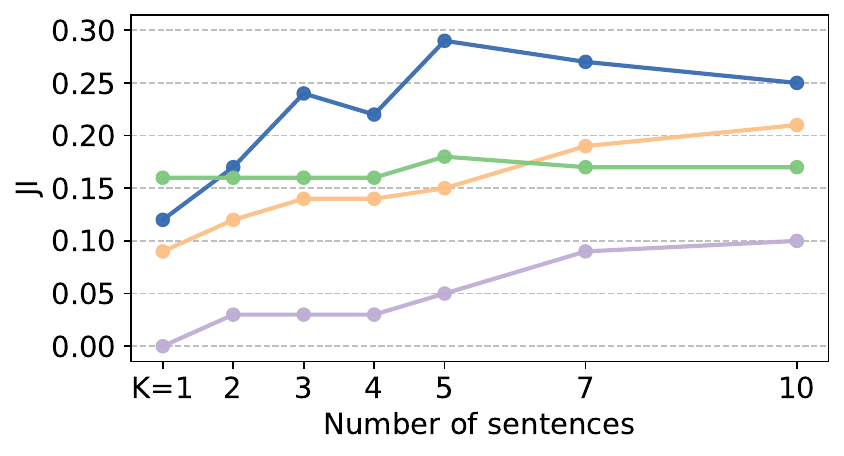}\\
\vspace{0.5cm} 
\includegraphics[width=0.24\linewidth]{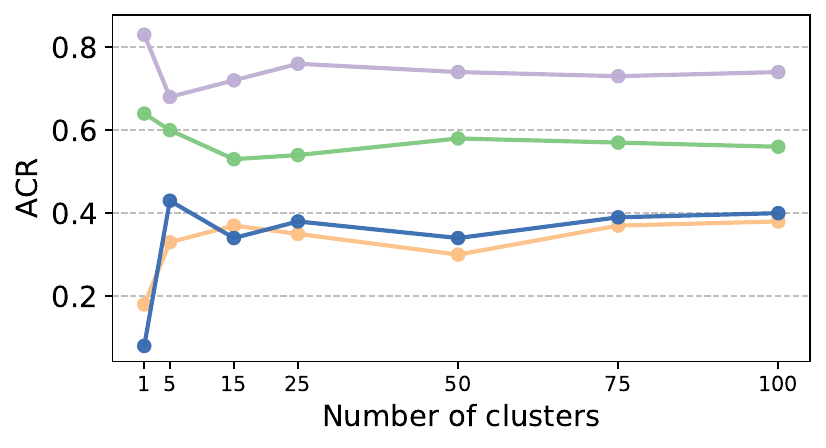}
\includegraphics[width=0.24\linewidth]{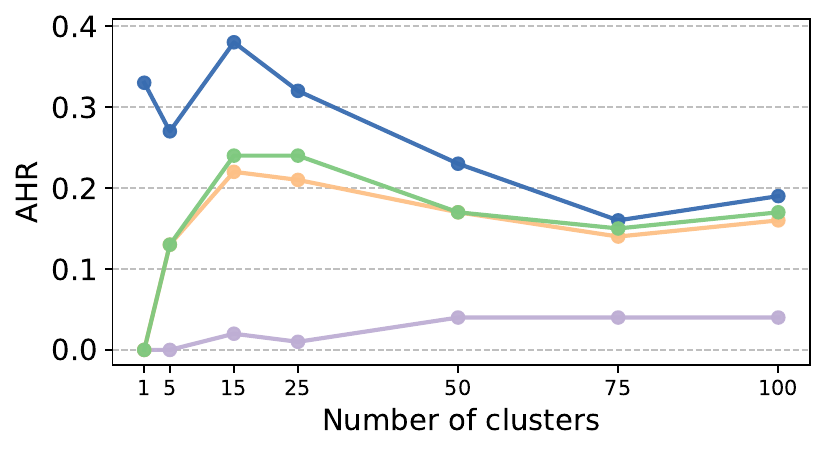}
\includegraphics[width=0.24\linewidth]{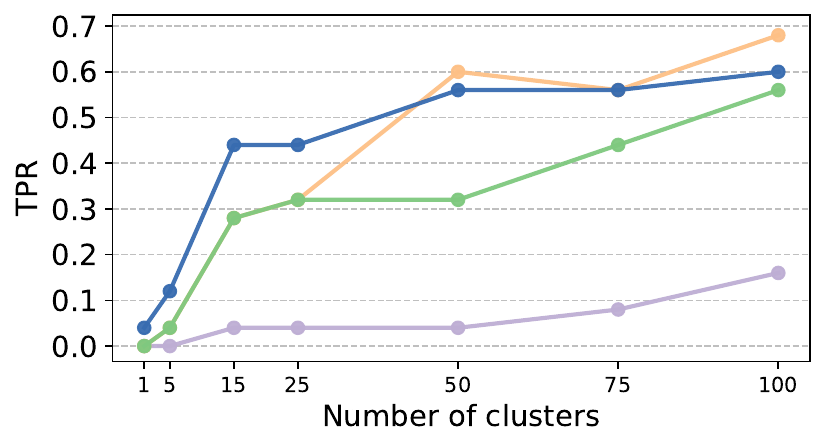}
\includegraphics[width=0.24\linewidth]{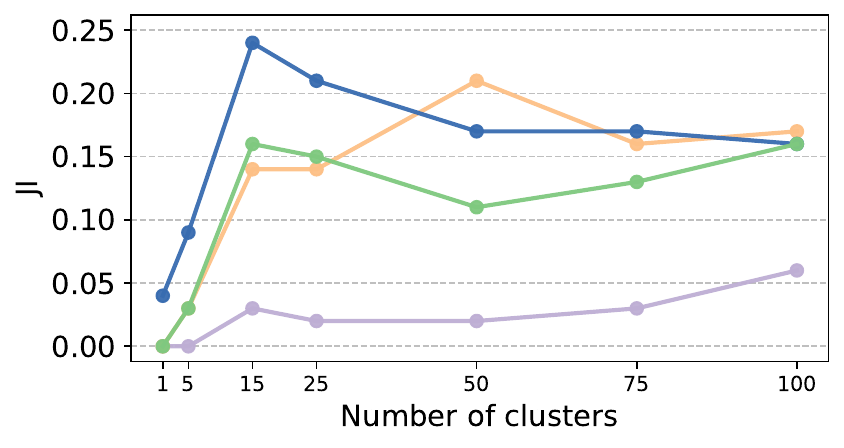}\\
    \caption{
    From left to right, we show how the ACR, AHR, TPR and JI metrics vary on the IDD experiment, by \textbf{varying the number of sentences selected for each cluster $K$} (top) and \textbf{varying the number of clusters $C$} (bottom). Note that $C=1$ corresponds to not performing any clustering and, hence, analyzing the whole hard set \vs the whole easy set. The very poor numbers obtained with $C=1$ emphasizes the importance of clustering. 
   }
   \vspace{0.5cm}
    \label{fig:IDD}
\end{figure*}

\subsubsection{ACDC}

\noindent To recap, we test models trained on Cityscapes (daylight and clear weather) on ACDC, which contains samples recorded under various weather and daylight conditions.  Results in \cref{fig:all_res} show that 
the best AHR performance is obtained with \tops{}, but also that  
this method is the worst performing one in terms of TPR and JI. For what concerns the latter metrics, 
\pdiff{}  shines as the best performing method.
To shed light on these results, we  complement them with 
\cref{tab:TPFP_UrbanSegmSmall} (and with the more detailed \cref{tab:TPFP_UrbanSegm})
showing  sentences from the sentence set $\calS$ that are either 
in $\calS^*_{\beta}$ or in any $\calR_\calS$, namely the ground truth
set of hard sentences and the ones retrieved by our methods, respectively.

From these results we can make the following observations.
\textit{i)} The main advantage of \tops{} comes from 
retrieving the sentences related to images taken at night or in the evening. \textit{ii)} The other methods 
fail in retrieving
these sentences because the closest easy clusters also contain night or evening images. \textit{iii)} On the other hand, these methods often select sentences referring to ``\textit{fog}'', or ``\textit{rain}'' 
(see \egs \cref{fig:qualityACDC}) that are indeed valid  characteristics of the images in the cluster,  but yet these conditions do not affect sufficiently the model to have a hardness score that satisfies the condition
$\omega_{\theta}^{s_n} <  \omega_{\theta}^\text{avg} - \beta$.

\myparagraph{Varying the easy/hard splitting strategy}
In order to evaluate  
the effectiveness of the entropy as a metric to split datasets into easy and hard partitions,
in  \cref{fig:ent_vs_pa_miou} we 
compare the results obtained by splitting the ACDC set with entropy \vs using metrics that
rely on annotated samples, namely the
pixel accuracy (average of correctly predicted pixels) or mIoU. 

We can observe that while using the GT mIoU improved significantly the 
ACR and AHR scores (averaged over all clusters), globally the TPR and JI changed only slightly. 
This suggests that, in general, the sentences 
retained by \pdiff, \fpdiff and \setdiff{} have better coverage 
in the cluster (higher ACR) and retain more often a  
hardness score that satisfies the required constraint. 
The TPR and JI scores are comparable across splitting strategy.
By analyzing the  sentences retrieved in each setting,
we discovered that
with GT mIoU the contrastive methods are able to recover 
sentences such as ``\textit{taken at night''}, ``\textit{taken in the evening}'' and ``\textit{taken at dusk}'',
but fail in retrieving
sentences such as ``\textit{with road barrier''/``with rail track on the road}'' and ``\textit{with motion blur``/''with underexposure}'', namely sentences that were retrieved by using the entropy.

Finally, sentences referring to \textit{fog} and \textit{rain}, having scores slightly below the required condition
($\omega_{\theta}^{s_n} <  \omega_{\theta}^\text{avg} - \beta$),  are  present in $\calR_\calS$  
for all splitting strategies and most methods. One could interpret those as less severe false positives.

\subsubsection{IDD}

\noindent In the case of IDD, the best AHR/TPR/JI results in \cref{fig:all_res} are obtained with \setdiff. Yet, the lower  ACR suggests that the sentences retained by this method have lower coverage in the cluster than the ones retained by the other methods.
IDD is a large dataset with high content variation, hence, posing more challenges.
Indeed, splitting into 15 clusters makes the content in each cluster very 
heterogeneous (see \cref{fig:qualityIDD}) 
and therefore the selection of the three most \textit{representative sentences} is  rather difficult.
 
To tackle this complexity, we carry out two additional IDD experiments, varying \textit{i)} the number of sentences selected  and \textit{ii)} the number of clusters.

\begin{figure*}[ttt]
{\footnotesize
    \centering
    {{\includegraphics[width=0.9\linewidth]{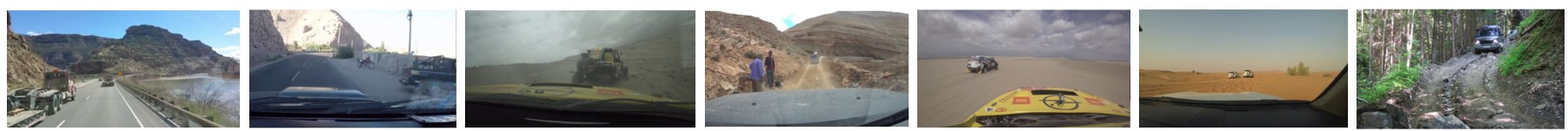} }}\\
     \begin{tabular}{ccc}
    \textbf{GT-based} & \textbf{User-defined}    & \textbf{LLM-based}  \\
 ``\textit{wheeled vehicle}'' \,\, \,\,& \,\,\,\, ``\textit{mud on the road}''  \,\, \,\,& \,\,\,\, ``\textit{off-road vehicle}'' 
  \\
 ``\textit{trailer}''  \,\, \,\,& \,\,\,\,``\textit{rocks on the road}'' 
  \,\, \,\,& \,\,\,\,``\textit{taken on a mountainous road}'' \\
 ``\textit{road}''    \,\, \,\,& \,\,\,\,``\textit{jeep on the road
}'' \,\, \,\,& \,\,\,\,``\textit{taken on a rocky road}'' 
    \end{tabular} \\
        {{\includegraphics[width=0.9\linewidth]{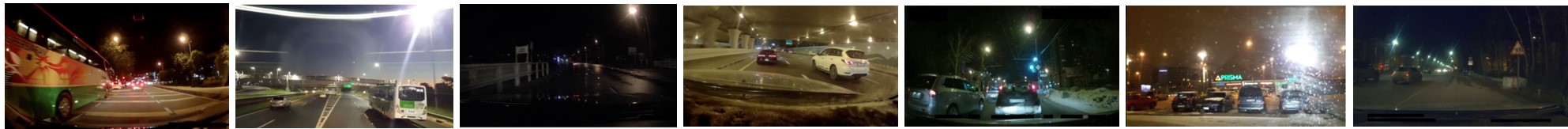} }}\\
     \begin{tabular}{ccc}
    \textbf{GT-based} & \textbf{User-defined}    & \textbf{LLM-based}  \\
 ``\textit{streetlight}'' \,\, \,\,& \,\,\,\, ``\textit{taken at night}''  \,\, \,\,& \,\,\,\, ``\textit{taken in a vibrant night}'' 
  \\
 ``\textit{traffic light}''  \,\, \,\,& \,\,\,\,``\textit{taken at dusk}'' 
  \,\, \,\,& \,\,\,\,``\textit{taken in the late snowy night}'' \\
 ``\textit{dashcammount}''    \,\, \,\,& \,\,\,\,``\textit{jeep on the road
}'' \,\, \,\,& \,\,\,\,``\textit{taken in a gloomy night}'' 
    \end{tabular} \\
          {{\includegraphics[width=0.9\linewidth]{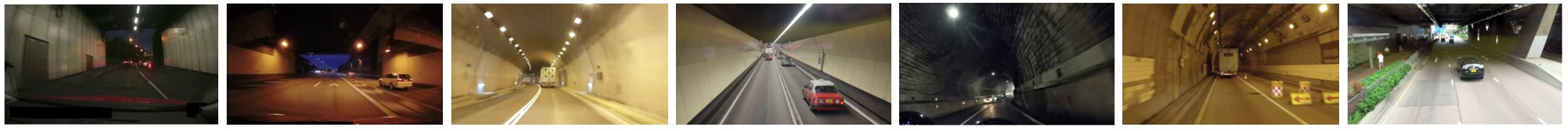} }}\\
     \begin{tabular}{ccc}
    \textbf{GT-based} & \textbf{User-defined}    & \textbf{LLM-based}  \\
 ``\textit{tunnel}'' \,\, \,\,& \,\,\,\, ``\textit{tunnel}''  \,\, \,\,& \,\,\,\, ``\textit{tunnel
}'' 
  \\
 ``\textit{streetlight}''  \,\, \,\,& \,\,\,\,``\textit{light reflection on the road}'' 
  \,\, \,\,& \,\,\,\,``\textit{underwater tunnel}'' \\
 ``\textit{dashcammount}''    \,\, \,\,& \,\,\,\,``\textit{motion blur}'' \,\, \,\,& \,\,\,\,``\textit{underpass}'' 
    \end{tabular} \\
    }
    \caption{
   Three out of 15 hard clusters from WD2  (unsupervised case)   explained by \fpdiff{} when we use different the sentence sets (GT based, User defined or LLM generated). }
   \vspace{0.5cm}
\label{fig:qualitativeWD2Sent}
\end{figure*}

\begin{figure*}[!ttt]
    \centering
    \includegraphics[width=0.5\linewidth]{legend_ADSeg_TPR_bar.pdf}\\
\includegraphics[width=0.24\linewidth]{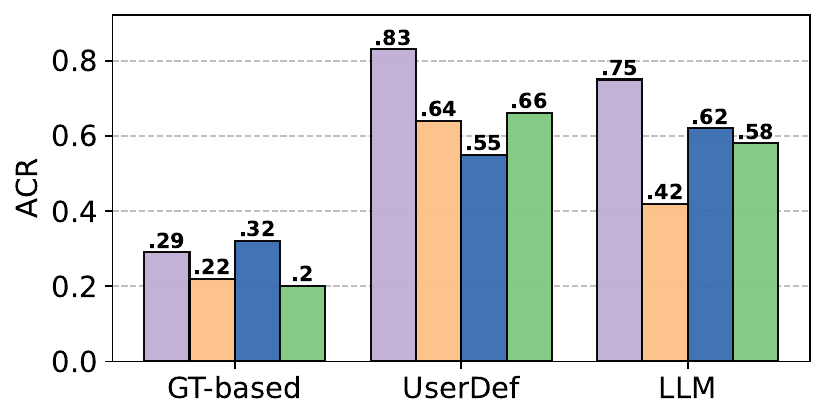}
\includegraphics[width=0.24\linewidth]{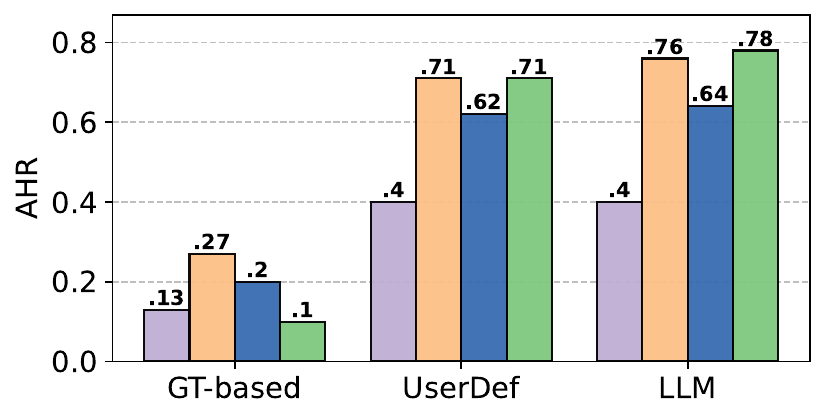}
\includegraphics[width=0.24\linewidth]{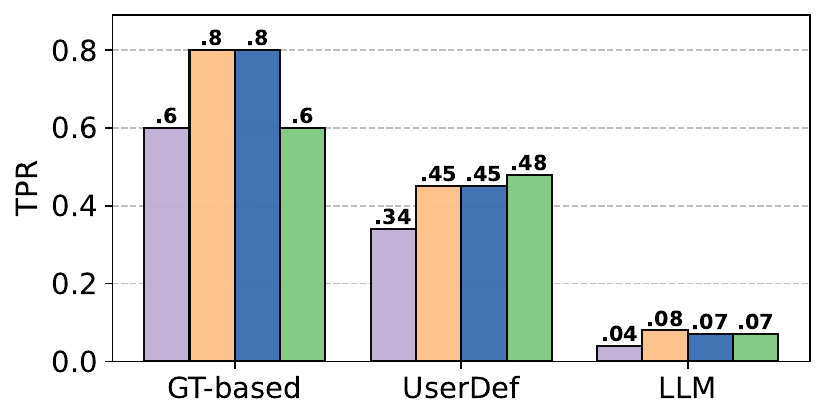}
\includegraphics[width=0.24\linewidth]{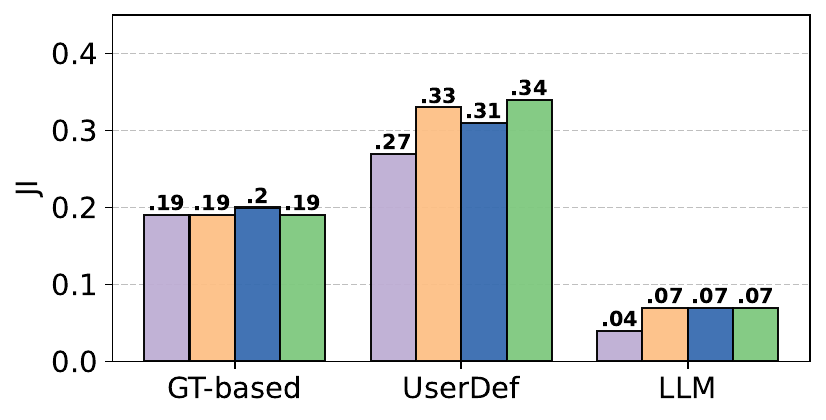}\\
    \caption{
    \textbf{
     Varying the sentence set.}  Results obtained on  WD2  when $\calS$ is the 
     GT-based, User-defined and the LLM-generated sentence set (see  \cref{sec:construcS}).}
    \label{fig:WD2Sent}
\end{figure*}

\myparagraph{Varying the number of retained sentences ($K$)}  We show  in \cref{fig:IDD} (top) results for IDD as we vary the number of retained sentences per cluster. By increasing the number of sentences retained, we observe that the ACR increases for 
\fpdiff{},  but decreases for the other methods; on the other hand, the ACR is stable starting from  $K=5$. The AHR  is generally stable as we vary $K$,
except for \fpdiff, where we obtain the best value with $K=1$. 
Both TPR and JI benefit from the increase of the number of sentences (higher $K$).
Overall,
\setdiff{} outperforms the other methods in terms of AHR, TPR and JI.
This result, combined with the low ACR, suggests that \setdiff{}
is more capable of detecting rarer failure reasons, for which there
are less support example images in the cluster/dataset.

\myparagraph{Varying the number of clusters ($C$)}
In  \cref{fig:IDD} (bottom) we present results obtained on IDD when we vary 
the number of clusters ($|\calC^h|=|\calC^e|=C$).
By increasing
the number of clusters, we observe an increase of the TPR (as expected) for all methods---with \pdiff{} being the one that benefits the most. Concerning the other metrics, the default value $C=15$  is a reasonable trade off.
These experiments suggest that IDD is overall a very challenging set, with some of the failure 
causes under-represented---especially the ones related to the weather 
conditions (see sentences in \cref{tab:TPFP_UrbanSegmSmall} and \cref{tab:TPFP_UrbanSegm}),  or not clustered together according to these characteristics (\textit{cf.}  \cref{fig:qualityIDD}).
Furthermore, note that as the filtering in \fpdiff relies on the \tops{} ranking, the very poor performance of the latter on this particular set (IDD) affects negatively the performance of the former, making \pdiff{} 
perform better than \fpdiff  when we consider more clusters or 
retain more sentences. Finally, the low performance obtained with $C=1$---corresponding to running the methods without the clustering step---shows 
that describing the full hard set by contrasting with the full easy set is a sub-optimal solution.

\subsubsection{WD2}

We analyse the performance of the different methods on WD2, varying the sentence set $\calS$. Indeed, for this dataset we have both User-defined (the same used throughout the experimental validation), a GT-based sentence sets, as well as a significantly larger LLM-based sentence set (see  \cref{sec:sentences}).  We show qualitative  comparison in \cref{fig:qualitativeWD2Sent} and numerical 
results in \cref{fig:WD2Sent}.

\myparagraph{Sensitivity to the choice of the sentence set}
First, we notice that both the User-defined and the LLM-generated sentence sets yield similar ACR and AHR scores. Yet, for the latter we have many sentences that are semantically related and carry similar hardness scores, for example \textit{``taken at night''/``mesmerizing moonlight''/``spellbound starlight''/``enigmatic night''/``miraculous midnight''}.
Since we limit the number of sentences each method can retrieve, in this case we 
cannot retrieve all of them with a small $K$ and, in turn,
this will result into significantly lower TPR and JI scores.  

When we use the GT-based sentence set with $o=0.2$, $\calS^{*}_{\beta}$ contains only five sentences  that are  ``\textit{showing a tunnel}'',  ``\textit{showing a bridge}'',  ``\textit{showing traffic light}'', ``\textit{showing a bus}'' and ``\textit{showing snow}''. 
Here, all methods introduce many false positives (explaining the low JI score). \pdiff{} and \setdiff{} are able to retain  four out of five sentences (except ``\textit{showing a bus}'') and  \tops{} and \fpdiff{} three out of five (missing also ``\textit{showing a bridge}''). The missing sentences had not enough image coverage in the clusters. The AHR and ACR scores are also low, due to the fact that it is not easy to find shared content for all clusters (see the low ACR even for \tops). Finally, note that the GT-based  sentence set 
mainly describes the presence of the classes (``\textit{Image showing} <class>''), therefore, it is not sufficient to describe the hard clusters to the desired level of details. This highlights the  importance of properly designing $\calS$, as well
as the importance of using additional textual metadata to analyze the performance of our models.

\begin{figure*}[!ttt]
    \centering
    \includegraphics[width=0.5\linewidth]{legend_ADSeg_TPR_bar.pdf}\\
\includegraphics[width=0.28\linewidth]{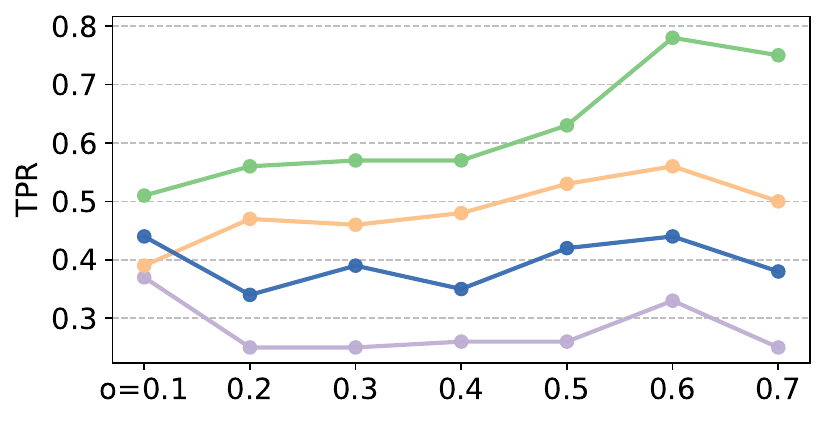}
\includegraphics[width=0.28\linewidth]{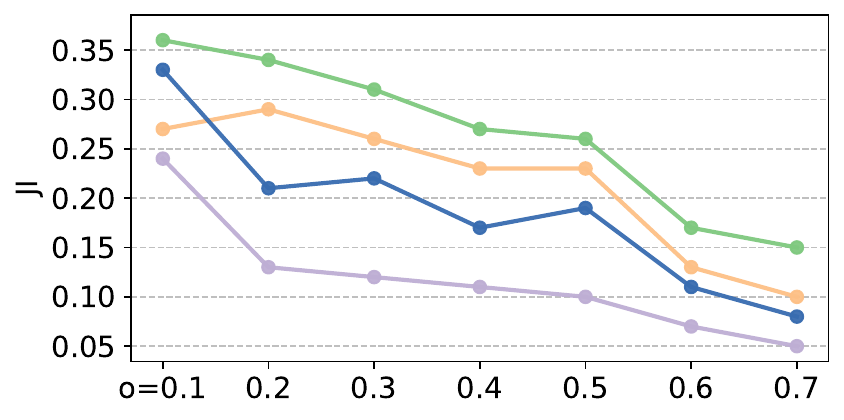}
\includegraphics[width=0.28\linewidth]{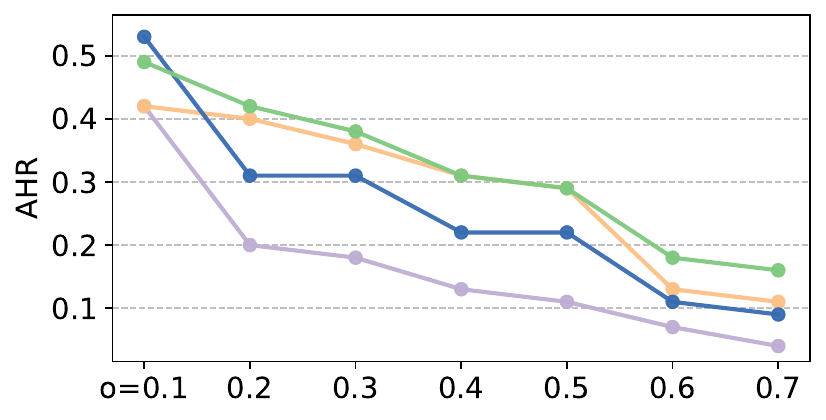}\\
    \caption{
     \textbf{
     Varying $\beta$ (by varying $o$).}
     Results obtained  on \nppm (unsupervised).  The value $\beta$ defines the ground truth set $\calS^{*}$, so its value impacts only the AHR/TPR/JI metrics 
     (not the output of the methods or ACR).  On x-axis: we plot $o$, where $\beta=o\cdot\omega_{\theta}^{\text{std}}$ with $\omega_{\theta}^{\text{std}}$ being the standard deviation of $\omega_{\theta}$ (entropy) computed over $\calX$.}
    \label{fig:BetaStudy}
\end{figure*}

\begin{figure*}[ttt]
{\footnotesize
    \centering
     {{\includegraphics[width=.8\linewidth]{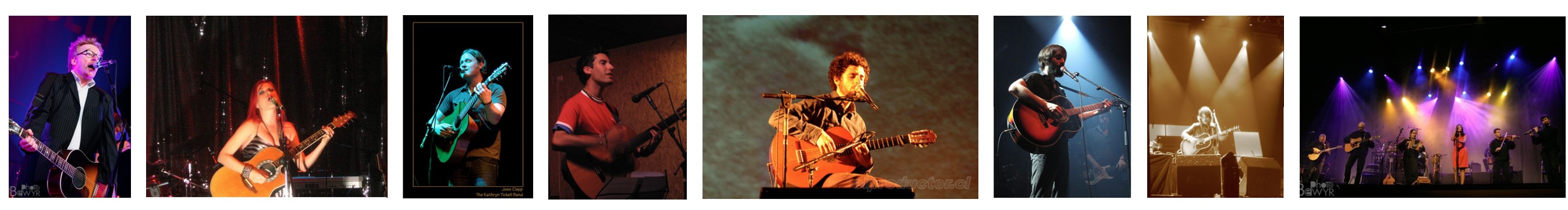} }}\\
     \begin{tabular}{cccc}
    \tops & \setdiff  & \pdiff  & \fpdiff \\
 ``\textit{stage indoor}''  \,\, \,\, & \,\, \,\,  ``\textit{basement}''  \,\, \,\, & \,\, \,\, ``\textit{with motion blur}''  \,\, \,\, & \,\, \,\, ``\textit{blurry image}''\\
``\textit{arena performance}''  \,\, \,\, & \,\, \,\, ``\textit{stage indoor}''  \,\, \,\, & \,\, \,\, ``\textit{blurry image}''    \,\, \,\, & \,\, \,\, ``\textit{stage outdoor}''\\
``\textit{taken in a basement}''    \,\, \,\, & \,\, \,\, ``\textit{thumbnail image}''  \,\, & \,\,``\textit{stage outdoor}'' \,\, & \,\,``\textit{stage indoor}''
    \end{tabular} \\
{{\includegraphics[width=.8\linewidth]{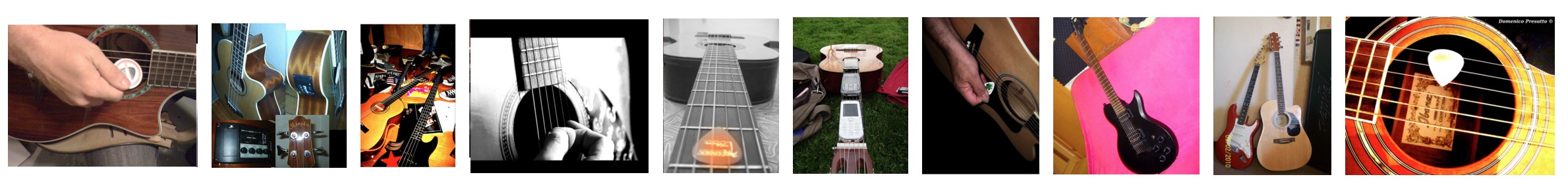} }}\\
  \begin{tabular}{cccc}
 \tops & \setdiff  & \pdiff  & \fpdiff \\
 ``\textit{acoustic guitar}''  \,\, \,\, & \,\, \,\,  ``\textit{taken in recreation room}''  \,\, \,\, & \,\, \,\, ``\textit{in dorm room}''  \,\, \,\, & \,\, \,\, ``\textit{taken in motel}''\\
``\textit{electric guitar}''  \,\, \,\, & \,\, \,\, ``\textit{taken in motel}''  \,\, \,\, & \,\, \,\, ``\textit{market indoor}''    \,\, \,\, & \,\, \,\, ``\textit{recreation room}''\\
``\textit{ banjo}''    \,\, \,\, & \,\, \,\, ``\textit{taken in shed}''  \,\, & \,\,``\textit{taken in storage room}'' \,\, & \,\,``\textit{music studio}''
    \end{tabular} \\
 {{\includegraphics[width=.8\linewidth]{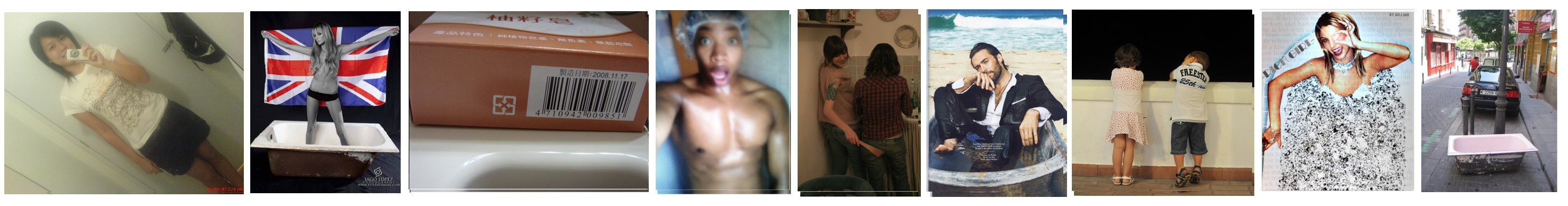} }}\\
\begin{tabular}{cccc}
 \tops & \setdiff  & \pdiff  & \fpdiff \\
 ``\textit{taken in dressing room}''  \,\, \,\, & \,\, \,\,  ``\textit{with reflections}''  \,\, \,\, & \,\, \,\, ``\textit{ taken in promenade}''  \,\, \,\, & \,\, \,\, ``\textit{taken in promenade}''\\
``\textit{taken in house}''  \,\, \,\, & \,\, \,\, ``\textit{with overexposure}''  \,\, \,\, & \,\, \,\, ``\textit{ taken in street}''    \,\, \,\, & \,\, \,\, ``\textit{ taken in street}''\\
``\textit{blurry image}''    \,\, \,\, & \,\, \,\, ``\textit{screen shot}''  \,\, & \,\,``\textit{stage outdoor}'' \,\, & \,\,``\textit{stage outdoor}''
 \end{tabular} \\
    }
    \caption{Analyzing EfficientNet performance on the ImageNet 1K. The first two
    rows are from the ``\textit{acoustic guitar}'' class, 
   and the third is from the ``\textit{bath tube}'' class. }
\label{fig:qualityEffNet}
\end{figure*}

\subsubsection{\nppm}

We provide in \cref{{fig:qualitativeNPPmain}} qualitative examples for \nppm, in the unsupervised case (where we use the prediction entropy to split the data). 
We observe that our methods can
capture the \textit{context-class} associations that were rarely seen in the training set\footnote{See in \cref{tab:NPPstat} the   distribution of the images in the validation set  over the different classes-context pairs. Note that the validation set follows the training distribution.}  and hence the model struggles with their classification. Indeed 
we find ``\textit{birds}'' or ``\textit{plants}'' in ``\textit{water}'' (top row), ``\textit{landways}'' in the ``\textit{rocks}'' (middle row) and ``\textit{landways}'' in `\textit{dim lighting}'' (bottom row). 

Next, we provide a sensitivity study on this dataset 
when we vary the hardness threshold value  $\beta$.

\myparagraph{Varying $\beta$ (by varying $o$)}
Recall that $\beta$ is the hyper-parameter used to select the ground-truth sentence set $\calS^{*}$, where higher $\beta$ values  imply retaining
only the sentences with higher hardness scores as valid failure descriptions, while lower values imply that also sentences with a  less strict  hardness score are also considered as valid explanations. 
In Fig.~\ref{fig:BetaStudy} we show the impact of varying $\beta$   on the AHR, TPR, and JI metrics.  We carried out this experiment on the spurious classification task with \nppm, considering the unsupervised case where we have GT image-sentence relevance scores (top row).  As the number of sentences retained by the method is fixed ($K=3$),  varying $\beta$ only affects
$\calS^{*}$ and not $\calR_\calS$.
Therefore, the value of ACR (that does not depend on $\beta$) is the same for each  $o$ value.  Intuitively, with this study we  assess how much each method focuses on the hardest sentences. Indeed, increasing the value of $o$ (and hence increasing $\beta$) corresponds to accepting fewer and fewer sentences as valid explanations, namely the ones with the highest hardness score (lowest average accuracy).  We observe that  the \fpdiff method, while experiencing a drop in performance as we make the problem harder, holds a better performance than the others methods for all the values of $o$.

\subsubsection{ImageNet 1K}

Finally, we focus on ImageNet 1K, where we treat each class separately. 
We can make several observations. 
\textit{i)} When the error slices (hard clusters)
mainly contain false positives from  semantically similar classes
or  false negatives  with shared content, \tops will generally capture 
the shared characteristics---see examples in~\cref{fig:qualityEffNet} (top two rows), \cref{fig:ImageNetFN} and \cref{fig:ImageNetFP} in \cref{sec:qualitative}. 
\textit{ii)} In the  case of false positives with
semantically similar classes, \eg,      ``\textit{electric guitar}'' and ``\textit{banjo}'' in place of the ``\textit{acoustic guitar}'' class (see first row in \cref{fig:qualityEffNet}), 
the main difficulty for the other methods is that the same sentences are also highly ranked for the  easy clusters, making it difficult for them to correctly identify the desired explanations---namely, contrasting the explanation between the easy and the hard clusters. This limitation is inherited from CLIP, which has difficulties to distinguish fine-grain classes.
\textit{iii)} When  the false negatives share some semantic similarities and there is a candidate sentence that well describes this,  all methods are able to find it. For example, see in \cref{fig:qualityEffNet} (second row) the difficulty of the model to label the ``\textit{acoustic guitar}'' when the photo is showing the person holding the guitar on a ``\textit{stage}''. Note  that for these images \pdiff and \fpdiff 
also identified  the ``\textit{blur}'' as potential challenge (or ``\textit{low contrast}''  and  ``\textit{dim weather}'' condition, in \cref{fig:ImageNetFN}).
\textit{iv)} Finally, we have clusters where even for a human is difficult to describe what is common in the images beyond the object class. 
When there is a mix between false positives and false negatives, all methods including \tops have difficulties to identify the failure reason. In these cases, the selected sentences are relevant to some of the images in the cluster but they do not point to
the reason for the failure (see \cref{fig:qualityEffNet} last row).

%% file: TPFPUrbanSegmSmall.tex
\begin{table*}[ttt]
    \footnotesize
    \centering
    \caption{
            \textbf{Examples of sentences from the User-defined sentence set $\calS$}. For each sentence we show if it belongs to   $\calS^*_{\beta}$ and to $\calR_\calS$ for \tops (\textbf{TS}), \pdiff (\textbf{PD}), \setdiff (\textbf{SD}) and  \fpdiff (\textbf{FP}) for the  datasets WD2, IDD and ACDC.  
            ``\gcheckmark'' means $s_n \in \calS^*_{\beta}$ and  ``\redx'' means $s_n \notin \calS^*_{\beta}$---namely,  
            the hardness score of $s_n$  is below the required level.  For the methods, ``\gplus''/``\rplus'' indicate that the sentence is in $\calR_\calS$, where ``\gplus''  means  true positive (\ie, the sentence is also in $\calS^*_{\beta}$) while  ``\rplus''  means  false positive. An empty space indicates that the sentence is not in the corresponding $\calR_\calS$.
    }
    \begin{tabular}{l|ccccc|ccccc|ccccc}
        \toprule
         & \multicolumn{5}{c}{WD2} &  \multicolumn{5}{c}{IDD} 
          & \multicolumn{5}{c}{ACDC}  \\
          \textbf{Sentences} (\textit{An image ...}) & $\calS^*_{\beta}$ & \textbf{TS} & \textbf{PD} & \textbf{SD} & \textbf{FP} & 
          $\calS^*_{\beta}$ & \textbf{TS} & \textbf{PD} & \textbf{SD} & \textbf{FP}  &
          $\calS^*_{\beta}$ & \textbf{TS} & \textbf{PD} & \textbf{SD} & \textbf{FP} \\
        \midrule
         ``taken at night'' &  \gcheckmark  &  \gplus & \gplus & \gplus  & \gplus & \gcheckmark & & \gplus & \gplus & \gplus &  \gcheckmark &  \gplus & & & \\
        ``taken in the evening'' &  \gcheckmark   &   & \gplus & \gplus  & \gplus &  \gcheckmark   &   &   \gplus &  & \gplus &  \gcheckmark &  \gplus & & &  \\
        ``taken at dusk'' & \gcheckmark   &  & \gplus & \gplus & \gplus &
        \gcheckmark   &  & \gplus & \gplus & \gplus & \gcheckmark & & & &   \\
   ``taken in a foggy weather''  & \gcheckmark &  &   & \gplus  &   & \gcheckmark & & & & &  \redx & \rplus  & \rplus &  \rplus & \rplus
        \\
        ``taken in a rainy weather'' &   \gcheckmark & \gplus & \gplus &   & \gplus & \gcheckmark & & & & &  \redx & \rplus  & \rplus &   & \rplus\\
            ``taken in a snowy weather''  & \gcheckmark &  & \gplus & \gplus & \gplus & \redx & & & & & \redx &  \rplus  & & & \\
             ``taken in a dull weather''  & \gcheckmark  & \gplus &  \gplus & \gplus & \gplus & \redx & & & & &  \redx & \rplus  & \rplus &  \rplus & \rplus\\
      ``with reflection on the road'' &   \gcheckmark & \gplus &  \gplus & \gplus & \gplus &   \gcheckmark & & &   \gplus & & \gcheckmark & & & & \gplus \\
        ``with shadows on the road'' & \redx &  &  &  & \rplus   & \redx & & & \rplus &  & \gcheckmark & & & \gplus & \\
         ``with water on the road'' &  \gcheckmark &   \gplus &  & & & \redx & & & & & \redx & & & \rplus & \rplus\\
          ``with mud on the road'' & \redx &  &  & \rplus &  \rplus & \redx & &  \rplus &  \rplus & & \redx & & & & \\
   ``with obstacle on the road'' &  \gcheckmark  & \gplus   &  &  & \gplus   & \redx &  \rplus & & & & \redx & & & &\\
    ``with road barrier on the road'' & \redx &  &  & \rplus & & \redx & &  & \rplus &  \rplus  &  \gcheckmark  & \gplus   &  \gplus  & \gplus  & \gplus\\
     ``with rail track on the road'' & \redx &  &   &  &   & \gcheckmark & & &  \gplus &  \gplus & \gcheckmark & \gplus & \gplus & \gplus &\gplus \\
      ``with rocks on the road'' & \redx &  &  & \rplus & \rplus & \redx &  &  & &  & \redx &    & &  & \\
        ``showing a highway scene'' & \redx &  &  & \rplus &  & \redx & & & \rplus & \rplus & \redx & \rplus & \rplus & \rplus & \rplus \\
         ``showing an industrial scene'' &  \gcheckmark &  &  &  &  & \gcheckmark &  &  &  & &  \redx &    & &  &\\
         ``showing construction site'' & \redx & & & & & \gcheckmark  & & \gplus &  \gplus & \gplus &  \redx & &  \rplus &  \rplus\\
        ``showing sub-urban scene'' &  \redx  & \rplus &  & \rplus &  &  \redx & \rplus &    &  & \rplus & \gcheckmark & & &  \\
        ``with rickshaw on the road'' &  \redx  & \rplus & \rplus &  \rplus  & \rplus & \redx & \rplus & \rplus &  & \rplus  & \gcheckmark & & \gplus & &  \\
        ``with motorbike on the road'' & \redx &  &  \rplus &  & \rplus & \redx & & & &  \rplus & \gcheckmark & & & & \gplus \\
 ``with vehicle on the road'' & \redx & \rplus &  &  &  & \redx & \rplus &  &  & &  \redx & \rplus &  \rplus  & \rplus & \rplus\\
        ``with tram on the road'' & \redx & & & &  & \redx &  &  & \rplus  &  & \redx & \rplus & \rplus & & \rplus \\
        ``with jeep on the road'' & \redx & \rplus &  & \rplus &  \rplus & \redx & & & & & \redx & & & &\\
        ``with animal on the road'' & \redx &  &   & \rplus &  \rplus  & \gcheckmark & & &  \gplus & \gplus & \redx & & & &\\
          ``with people on the road'' & \redx & \rplus  &  & & \rplus & \redx & \rplus  &  & & \rplus  & \redx & & \rplus & &\\
          ``with crowded foreground'' & \redx & & & & & \redx & & \rplus& \rplus & \rplus & \redx & & & &\\
          ``with motion blur'' &  \gcheckmark &  \gplus &  \gplus & & \gplus & \gcheckmark & & & \gplus & \gplus & \gcheckmark & & \gplus & &  \\
         ``with underexposure''  & \redx  &  &  & \rplus & \rplus  & \redx & & & & & \gcheckmark & & & \gplus  &  \\
        ``with low contrast'' & \gcheckmark &  &  &  & & \redx & & & \rplus & \rplus  &  \redx & & & \rplus &\\
           ``of a tunnel'' &  \gcheckmark & \gplus & \gplus & \gplus & \gplus &  \gcheckmark & &  \gplus & & &  \gcheckmark & \gplus & \gplus & \gplus & \gplus \\
          ``of fences'' & \redx & & & &  & \redx & & \rplus  & &  &  \gcheckmark & & \gplus & & \\
           ``of guard-rail'' & \redx & & & &  & \redx & &   & &  &  \redx & & \rplus & \rplus & \rplus \\
         ``of a traffic jam'' &  \gcheckmark &  &  &  &  & \gcheckmark & \gplus & \gplus & & \gplus & \redx & &  & & \\
          ``of traffic lights'' & \gcheckmark &  &  &  & &  \gcheckmark &  &  & &  & \redx & & \rplus & &\\
        ``of a parking'' & \redx &  &   \rplus &  &  & \gcheckmark & & \gplus & \gplus & \gplus & \redx & &  \rplus & &\\
        \bottomrule
    \end{tabular}
    \label{tab:TPFP_UrbanSegmSmall}
\end{table*}

%% file: 7_conclusion.tex
\section{Concluding remarks}
\label{sec:ccl}

Assessing model performance, and in particular failure modes, for arbitrary task in arbitrary environments
should be \textit{easy},   \textit{efficient}
and \textit{interpretable}.  While the gold
standard for performance evaluation is using a human-annotated
test set, this process is costly and cannot scale
for computer vision models to be deployed in many and
diverse scenarios. Moreover, quantitative evaluation
measures such as dataset-level accuracy values do not
fully reflect the details of the model performance. 
It is important to know what lies beyond performance values and if
the model is failing on particularly critical images.

In this work,
we posit that it is of the utmost importance being able to assess model performance on non-annotated samples, and to move beyond hard-to-interpret numbers. 
We address this by first formulating the  \task{} problem  and
designing a family of \textit{task-agnostic} approaches
that do not require  error specifications 
or user-provided annotations. We complement these 
two  contributions with different metrics to benchmark methods for \task{}
and rely on those to carry out in-depth analyses. 
We show that we can retrieve relevant sentences pointing to important errors, \eg, related to environments beyond our model’s comfort zone.

In the future, we plan to  overcome the need to define the sentence set $\calS$ a priori. Indeed, while this design choice makes the evaluation more tractable and allows deploying our methods off the shelf, the set might omit certain unexpected failure reasons.
A  possible solution 
could be to  build the sentence set $\calS$ directly from the target set $\calX$, \eg, by running a captioner
to describe all its images and 
merge these descriptions into the sentence set.
Yet, in addition to being prohibitively expensive, a solution like this needs to be iterated for every new target set before running any method.
Furthermore, in the case of a large image set, it would also require further steps to remove redundancy and to regroup similar sentences. 

%% file: 8_Appendix.tex
\section{Datasets and Tasks}
\label{sec:datasets}

In the following, we detail the
three main tasks we tackle in this paper: 
semantic segmentation of urban scenes, classification in the presence of spuriously correlated data, and ImageNet-1K classification.

\myparagraph{Urban scene segmentation}
We consider a  ConvNeXt~\cite{LiuCVPR22AConvNet} segmentation model
trained on Cityscapes~\cite{CordtsCVPR16CityscapesDataset}, comprised of images from 50 European cities collected at daytime in clear weather conditions. 
We build our evaluation sets $\calX$ using three
datasets: 
\textit{i)} WildDash 2 (WD2)~\cite{ZendelECCV18WildDashCreatingHazardAwareBenchmarks}, which contains  challenging visual conditions such as motion blur, various road types, 
difficult
weather,  
\etc \ and \textit{ii)} India Driving Dataset (IDD)~\cite{VarmaWACV19IDDDatasetExploringADUnconstrainedEnvironments}, which contains
images from Hyderabad, Bangalore and their peripheries, \textit{iii)} ACDC~\cite{SakaridisICCV21ACDCAdverseConditionsDatasetSIS}
contains images recorded in adverse conditions, namely \textit{fog}, \textit{rain}, \textit{snow} and \textit{night} representing 
challenging domain shift for the Cityscapes model. For the main experiments we
use the user-defined
sentence set $\mathcal{S}$ with 148 sentences describing 
content  related to urban scenes, road conditions, and image quality (described in \cref{sec:sentences}). In  Sec. {\color{red} 6.1.3} 
we compare these results with the results obtained on WD2 when using both a  smaller GT-based sentence set derived from the metadata information available with the dataset and a larger  sentence set  of  1016 sentences generated  automatically by an LLM (both sets detailed in \cref{sec:sentences}).

\myparagraph{Classification with spurious correlations}
We use the \npp~\cite{ZhangCVPR23NICOPP} dataset, consisting of real-world images of concept classes (\textit{mammals}, \textit{birds}, \textit{plants}, \textit{airways}, \textit{landways} and \textit{waterways}) taken in six  contexts (\textit{dim lighting}, \textit{outdoor}, \textit{grass}, \textit{rock}, \textit{autumn}, \textit{water}). We follow the train, biased validation, and unbiased test   splits from \cite{YenamandraICCV23FACTS},
aimed at creating different levels of class-context correlations. We use
three partitionings \nppl, \nppm and \npph with increased levels
of correlation.\footnote{This means setting the correlation level to 75\%, 85\% and 95\% respectively  
between a class and its natural context in the training/validation set, as detailed in    \cite{YenamandraICCV23FACTS}.}
We train a ResNet-50 model on each training set, and use the unbiased test set for evaluation.

To evaluate error descriptions of the hard clusters,
we build a sentence set $\mathcal{S}$ with 130 sentences that include information related to the the six classes, the six contexts, various sub-classes and their combination. 
Note that in this case we have  GT relevance $\Gamma (\bfx_m,s_n)$  between sentences and images.

We consider two cases, an unsupervised and a supervised one. In the former case,  we consider the full set and split it with entropy values computed on class predictions. In the latter case, similarly to \cite{EyubogluICLR22DominoDiscoveringSystematicErrors,YenamandraICCV23FACTS}, we consider images from each class separately and use the class probability scores to split the data into easy and hard sets.

\myparagraph{ImageNet-1K  classification}
We consider three different architectures (ResNet-50~\cite{HeCVPR16DeepResidualLearning}, VIT-B-16~\cite{DosovitskiyICLR21ImageIsWorth16x16WordsViT}, and EfficientNet-B1~\cite{TanICML19EfficientNet}) trained on the ImageNet-1K~\cite{RussakovskyIJCV15Imagenet} training set and evaluate these models
on the ImageNet-1K validation set.  We build $\calS$ as the union of a set of sentences 
\{``\textit{An image of a} \texttt{<class>}''\} (for the 1K classes) with
a set of 
sentences corresponding to some place classes from Places365~\cite{ZhouPAMI18Places365}
in form of \{``\textit{An image taken at} \texttt{<location>}''\},
and a set of sentences describing image type (photo, drawing), image quality (blurry, noise, JPEG compression), weather conditions, \etc \ In total we have 1417 sentences (see details in \cref{sec:sentences}).
To assess the relevance between a sentence and an image, we use a  combination of ground truth  based information (for the first 1000 sentences corresponding to the 1000 class names) and a VQA model to assess the relevance for the other sentences (as described in \cref{sec:pseudoGT}).

With this dataset,  we focus on the top-1 classification  performance of different models and aim at analyzing failure cases for this task---namely, the reasons for not predicting the correct class on top. In ImageNet classification, several errors arise from confusions between semantically similar classes. For example, the model will miss-classify dog breeds, bird/fish species, similar objects, \etc  In such cases, knowing the correct class label can help to deduce the main reason from the retrieved description. Indeed, for example, if the selected sentence is  
\{``\textit{An image of a water snake}''\}  without the knowledge of  the correct class, it is difficult to interpret whether
the model predicts false positives (images of different type of snakes) or the model failed to recognize  the  water snake (false negative) for some reasons. 
Furthermore, while, class independent failure reasons 
such as those related to image capturing conditions (\eg blurred or low contrast images)
 might emerge when  processing the full dataset as a whole, errors related to the image content
is more difficult to interpret  if it is a failure reason 
without the knowledge of  the target  class (\eg finding that the image was taken at a \textit{beach} has different effect on the model for the \textit{swimming suit} class than for the \textit{cow} or \textit{ski suit} class). 
Finally,  even if we  use a very large number of clusters with the full set, the content of the nearest easy cluster  might differ in many aspects from the hard one, making rather it difficult to select in priority which is the one causing the model to fail.

For all these reasons we opted to follow the prior art and  apply \task{} methods by analyzing the data for each class independently.
Still, in contrast to prior art, where only false negative errors are considered (the images that belong to the given class but labeled with a wrong class label), we also 
include false positives. Hence, the easy set
$\calX^e$ contains the images where the class has been correctly predicted (images from the current class) while the hard set is a union of false positives and false negatives.

\input{npp_biased_stats}

\section{Comparison with prior art}
\label{sec:npp}

DOMINO~\cite{EyubogluICLR22DominoDiscoveringSystematicErrors} and FACTS~\cite{YenamandraICCV23FACTS} mainly focus on evaluating the error mode discovery~\cite{EyubogluICLR22DominoDiscoveringSystematicErrors,YenamandraICCV23FACTS}, assuming to know the set of error types, and limit the language description of these error 
modes to a qualitative role. In contrast, in
our problem formulation, we treat the language-based description of errors made by
computer vision models as the problem itself. For this reason, 
a direct comparison with prior art is not straightforward. 
In this section,  we make an attempt by simply comparing
our split-and-cluster partitioning to the more complex, learnt partitioning methods by~\cite{EyubogluICLR22DominoDiscoveringSystematicErrors,YenamandraICCV23FACTS}, following their evaluation protocol. 

We perform this analysis using the \nppl, \nppm and \npph 
sets from \cite{YenamandraICCV23FACTS}, corresponding to a 75\%, 85\% and 95\% correlation level, respectively, between the classes and their natural context in the training set. This artificially introduces strong spurious correlations between classes and contexts, 
makes the model to fail in scenarios where the test set does not follow the same biased distribution seen at training.

To tackle this setting, \cite{YenamandraICCV23FACTS} propose to 
learn  a partitioning Gaussian Mixture model  
on a \biased{} validation set that follows the same bias as the training set and evaluate the partitioning on an \unbiased{}   set 
where all class-context are equally represented
(see  statistics in \cref{tab:NPPstat}).  Note that this is done for each class independently (per-class case).
Using the code provided by \cite{YenamandraICCV23FACTS}\footnote{\url{https://github.com/yvsriram/FACTS}}, we learn and evaluate 
all possible set combinations (see \cref{tab:NPP-mAP}). Note that
(\biased-\biased) and (\unbiased-\unbiased) sets means that we learnt and test the partitioning function on the exact same data.

In our case, there is no learning involved: we simply 
split and cluster the data as discussed in~\cref{sec:method}. Then,  for a given ($A,B$) pair of sets,  where $A,B \in \{\biased,\unbiased\}$, we assign images from 
the set $B$ to the cluster prototypes obtained with the images in set $A$, where the union of easy and hard prototypes is considered.
We use the class probability to split the data---the same measure
used by FACTS~\cite{YenamandraICCV23FACTS} to learn the partitioning function, except that we only use the probability of the target class while they use the probabilities of all classes.

\myparagraph{Evaluation protocol} Before discussing the comparative results, we recall the metrics and the evaluation protocol we use, proposed by \cite{EyubogluICLR22DominoDiscoveringSystematicErrors,YenamandraICCV23FACTS}. In particular they propose to rely on the  Precision-at-K (\patk) to
measure  how accurately a partitioning method 
aggregates samples from the same context given a class. 
Formally, let $\calX$ denote the test set, $\{\bfz_n\}_{n=1}^{N}$ the ground-truth partitions and $\{\widehat{\bfz}_m\}_{m=1}^{M}$ the predicted set of discovered partitions. Given $K$, for each 
 ground-truth partition $\bfz_n$ the  most 
"similar"  predicted partition $\widehat{\bfz}_m$ is selected based on the size of the intersection  $\bfz_n \cap \widehat{\bfz}^K_m$, 
where $\widehat{\bfz}^K_m$ is the subset containing the top-$K$ elements of $\calX$ according to the likelihood of belonging to $\widehat{\bfz}_m$. For FACTS/DOMINO the ranking corresponds to the partitioning function's output, while in our case to the assignment to the closest prototype. If we denote by $\widehat{\bfz}^K_{nm}$ the set that has the largest overlap with $\bfz_n$ given $K$, the \patk is defined as 
\begin{equation*} 
P\at K= \frac{1}{N}  \sum_{n=1}^N   \frac{\lvert  \bfz_n \cap \widehat{\bfz}^K_{nm}\rvert}{K} 
\end{equation*} 
where  the average is computed, as in \cite{YenamandraICCV23FACTS},
over the five bias-conflicting groups (excluding the spuriously correlated GT partitions, such as  the context \textit{water} for \textit{waterways}).

\input{main_NPP_results_mAP_all}

\myparagraph{Experimental results} We show in \cref{tab:NPP-mAP} the \patk values averaged over the six classes  for the different \npp sets.
Each time we show results averaged over three different variants of the current
model trained with different seed\footnote{Since we were not able to reproduce the original paper's
numbers with the code, namely, the results corresponding to the (\biased,\unbiased) pairs, 
for fairness we ran their code with three different seeds and provide here the averaged results. We do not show the  variances in the table to keep it simple, but the values are most often between 1 and 2.}. In our case, we used the exact same ResNet50 models and the same \biased{} and \unbiased{} sets on  which we perform  the split and the clustering. We use five easy and five hard clusters as it is our default setting for the per-class case and as in the code  from \cite{YenamandraICCV23FACTS} the default number of slices was also set to ten.

From  \cref{tab:NPP-mAP} we can see that our split-and-cluster
method provides competitive, or sometimes even better, slice discovery performance
than the more complex partition learning methods. 
Our method performs slightly worse than FACTS on the \unbiased{} set both when their partitioning  was trained on the \unbiased{} or 
on the \biased{} set. Surprisingly though, both FACTS and DOMINO underperform
when tested on the 
\biased{} set even when the partitioning function was trained on this set. In contrast, our method is more robust and performs similarly on both sets. 
When the  FACTS model is trained on the \biased{} set, which strongly relies on priors about class-context associations (the configuration shown in their paper), it outperforms our model, for which the P\at K decreases as the amount of class-context correlation increases (bottom part of \cref{tab:NPP-mAP}).
We hypothesize that the reason behind this result is that
we do not have enough data for a proper clustering  (see \cref{tab:NPPstat}), while FACTS manages to learn the partitioning function even with a a few samples by exploiting the strong correlation between the classes and dominant context. 
As the model is trained mainly with 
images with specific combinations of class and correlated context---for example, mammals in rocky environments---the model rarely sees images from other classes in this context. Consequently, when the model learns to predict a class "$c$" with high probability, it also associates this high probability with the presence of the corresponding context (\eg rocks) even if the class (\textit{mammals}) is
absent. Therefore, when analyzing another class, such as \textit{plants}, and learn the partitioning based on all six probabilities, the partitioning function  primarily  learns to  assigns images based on the highest probability score, which is an indicator of the context.  For instance, a plant image taken in a rocky environment will have a high probability score for the mammal class, indicating the presence of rocks. Note that in our case, when analyzing the plant class, we disregard the probabilities of other classes, thereby using much less prior information about the image context.

In summary, despite the simplicity of our method, it performs favorably to prior art in  their proposed settings, while also being able to provide natural language descriptions of error modes as shown in \cref{sec:res}.

\section{Defining the sentence sets}
\label{sec:sentences}

Our paper focuses on two objectives, 1)
identifying error modes in computer vision models and 2)
describing them with natural language.
To achieve this, we rely on a large set of sentences ($\calS$), that 
can either be provided or automatically generated. This sentence set is expected to cover various potential reasons for model failures given a task. In this section, we first discuss 
three approaches to construct such sentence sets (\cref{sec:construcS}); then 
we detail how we create the ground-truth sentence set $\calS^{*}_{\beta}$ for \task{} (\cref{sec:GTSentenceSet}); finally, we show how to obtain pseudo-GT relevance scores  between images and sentences using Visual-Question Answering  (VQA) tools when not available otherwise (\cref{sec:pseudoGT}).

\subsection{Building the sentence set $\calS$}
\label{sec:construcS}

There are several possibilities to construct the sentence sets $\calS$. We describe a few possibilities below.

\myparagraph{User-defined} The  set of sentences can be designed ad hoc for the task at hand by the end user who has the expert knowledge and can envisage the main difficulties a model might face. While such sentence set might be non-exhaustive, it can  cover the most important
failure causes. Our method aim to select from them the ones that confirm the user's concerns (with additional visual support). In this case, to get pseudo-ground-truth relevance  between images and a sentence, we resort  to VQA (as described in  \cref{sec:pseudoGT}).

We follow this setting
to build $\calS$ in our experiments on the urban scene segmentation task.
More specifically, we manually defined
148 sentences describing content related to urban scenes (buildings, pedestrians, traffic, two-wheels, trees, animals, garbage, \etc) in the form of  \{``\textit{An image of a} \texttt{<urban scene content>}''\},  road  conditions (weather, lighting, season) in form of   \{``\textit{An image taken in/at} \texttt{<condition>}''\} or  image quality (underexposure, motion blur, \etc) in form of 
\{``\textit{An image with} \texttt{<effect>}''\}. A large subset of this sentence set can be seen in \cref{tab:TPFP_UrbanSegm}.

\myparagraph{GT-based} If available, we can use ground-truth information (class labels and metadata) to generate a sentence set. This is quite limiting in general: we mainly use this option because  it yields the advantage of providing the
GT relevance between images and sentences.
We used this option in all \npp experiments, where we build  130 sentences that include combinations from  the six classes, the six contexts, and various sub-class information. For example,  we have  \{``\textit{A photo of}  \texttt{<sub-class>} \textit{in the} \texttt{<context>}''\} such as 
``\textit{A photo of a cactus in the water}''.

We also built a GT-based sentence set for the WD2 dataset~\cite{ZendelECCV18WildDashCreatingHazardAwareBenchmarks} tested in the context of urban scene segmentation task. It contains annotations for 81 classes, from which we selected 74 classes (removing the ambiguous ones most difficult for VLMs to handle, namely \textit{`unlabeled'}, \textit{`ego-vehicle'}, \textit{`overlay'}, \textit{`out-of-roi'}, \textit{`static'}, \textit{`dynamic'}, \textit{`curb terrain'}).  The sentences have the form \{``\textit{An image showing a}  \texttt{<class>}''\}, where \texttt{<class>} was replaced each time by one of the 
74 classes selected.  While this allows us to generate GT relevance scores for
each image in WD2, this set limits our methods in selecting only from failure causes  related to the presence of these classes in the image and hence they do not cover 
all the possible failure reasons that may explain the
model's behavior.

In the case of  ImageNet-1K~\cite{RussakovskyIJCV15Imagenet},  the first 1K
sentences (out of 1417) were also generated from the GT information, namely the class names. We consider  sentences with the form ``\{An image of a \texttt{<class>}''\}. 
This was complemented by a set of user-defined sentence set related to various places 
in  the form \{``\textit{An image taken in} \texttt{<location>}''\} where we filled the \texttt{<location>} with the  place class names from Places365~\cite{ZhouPAMI18Places365}. Finally
we also added a set of sentences describing image type (photo, drawing), image quality (blurry, noise, JPEG compression), weather conditions, \etc  yielding to a total of 
1417 sentence set used for ImageNet-1K.

\myparagraph{LLM-generated} In the case of urban scene segmentation task, we also consider a set of sentences generated by a large language model (LLM). We provided GPT-3.5 with the context and the sentence structure and prompted it to provide lists of sentences related to weather conditions, 
vehicles, urban objects, time of day and road conditions.  For each of them we used either the sentence form 
\{``\textit{An image taken in/at} <\texttt{weather/road condition} or \texttt{time of day>}''\} or
\{``\textit{An image  showing a/an} <\texttt{vehicle}/\texttt{urban object>}''\},  where 
 the template is filled by the LLM.
This yields to a set of 1016 sentences,  including most of the ones we design manually in the  User-defined set. On the other end, the limitation of this approach is the noise in the set, which can also include sentences hardly helpful for the task at end, \eg
 ``\textit{An image taken on an unpredictable/eclectic/dystopian road}'' or ``\textit{An image taken in a corn maze/trick-or-treat/pajama weather}''.

\subsection{Defining the ground-truth  $\calS_\beta^{*}$  for \task{}}
\label{sec:GTSentenceSet}

To evaluate the predicted sentence set   $\calR_\calS  \subset \calS$ output by an \task{}  approach,
we must define a set of ground-truth sentences  $\calS^{*}  \subset \calS$ to compare  our predictions to---namely, ground-truth error mode descriptions.
To derive such GT set, we consider the function $\omega_{\theta}: \mathbb{R}^{H\times W\times 3} \rightarrow \mathbb{R}$
that measures the model's performance on each image $\bfx_i$. 
This function can be the accuracy in the case of a classification task, the  mIoU (mean Intersection over Union) score for semantic segmentation, or any other task-specific metric.
Given the set of images $\calX$,  the average value of this measure over the entire dataset
\begin{equation}
\omega^{\text{avg}}_{\theta} = \frac{1}{\lvert \calX \rvert} \sum_{\bfx_m\in \calX} \omega_{\theta}(\bfx_m)
    \enspace,
\end{equation}
is the overall performance of $\calM_{\theta}$ on $\calX$.

Given this value and a threshold $\beta$, 
we define the ground-truth sentence set $\calS^{*}$, 
associated with  failure cases of $\calM_{\theta}$
on $\calX$, as the set of sentences for which the mean performance across images associated with the sentence 
 falls beneath the overall mean performance by a given margin $\beta$.
To retrieve the images associated with a sentence, we define 
a binary  function $\Gamma (\bfx_m,s_n)$ that outputs 1 if the sentence $s_n$ describes the image $\bfx_m$ (\ie, is relevant to it), and 0 otherwise. For example, if $\bfx_m$ is an image taken at night, $\Gamma (\bfx_m,``\textit{Image taken at night}'')=1$ whereas $\Gamma (\bfx_m,``\textit{Image taken at sunset}'')=0$. If this relevance cannot be derived directly from ground truth information (most of the cases),
we resort to VQA~\cite{AntolICCV15VQAVisualQuestionAnswering} to obtain such information for arbitrary sentences on arbitrary datasets---as detailed  in \cref{sec:pseudoGT}. 

The set of images  associated with a sentence  $s_n$  is defined as 
\begin{equation}
    \calX_{s_n} = \{\bfx_m \in \calX | \Gamma (\bfx_m,s_n)=1 \} 
    \enspace.
\end{equation}
Then, for each sentence we compute a score by averaging the model performance over the associated images
\begin{equation}\label{eq:hardness}
\omega_{\theta}^{s_n} =\frac{1}{\lvert \calX_{s_n} \rvert} \sum_{\bfx_m \in \calX_{s_n} } \omega_\theta(\bfx_m)   \enspace .
\end{equation}
We use this quantity to represent the sentence's \textit{hardness} score (according to the measure $\omega_{\theta}$)
and we consider a sentence as \textit{describing a failure mode} if its value is below 
the global average $\omega^{\text{avg}}_{\theta}$ by at least a margin $\beta$. 
Therefore, the desired output $\calS^{*}$ is the collection of all of these sentences:
\begin{equation}\label{eq:prob-form-2}
\calS^{*} = \{s_n \in \calS | \,  \omega_{\theta}^{s_n} <  \omega_{\theta}^\text{avg} - \beta  \} \enspace . 
\end{equation}
Varying $\beta$ allows to evaluate the methods with different 
% set of 
severity levels
for the desired error descriptions, \ie, a higher $\beta$ restricts the desired set $\calS^{*}$ to more challenging sentences (see \cref{fig:BetaStudy}).

\subsection{Pseudo-GT image-sentence relevance}
\label{sec:pseudoGT}

When we do not have access to GT information,
we rely on  VQA~\cite{AntolICCV15VQAVisualQuestionAnswering}
to determine whether  
an image $x_m \in \calX$ can be associated with  a sentence $s_n$ (description) or not. These 
binary relevance scores $\Gamma (x_m,s_n)$ are necessary to define 
the GT sentence set $\calS^{*}$  for \task{} (see \cref{sec:GTSentenceSet}) but also to derive 
the sentence coverage ratio in the cluster (CR defined in \cref{sec:LBEquality}).

In particular, we use
OFA~\cite{WangICML22OFAUnifyingArchitecturesTasksModalities}
and LLaVA~\cite{LiuNIPS23VisualInstructionTuning}  with their publicly available pre-trained weights---but any other (possibly more recent) 
model can be used instead. 
We do not simply use OFA and LLaVA alone, but we also combine them:
combining the CLIP-based  LLaVA with non CLIP-based OFA representation allow us to significantly decrease the CLIP bias from the pseudo-GT relevance scores. 

To proceed with VQA, we first turn each sentence $s_n \in \calS$
into a question $q_n$ such that the expected answer is 
\textit{yes} or \textit{no}. For example 
the sentence ``\textit{An image taken at night}'' was turned into the prompt: ``\textit{Was the 
image taken at night? Reply simply with `yes' or `no'}. ''.
Then we provide  VQA model with the image-question pairs $(x_i,q_n)$  and turn the yes/no
answer into a binary score that is assigned to $\Gamma (x_i, s_n)$. 

\input{vqa_gt_eval.tex}

\myparagraph{Comparing VQA pseudo-GT to manually annotated GT}
Given the key role that VQA 
plays in our experimental validation,
we carry out an experimental analysis to compare
the pseudo-GT image-sentence associations obtained with  
LLaVA~\cite{LiuNIPS23VisualInstructionTuning} and/or OFA~\cite{WangICML22OFAUnifyingArchitecturesTasksModalities} to manually annotated GT.  To this end, we rely on
the GT annotations from WD2~\cite{ZendelECCV18WildDashCreatingHazardAwareBenchmarks} and the derived 74 
sentences (described in \cref{sec:construcS}) from which we compute the
GT scores for $\Gamma (x_i, s_n)$. 
Turning these sentences into questions, we  can gather the pseudo-GT relevance scores 
 $\Gamma_{\text{OFA}} (x_i, s_n)$ and  $\Gamma_{\text{LLAVA}} (x_i, s_n)$.

We evaluate the accuracy of these pseudo-GT associations  independently, as well as combined  with AND/OR logical operations (\textbf{OFA \& LLaVA} and \textbf{OFA | LLaVA}, respectively) and report our results in~\cref{tab:vqa-gt-eval}.
Combining the output from both models yields the highest accuracy with respect to the ground-truth. Moreover, the 
accuracy of both models achieves an accuracy above  70\% of the GT performance, with \textbf{LLAVA} hallucinating more (FP) and \textbf{OFA} missing more associations (FN).  In light of this study, we use \textbf{OFA \& LLaVA} to create pseudo-GT sentence-image relevance scores whenever GT information is not available. Note that the overhead 
of this computation does not affect the methods themselves, as it is used  for evaluation only.

\section{Design choices}
\label{sec:param_select}

In this section we recall and detail our default parameter setting.
We use Open-CLIP~\cite{openclip21} trained on LAION-2B~\cite{SchuhmannNIPS22LAION5BAnOpenLargeScaleDataset}
as our default visual-textual embedding space, but one could use other VLM models, such as BLIP~\cite{LiICML22BLIPBootstrappingLanguageImagePreTraining},
\etc To generate the GT sentence set $\calS^{*}_{\beta}$, 
we use $\beta = o *\omega_{\theta}^{\text{std}}$, where $\omega_{\theta}^{\text{std}}$ is the standard deviation of $\omega_{\theta}$ computed over $\calX$ with $o=0.2$ as default value. 
We set the number of hard and easy clusters to $C=15$ for large datasets (when we 
analyze the full set, namely urban scene segmentation and the unsupervised analyses on the \npp datasets) and to
$C=5$ for small  sets, namely when we analyze the dataset for each class independently (per-class analyses of \npp and  ImageNet). We set the number of retained sentences per cluster for all methods to $K=3$.

\begin{figure*}[ttt]
{\footnotesize
    \centering
    {{\includegraphics[width=.85\linewidth]{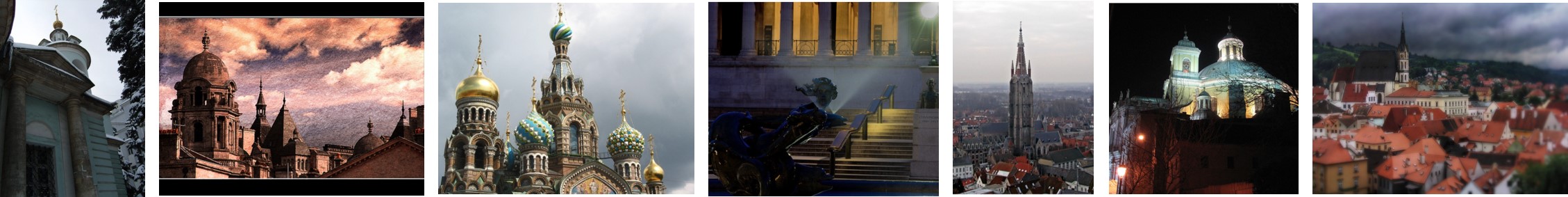} }}\\
     \begin{tabular}{cccc}
    \tops & \setdiff  & \pdiff  & \fpdiff \\
 ``\textit{dome}''  \,\, \,\, & \,\, \,\,  ``\textit{taken in a dim weather}''  \,\, \,\, & \,\, \,\, ``\textit{taken in a snowy weather}''  \,\, \,\, & \,\, \,\, ``\textit{taken in a dim weather}''\\
``\textit{church}''  \,\, \,\, & \,\, \,\, ``\textit{taken at dusk}''  \,\, \,\, & \,\, \,\, ``\textit{taken in roof garden}''    \,\, \,\, & \,\, \,\, ``\textit{taken in a foggy weather}''\\
``\textit{taken in tower}''    \,\, \,\, & \,\, \,\, ``\textit{taken at night}''  \,\, & \,\,``\textit{taken in a dim weather}'' \,\, & \,\,``\textit{taken in a rainy weather}''
    \end{tabular} \\
      {{\includegraphics[width=.85\linewidth]{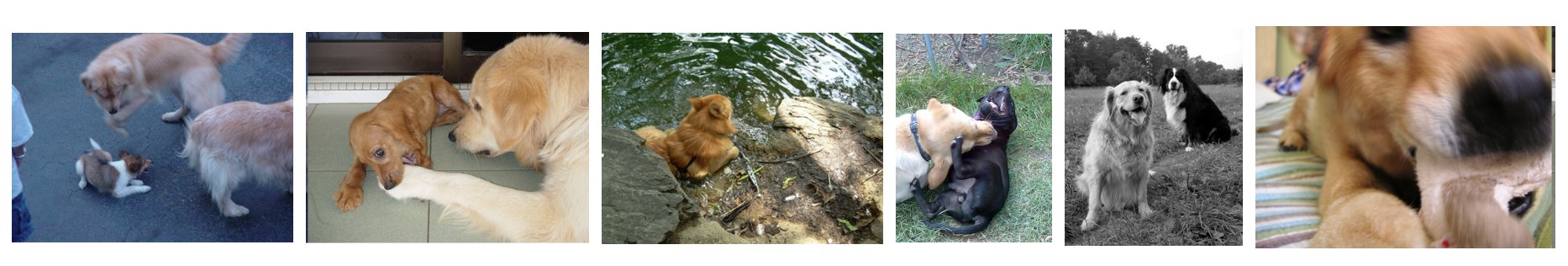} }}\\
       \begin{tabular}{cccc}
    \tops & \setdiff  & \pdiff  & \fpdiff \\
 ``\textit{taken in pet shop}''  \,\, \,\, & \,\, \,\,  ``\textit{taken in pet shop}''  \,\, \,\, & \,\, \,\, ``\textit{Shetland sheepdog}''  \,\, \,\, & \,\, \,\, ``\textit{taken in pet shop}''\\
``\textit{taken in kennel outdoor}''  \,\, \,\, & \,\, \,\, ``\textit{veterinarians office}''  \,\, \,\, & \,\, \,\, ``\textit{taken in pet shop}''    \,\, \,\, & \,\, \,\, ``\textit{veterinarians office}''\\
``\textit{golden retriever}''    \,\, \,\, & \,\, \,\, ``\textit{a thumbnail image}''  \,\, & \,\,``\textit{taken in hospital room}'' \,\, & \,\,``\textit{a chow}''
    \end{tabular} \\
     {{\includegraphics[width=.85\linewidth]{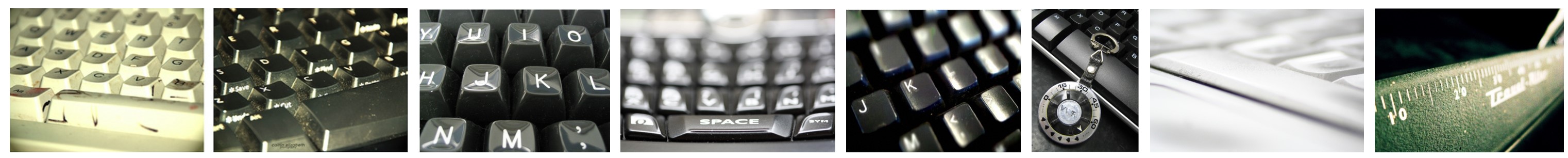} }}\\
     \begin{tabular}{cccc}
    \tops & \setdiff  & \pdiff  & \fpdiff \\
 ``\textit{typewriter keyboard}''  \,\, \,\, & \,\, \,\,  ``\textit{with low contrast}''  \,\, \,\, & \,\, \,\, ``\textit{taken in clean room}''  \,\, \,\, & \,\, \,\, ``\textit{rule}''\\
``\textit{computer keyboard}''  \,\, \,\, & \,\, \,\, ``\textit{rule}''  \,\, \,\, & \,\, \,\, ``\textit{with studio lighting}''    \,\, \,\, & \,\, \,\, ``\textit{hardware store}''\\
``\textit{space bar}''    \,\, \,\, & \,\, \,\, ``\textit{taken in conference room}''  \,\, & \,\,``\textit{taken in entrance hall}'' \,\, & \,\,``\textit{with low contrast}''
    \end{tabular} \\
 {{\includegraphics[width=.85\linewidth]{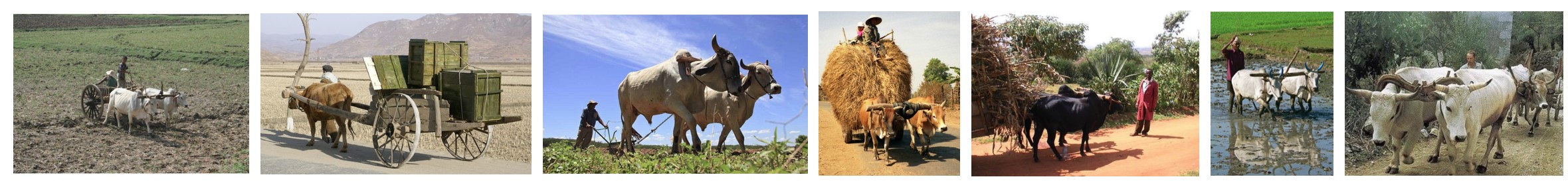} }}\\
     \begin{tabular}{cccc}
    \tops & \setdiff  & \pdiff  & \fpdiff \\
 ``\textit{oxcart}''  \,\, \,\, & \,\, \,\,  ``\textit{taken in medina}''  \,\, \,\, & \,\, \,\, ``\textit{oxcar}''  \,\, \,\, & \,\, \,\, ``\textit{oxcar}''\\
``\textit{taken in a village}''  \,\, \,\, & \,\, \,\, ``\textit{taken in barndoor}''  \,\, \,\, & \,\, \,\, ``\textit{taken in slum}''    \,\, \,\, & \,\, \,\, ``\textit{taken in medina}''\\
``\textit{taken in farm}''    \,\, \,\, & \,\, \,\, ``\textit{taken in hayfield}''   \,\,\,\, & \,\, \,\, ``\textit{taken in medina}''  \,\, \,\, & \,\, \,\, ``\textit{taken in a rice paddy}''
    \end{tabular} \\
     {{\includegraphics[width=.85\linewidth]{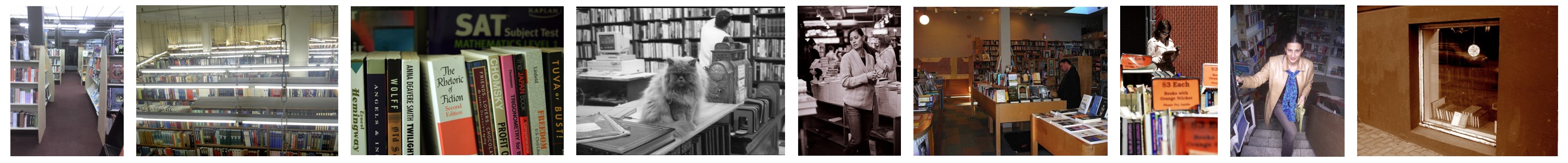} }}\\
        \begin{tabular}{cccc}
    \tops & \setdiff  & \pdiff  & \fpdiff \\
 ``\textit{taken in bookstore}''  \,\, \,\, & \,\, \,\,  ``\textit{taken in archive}''  \,\, \,\, & \,\, \,\, ``\textit{taken in slum}''  \,\, \,\, & \,\, \,\, ``\textit{taken in archive}''\\
``\textit{taken in library indoor}''  \,\, \,\, & \,\, \,\, ``\textit{taken in office}''  \,\, \,\, & \,\, \,\, ``\textit{taken in street}''    \,\, \,\, & \,\, \,\, ``\textit{taken in office}''\\
``\textit{bookshop}''    \,\, \,\, & \,\, \,\, ``\textit{taken in shed}''  \,\, & \,\,``\textit{taken in porch}'' \,\, & \,\,``\textit{taken in ticket booth}''
    \end{tabular} \\
    }
    \caption{
   Hard clusters from ImageNet 1K validation set tested with the pre-trained Vit-B16 model. Here we show examples  containing  mainly false negative (images from the current class not recognized).  From top to bottom we
   have the classes ``\textit{church}'', ``\textit{golden retriever}'', ``\textit{computer keyboard}'', ``\textit{ox}'' and ``\textit{bookshop}''.}    
\label{fig:ImageNetFN}
\end{figure*}

In the unsupervised  cases, we split $\calX$ according to $t_{\varphi}^h=\varphi_{\theta}^{\text{avg}} +  a*\varphi_{\theta}^{\text{std}}$  and $t_{\varphi}^e = \varphi_{\theta}^{\text{avg}} -  a*\varphi_{\theta}^{\text{std}}$,
where $\varphi_{\theta}$ is the model's output entropy and we set $a=0.2$ as default value. 
In the per-class analysis using \npp datasets, we use class probabilities $\rho_{\theta}$ and split according to  
$t_{\rho}^h=\rho_{\theta}^{\text{avg}} -  a*\rho_{\theta}^{\text{std}}$  and $t_{\rho}^e = \rho_{\theta}^{\text{avg}} -  a*\rho_{\theta}^{\text{std}}$. 
For ImageNet, the splitting is done according to the ranking, with the hard set including both false positives (the GT class not ranked first)
and false negatives (an image not labeled with the current class where the model ranked the current class as first)---as described in \cref{sec:datasets}.

\section{Further qualitative analyses}
\label{sec:qualitative}
\begin{figure*}[ttt]%
{\footnotesize
    \centering
    {{\includegraphics[width=.85\linewidth]{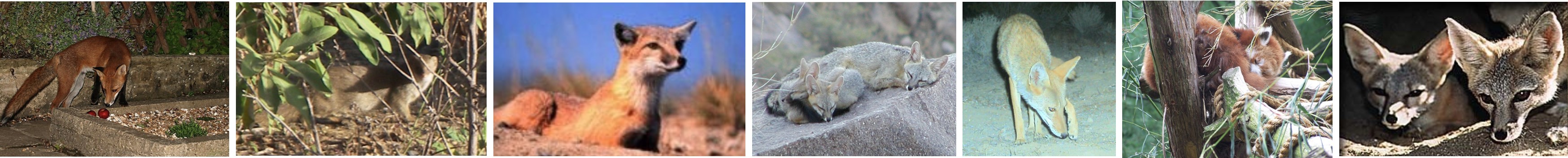} }}\\
     \begin{tabular}{cccc}
    \tops & \setdiff  & \pdiff  & \fpdiff \\
 ``\textit{kit fox}''  \,\, \,\, & \,\, \,\,  ``\textit{mongoose}''  \,\, \,\, & \,\, \,\, ``\textit{mongoose}''  \,\, \,\, & \,\, \,\, ``\textit{mongoose}''\\
``\textit{grey fox}''  \,\, \,\, & \,\, \,\, ``\textit{taken in yard}''  \,\, \,\, & \,\, \,\, ``\textit{a grey fox}''    \,\, \,\, & \,\, \,\, ``\textit{a grey fox
}''\\
``\textit{red fox}''    \,\, \,\, & \,\, \,\, ``\textit{taken in wild field}''  \,\, & \,\,``\textit{desert vegetation}'' \,\, & \,\,``\textit{coyote}''
    \end{tabular} \\
      {{\includegraphics[width=.85\linewidth]{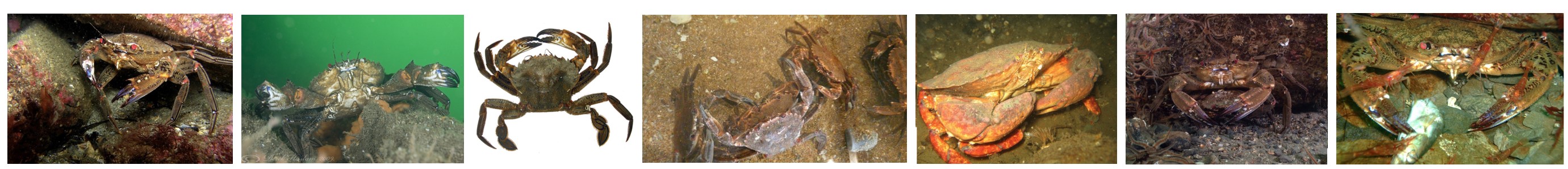} }}\\
       \begin{tabular}{cccc}
    \tops & \setdiff  & \pdiff  & \fpdiff \\
 ``\textit{rock crab}''  \,\, \,\, & \,\, \,\,  ``\textit{thumbnail image}''  \,\, \,\, & \,\, \,\, ``\textit{SD generated image}''  \,\, \,\, & \,\, \,\, ``\textit{taken in fishpond}''\\
``\textit{Dungeness crab}''  \,\, \,\, & \,\, \,\, ``\textit{damaged image}''  \,\, \,\, & \,\, \,\, ``\textit{taken in galley}''    \,\, \,\, & \,\, \,\, ``\textit{Dungeness crab}''\\
``\textit{fiddler crab}''    \,\, \,\, & \,\, \,\, ``\textit{taken in alcove
}''  \,\, & \,\,``\textit{taken at night}'' \,\, & \,\,``\textit{taken in lagoon}''
    \end{tabular} \\
     {{\includegraphics[width=.85\linewidth]{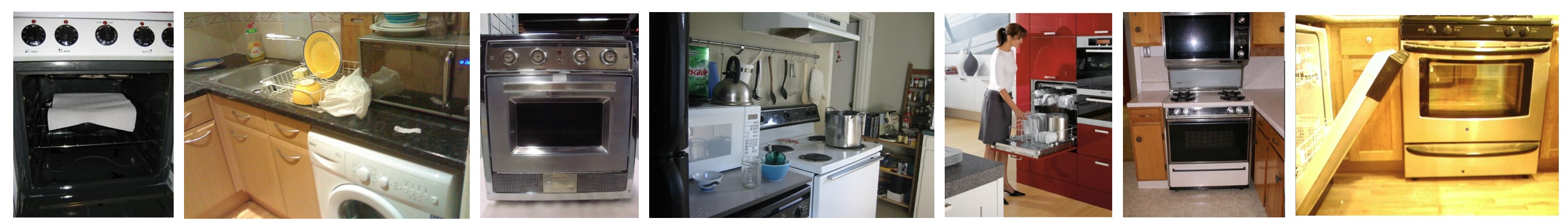} }}\\
     \begin{tabular}{cccc}
    \tops & \setdiff  & \pdiff  & \fpdiff \\
 ``\textit{stove}''  \,\, \,\, & \,\, \,\,  ``\textit{taken in archive}''  \,\, \,\, & \,\, \,\, ``\textit{stove}''  \,\, \,\, & \,\, \,\, ``\textit{stove}''\\
``\textit{taken in kitchen}''  \,\, \,\, & \,\, \,\, ``\textit{thumbnail image
}''  \,\, \,\, & \,\, \,\, ``\textit{dishwasher}''    \,\, \,\, & \,\, \,\, ``\textit{dishwasher}''\\
``\textit{taken in galley}''    \,\, \,\, & \,\, \,\, ``\textit{taken in laundromat}''  \,\, & \,\,``\textit{SD generated image}'' \,\, & \,\,``\textit{taken in kitchen}''
    \end{tabular} \\
 {{\includegraphics[width=.85\linewidth]{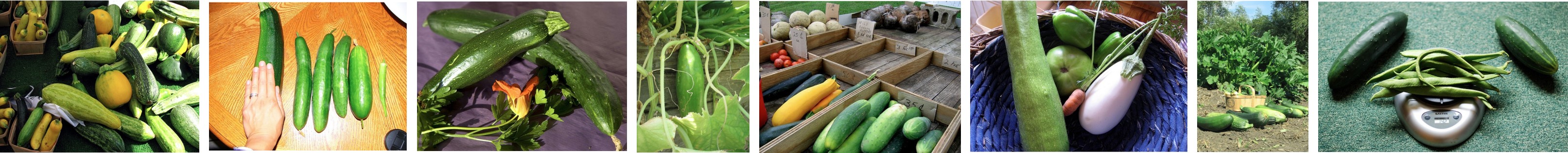} }}\\
     \begin{tabular}{cccc}
    \tops & \setdiff  & \pdiff  & \fpdiff \\
 ``\textit{zucchini}''  \,\, \,\, & \,\, \,\,  ``\textit{taken in a farm}''  \,\, \,\, & \,\, \,\, ``\textit{vegetable garden}''  \,\, \,\, & \,\, \,\, ``\textit{vegetable garden}''\\
``\textit{cucumber}''  \,\, \,\, & \,\, \,\, ``\textit{taken in a market outdoor}''  \,\, \,\, & \,\, \,\, ``\textit{taken in a farm}''    \,\, \,\, & \,\, \,\, ``\textit{taken in a farm}''\\
``\textit{vegetable garden}''    \,\, \,\, & \,\, \,\, ``\textit{taken in yard}''  \,\, & \,\,``\textit{taken in market outdoor}'' \,\, & \,\,``\textit{taken in market outdoor}''
    \end{tabular} \\
     {{\includegraphics[width=.85\linewidth]{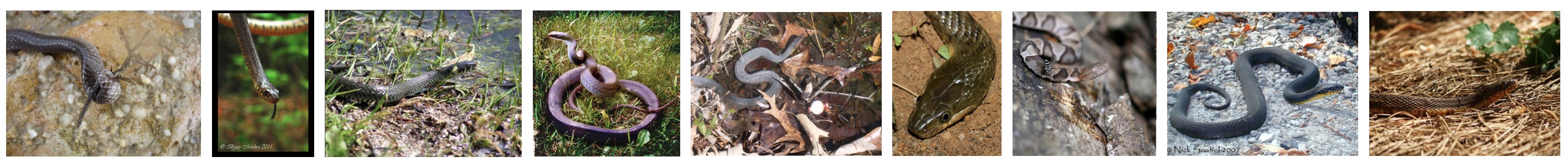} }}\\
        \begin{tabular}{cccc}
    \tops & \setdiff  & \pdiff  & \fpdiff \\
 ``\textit{water snake}''  \,\, \,\, & \,\, \,\,  ``\textit{green mamba}''  \,\, \,\, & \,\, \,\, ``\textit{thumbnail image}''  \,\, \,\, & \,\, \,\, ``\textit{thumbnail image}''\\
``\textit{ringneck snake}''  \,\, \,\, & \,\, \,\, ``\textit{thumbnail image}''  \,\, \,\, & \,\, \,\, ``\textit{taken in forest broadleaf}''    \,\, \,\, & \,\, \,\, ``\textit{ringneck snake}''\\
``\textit{vine snake}''    \,\, \,\, & \,\, \,\, ``\textit{taken in arch}''  \,\, & \,\,``\textit{taken in a cloudy weather}'' \,\, & \,\,``\textit{green mamba}''
    \end{tabular} \\
    }
    \caption{
   Hard clusters from ImageNet 1K validation set tested with the pre-trained Vit-B16 model. Here we show examples  containing  mainly false positive (images from other classes wrongly labeled with the current class name).  From top to bottom we
   have the classes ``\textit{red fox}'', ``\textit{rock crab}'', ``\textit{microwave}'', ``\textit{cucumber}'' and ``\textit{water snake}''.}    
\label{fig:ImageNetFP}
\end{figure*}

\begin{figure*}[ttt]
{\footnotesize
    \centering
    {{\includegraphics[width=.85\linewidth]{NPP85_cl8.jpg} }}\\
     \begin{tabular}{cccc}
    \tops & \setdiff    & \pdiff  & \fpdiff\\
 ``\textit{taken in autumn}''  \,\, \,\, & \,\, \,\, ``\textit{taken in autumn}''  \,\, \,\, & \,\, \,\, ``\textit{sailboat in the water}'' 
\,\, & \,\,  ``\textit{sailboat in the water}'' \\
``\textit{with waterways}''  \,\, & \,\,``\textit{with waterways}''   \,\, \,\, & \,\, \,\, ``\textit{sailboat in the rocks}''  \,\, \,\, & \,\, \,\, 
 ``\textit{sailboat}''\\
``\textit{plants in the water}''    \,\, & \,\,``\textit{plants in the water}'' \,\, & \,\,``\textit{goose in the water}'' \,\, & \,\,``\textit{birds in the water}''
    \end{tabular} \\
        {{\includegraphics[width=.85\linewidth]{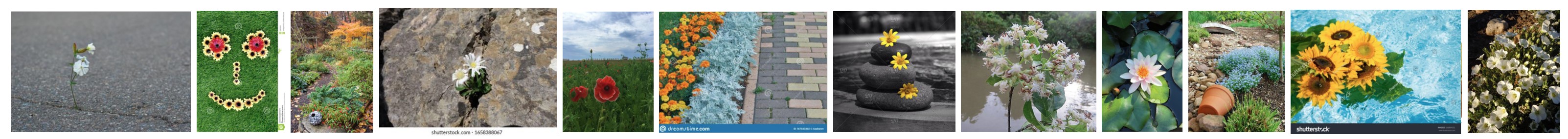} }}\\
     \begin{tabular}{cccc}
    \tops \,\, & \,\,\setdiff    \,\, & \,\,\pdiff  \,\, & \,\,\fpdiff\\
 ``\textit{flowers}''  \,\, \,\, & \,\, \,\, ``\textit{plants}''  \,\, \,\, & \,\, \,\, ``\textit{flowers in the rocks}'' \,\, \,\, & \,\, \,\, ``\textit{flowers in the rocks}''\\
``\textit{plants}''  \,\, & \,\,``\textit{flowers in the rocks}''   \,\, \,\, & \,\, \,\, ``\textit{plants in the rocks}''  \,\, \,\, & \,\, \,\, ``\textit{plants in the rocks}''\\
``\textit{flowers in the rocks}''    \,\, & \,\,``\textit{flowers in the water
}'' \,\, & \,\,``\textit{monkey in the water}'' \,\, & \,\,``\textit{flowers in the water}''
    \end{tabular} \\
            {{\includegraphics[width=.85\linewidth]{NPP85_cl4.jpg} }} \\
     \begin{tabular}{cccc}
    \tops \,\, & \,\,\setdiff   \,\, & \,\,\pdiff   \,\, & \,\,\fpdiff \\
 ``\textit{train}''  \,\, \,\, & \,\, \,\, ``\textit{train}''  \,\, \,\, & \,\, \,\, ``\textit{ train in the rocks}'' \,\, \,\, & \,\, \,\, ``\textit{ train in the rocks}''\\
``\textit{train in the rocks}''  \,\, & \,\,``\textit{train in the rocks}''   \,\, \,\, & \,\, \,\, ``\textit{bus in the rocks}'' \,\, \,\, & \,\, \,\, ``\textit{bus in the rocks}''\\
``\textit{bus in the rocks}''    \,\, & \,\,``\textit{bus in the rocks}'' \,\, & \,\,``\textit{train in the grass}'' \,\, & \,\,``\textit{train in the grass}''
    \end{tabular} \\
    {{\includegraphics[width=.85\linewidth]{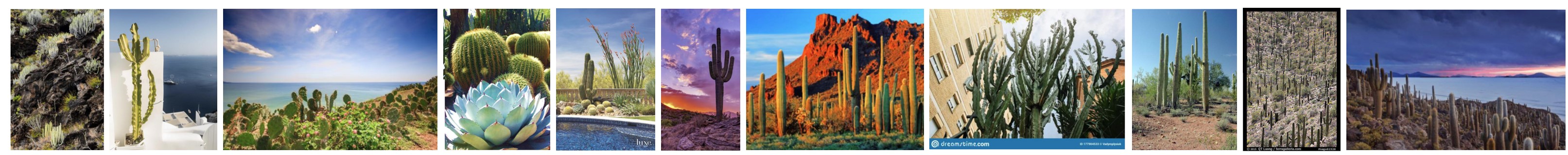} }} \\
    \begin{tabular}{cccc}
    \tops \,\, & \,\,\setdiff  \,\, & \,\,\pdiff    \,\, \,\, & \,\, \,\, \fpdiff \\
 ``\textit{cactus}''  \,\, \,\, & \,\, \,\, ``\textit{cactus in the rocks}''  \,\, \,\, & \,\, \,\, ``\textit{cactus in the rocks}'' \,\, \,\, & \,\, \,\, ``\textit{cactus in the rocks}'' \\
``\textit{cactus in the rocks}''  \,\, & \,\,``\textit{cactus in the water}''   \,\, \,\, & \,\, \,\, ``\textit{cactus in the water}'' \,\, \,\, & \,\, \,\, ``\textit{cactus in the water}''\\
``\textit{cactus in the grass}''    \,\, \,\, & \,\, \,\, ``\textit{plants in the rocks}'' \,\, & \,\,``\textit{cactus}'' \,\, & \,\,``\textit{cactus}''\\
    \end{tabular} \\
{{\includegraphics[width=.85\linewidth]{NPP85_cl14.jpg} }}\\
  \begin{tabular}{cccc}
    \tops \,\, & \,\,\setdiff   \,\, & \,\,\pdiff   \,\, & \,\,\fpdiff \\
 ``\textit{truck taken at sunset}''  \,\, \,\, & \,\, \,\, ``\textit{truck taken at sunset}''  \,\, \,\, & \,\, \,\, ``\textit{truck taken at sunset}''  \,\, \,\, & \,\, \,\, ``\textit{truck taken at sunset}''\\
``\textit{bus taken at sunset}''  \,\, & \,\,``\textit{bus taken at sunset}''   \,\, \,\, & \,\, \,\, ``\textit{bus taken at sunset}'' \,\, \,\, & \,\, \,\, ``\textit{bus taken at sunset}''\\
``\textit{train taken at sunset}''    \,\, & \,\,``\textit{train taken at sunset}'' \,\, & \,\,``\textit{truck in the water}''  \,\, & \,\,``\textit{truck in the water}''
    \end{tabular} \\
    {{\includegraphics[width=.85\linewidth]{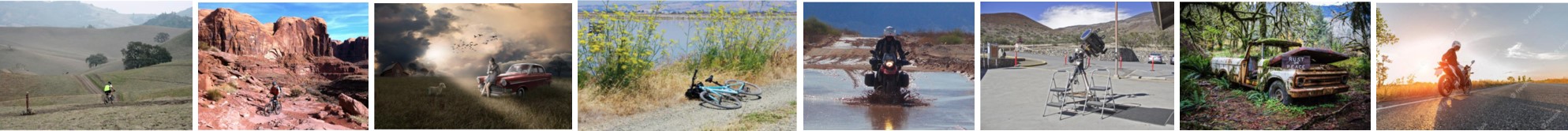} }}\\
  \begin{tabular}{cccc}
    \tops \,\, & \,\,\setdiff   \,\, & \,\,\pdiff   \,\, & \,\,\fpdiff \\
 ``\textit{motorcycle in the rocks}''  \,\, \,\, & \,\, \,\, ``\textit{motorcycle in the rocks}''  \,\, \,\, & \,\, \,\, ``\textit{motorcycle in the rocks}''  \,\, \,\, & \,\, \,\, ``\textit{motorcycle in the rocks}''\\
``\textit{motorcycle}''  \,\, & \,\,``\textit{motorcycle}''   \,\, \,\, & \,\, \,\, ``\textit{bicycle in the rocks}'' \,\, \,\, & \,\, \,\, ``\textit{bicycle in the rocks}''\\
``\textit{landways}''    \,\, & \,\,``\textit{landways}'' \,\, & \,\,``\textit{car in the rocks}''  \,\, & \,\,``\textit{car in the rocks}''
    \end{tabular} \\
        {{\includegraphics[width=.85\linewidth]{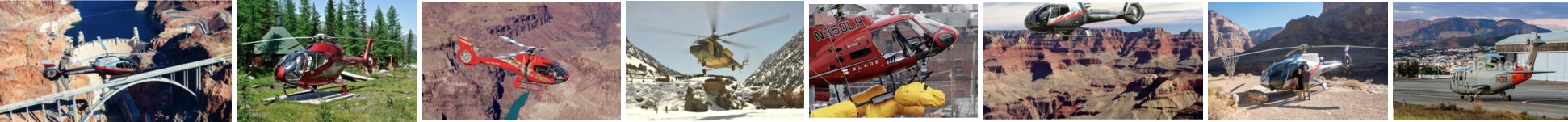} }}\\
  \begin{tabular}{cccc}
    \tops \,\, & \,\,\setdiff   \,\, & \,\,\pdiff   \,\, & \,\,\fpdiff \\
 ``\textit{helicopter in the rocks}''  \,\, \,\, & \,\, \,\, ``\textit{hot air balloon in the rocks}''  \,\, \,\, & \,\, \,\, ``\textit{hot air balloon in the rocks}''  \,\, \,\, & \,\, \,\, ``\textit{hot air balloon in the rocks}''\\
``\textit{helicopter}''  \,\, & \,\,``\textit{airways}''   \,\, \,\, & \,\, \,\, ``\textit{bicycle in the rocks}'' \,\, \,\, & \,\, \,\, ``\textit{helicopter in the rocks}''\\
``\textit{helicopter taken at sunset}''    \,\, & \,\,``\textit{airplane in the water}'' \,\, & \,\,``\textit{ train in the rocks}''  \,\, & \,\,``\textit{airplane in the rocks}''
    \end{tabular} \\
    }
    \caption{
   Seven out of 15 hard clusters from \nppm  (unsupervised case)   explained by different methods.
    }    
\label{fig:qualitativeNPP}
\end{figure*}

\begin{figure*}[ttt]%
{\footnotesize
    \centering
    {{\includegraphics[width=.8\linewidth]{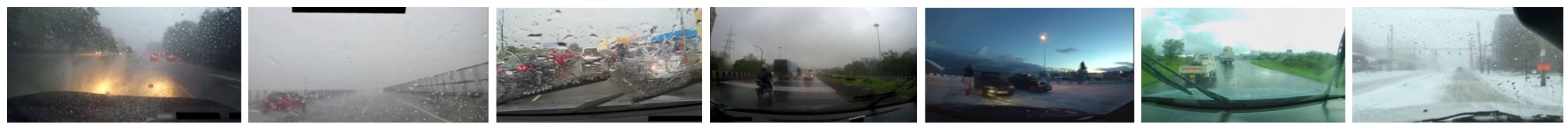} }}\\
     \begin{tabular}{cccc}
    \tops & \setdiff    & \pdiff  & \fpdiff\\
 ``\textit{rainy weather}''  \,\,\,\, & \,\,\,\, ``\textit{rainy weather}''  \,\,\,\, & \,\,\,\, ``\textit{rainy weather}'' 
\,\,\,\, & \,\,\,\,  ``\textit{rainy weather}'' \\
``\textit{vehicle on the road}''  \,\,\,\, & \,\,\,\,``\textit{stormy weather}''   \,\,\,\, & \,\,\,\, ``\textit{stormy weather}''  \,\,\,\, & \,\,\,\, 
 ``\textit{stormy weather}''\\
``\textit{dull weather}''    \,\,\,\, & \,\,\,\,``\textit{dust on the road}'' \,\,\,\, & \,\,\,\,``\textit{dull weather}'' \,\,\,\, & \,\,\,\,``\textit{dull weather}''
    \end{tabular} \\
        {{\includegraphics[width=.8\linewidth]{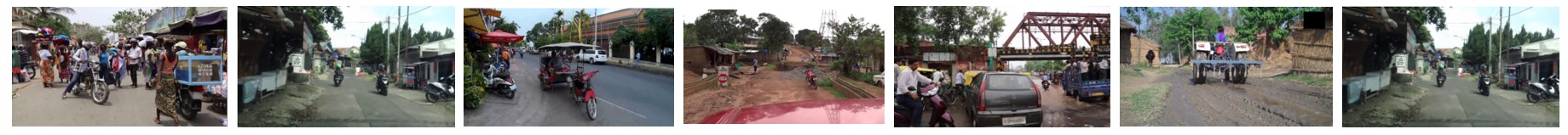} }}\\
     \begin{tabular}{cccc}
    \tops \,\,\,\, & \,\,\,\,\setdiff    \,\,\,\, & \,\,\,\,\pdiff  \,\,\,\, & \,\,\,\,\fpdiff\\
 ``\textit{rickshaw on the road}''  \,\,\,\, & \,\,\,\, ``\textit{sub-urban scene}''  \,\,\,\, & \,\,\,\, ``\textit{market}'' \,\,\,\, & \,\,\,\, ``\textit{rickshaw on the road}''\\
``\textit{vehicle on the road}''  \,\,\,\, & \,\,\,\,``\textit{dull weather}''   \,\,\,\, & \,\,\,\, ``\textit{restaurants}''  \,\,\,\, & \,\,\,\, ``\textit{taken in the morning}''\\
``\textit{obstacle on the road}''    \,\,\,\, & \,\,\,\,``\textit{low contrast}'' \,\,\,\, & \,\,\,\,``\textit{with defocus blur}'' \,\,\,\, & \,\,\,\,``\textit{taken in the afternoon}''
    \end{tabular} \\
            {{\includegraphics[width=.8\linewidth]{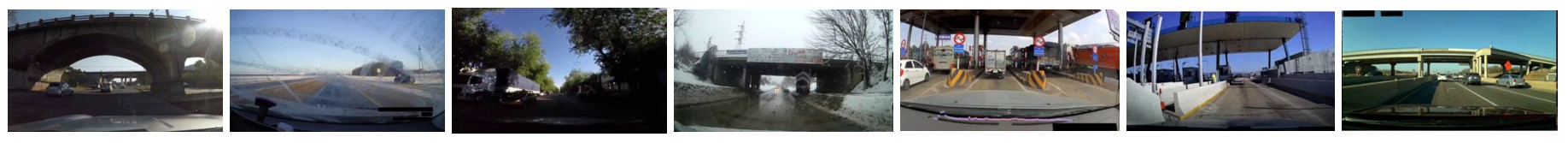} }} \\
     \begin{tabular}{cccc}
    \tops \,\,\,\, & \,\,\,\,\setdiff   \,\,\,\, & \,\,\,\,\pdiff   \,\,\,\, & \,\,\,\,\fpdiff \\
 ``\textit{vehicle on the road}''  \,\,\,\, & \,\,\,\, ``\textit{tunnel}''  \,\,\,\, & \,\,\,\, ``\textit{tunnel}'' \,\,\,\, & \,\,\,\, ``\textit{shadows on the road}''\\
``\textit{car on the road}''  \,\,\,\, & \,\,\,\,``\textit{guard-rail}''   \,\,\,\, & \,\,\,\, ``\textit{guard-rail}'' \,\,\,\, & \,\,\,\, ``\textit{light reflection on the road}''\\
``\textit{motion blur}''    \,\,\,\, & \,\,\,\,``\textit{dirt on the road}'' \,\,\,\, & \,\,\,\,``\textit{snowy weather}'' \,\,\,\, & \,\,\,\,``\textit{motion blur}''
    \end{tabular} \\
   {{\includegraphics[width=.8\linewidth]{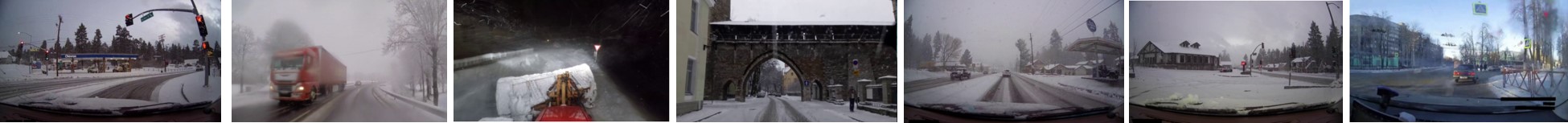} }} \\
    \begin{tabular}{cccc}
    \tops \,\,\,\, & \,\,\,\,\setdiff  \,\,\,\, & \,\,\,\,\pdiff    \,\,\,\, & \,\,\,\, \fpdiff \\
 ``\textit{vehicle on the road}''  \,\,\,\, & \,\,\,\, ``\textit{snowy weather}''  \,\,\,\, & \,\,\,\, ``\textit{snowy weather}'' \,\,\,\, & \,\,\,\, ``\textit{snowy weather}'' \\
``\textit{clear weather}''  \,\,\,\, & \,\,\,\,``\textit{rainy weather}''   \,\,\,\, & \,\,\,\, ``\textit{rainy weather}'' \,\,\,\, & \,\,\,\, ``\textit{rainy weather}''\\
``\textit{snowy weather}''    \,\,\,\, & \,\,\,\, ``\textit{stormy weather}'' \,\,\,\, & \,\,\,\,``\textit{stormy weather}'' \,\,\,\, & \,\,\,\,``\textit{taken in winter}''\\
    \end{tabular} \\
    {{\includegraphics[width=.8\linewidth]{WD2_cl7.jpg} }} \\
  \begin{tabular}{cccc}
    \tops  & \setdiff    & \pdiff    & \fpdiff \\
``\textit{vehicle on the road}''    \,\,\,\, & \,\,\,\,  ``\textit{mud on the road}''   \,\,\,\, & \,\,\,\,  ``\textit{mountain
}''    \,\,\,\, & \,\,\,\, ``\textit{mud on the road}''\\
``\textit{obstacle on the road}''    \,\,\,\, & \,\,\,\,  ``\textit{taken in a windy weather}''     \,\,\,\, & \,\,\,\,  ``\textit{terrain}''   \,\,\,\, & \,\,\,\,  ``\textit{rocks on the road}''\\
``\textit{jeep on the road}''     \,\,\,\, & \,\,\,\,  ``\textit{round-about scene}''   \,\,\,\, & \,\,\,\, ``\textit{forest
}''    \,\,\,\, & \,\,\,\,  ``\textit{jeep on the road}''
    \end{tabular} \\
 {{\includegraphics[width=.8\linewidth]{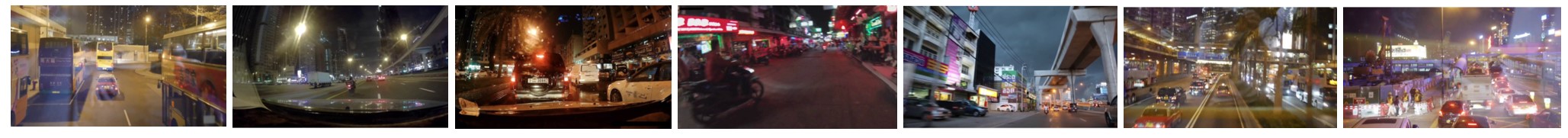} }} \\
  \begin{tabular}{cccc}
    \tops \,\,\,\, & \,\,\,\,\setdiff   \,\,\,\, & \,\,\,\,\pdiff   \,\,\,\, & \,\,\,\,\fpdiff \\
 ``\textit{traffic}''  \,\,\,\, & \,\,\,\, ``\textit{taken at night}''  \,\,\,\, & \,\,\,\, ``\textit{taken at night}''  \,\,\,\, & \,\,\,\, ``\textit{taken at night}''\\
``\textit{showing a sub-urban scene}''  \,\,\,\, & \,\,\,\,``\textit{taken in the evening}''   \,\,\,\, & \,\,\,\, ``\textit{taken at dusk}'' \,\,\,\, & \,\,\,\, ``\textit{taken in the evening}''\\
``\textit{vehicle on the road}''    \,\,\,\, & \,\,\,\,``\textit{with underexposure}'' \,\,\,\, & \,\,\,\,``\textit{taken in the evening}''  \,\,\,\, & \,\,\,\,``\textit{taken in a rainy weather}''
    \end{tabular} \\
{{\includegraphics[width=.8\linewidth]{WD2_cl11.jpg} }} \\
  \begin{tabular}{cccc}    \tops  & \setdiff    & \pdiff    & \fpdiff \\ ``\textit{tunnel}''   \,\,\,\, & \,\,\,\,  ``\textit{tunnel}''   \,\,\,\, & \,\,\,\,  ``\textit{tunnel}''   \,\,\,\, & \,\,\,\,  ``\textit{tunnel}''\\ ``\textit{vehicle on the road}''   \,\,\,\, & \,\,\,\, ``\textit{taken at night}''  \,\,\,\, & \,\,\,\,  ``\textit{taken at night}'' \,\,\,\, & \,\,\,\,  ``\textit{light reflections on the road}''\\ ``\textit{car on the road}''   \,\,\,\, & \,\,\,\, ``\textit{with high brightness}''  \,\,\,\, & \,\,\,\,  ``\textit{light reflections on the road}''   \,\,\,\, & \,\,\,\,  ``\textit{with motion blur}''     \end{tabular} \\
    {{\includegraphics[width=.8\linewidth]{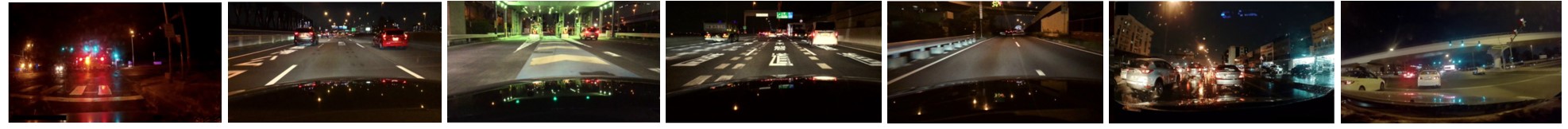} }} \\
  \begin{tabular}{cccc}
    \tops  & \setdiff    & \pdiff    & \fpdiff \\
 ``\textit{light reflections on the road}''   \,\,\,\, & \,\,\,\,  ``\textit{taken at night}''   \,\,\,\, & \,\,\,\, ``\textit{taken at night}''   \,\,\,\, & \,\,\,\,  ``\textit{taken at night}''\\
``\textit{taken at night}''   \,\,\,\, & \,\,\,\, ``\textit{taken in the evening}''    \,\,\,\, & \,\,\,\,  ``\textit{taken at dusk}''  \,\,\,\, & \,\,\,\,  ``\textit{taken at dusk}''\\
``\textit{vehicle on the road}''    \,\,\,\, & \,\,\,\,  ``\textit{taken at dusk}''  \,\,\,\, & \,\,\,\, ``\textit{light reflections on the road}''  \,\,\,\, & \,\,\,\, ``\textit{light reflections on the road}''
    \end{tabular} \\
{{\includegraphics[width=.8\linewidth]{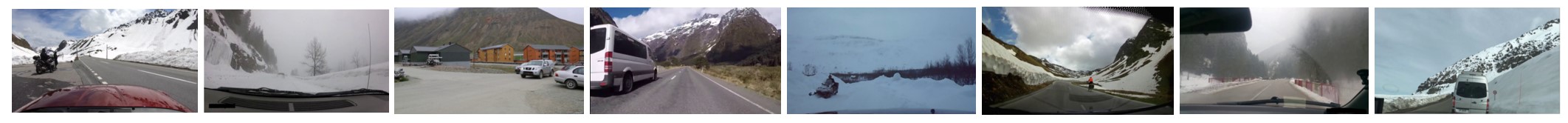} }} \\
  \begin{tabular}{cccc}
    \tops \,\,\,\, & \,\,\,\,\setdiff   \,\,\,\, & \,\,\,\,\pdiff   \,\,\,\, & \,\,\,\,\fpdiff \\
 ``\textit{vehicle on the road}''  \,\,\,\, & \,\,\,\, ``\textit{taken in winter}''  \,\,\,\, & \,\,\,\, ``\textit{snowy weather}''  \,\,\,\, & \,\,\,\, ``\textit{taken in winter}''\\
``\textit{clear weather}''  \,\,\,\, & \,\,\,\,``\textit{snowy weather}''   \,\,\,\, & \,\,\,\, ``\textit{taken in winter}'' \,\,\,\, & \,\,\,\, ``\textit{dull weather}''\\
``\textit{taken at daytime}''    \,\,\,\, & \,\,\,\,``\textit{tunnel}'' \,\,\,\, & \,\,\,\,``\textit{mountain}''  \,\,\,\, & \,\,\,\,``\textit{snowy weather}''
    \end{tabular} \\
    }
    \caption{
   Nine out of 15 hard clusters from WD2  explained by different methods.
    }    
\label{fig:qualitativeWD2}
\end{figure*}

First, we  provide in \cref{tab:TPFP_UrbanSegm} an extended version of 
\cref{tab:TPFP_UrbanSegmSmall} 
We show qualitative examples  
for ImageNet 1K using the  pre-trained Vit-B16 model in~\cref{fig:ImageNetFN,fig:ImageNetFP} , for \nppm in \cref{fig:qualitativeNPP},  
and finally further examples obtained for the urban scene segmentation model on   WD2 in \cref{fig:qualitativeWD2}.
We conclude the section   with \cref{tab:TPFP_UrbanSegm}
which is an extended version of \cref{tab:TPFP_UrbanSegmSmall} 
 in the main paper.

%\newpage
\input{TPFPUrbanSegm}

%% file: npp_biased_stats.tex
\begin{table*}[ttt]
    \centering
    \caption{The  number of images in the \biased{} validation set for each (class, context) pair in corresponding three \npp datasets. The \unbiased{} validation sets have 50 images for each (class, context) pairs in each of the three sets. The spurious contexts (1 to 6)  are in order  \textit{rock, grass, dim
lighting, autumn, water, outdoor}. 
    }  \label{tab:NPPstat}
       \resizebox{\linewidth}{!}{
    \begin{tabular}{|l||c||c||c|}
    \toprule
    \begin{tabular}{c}
    \npp set \\
     \hline
class $\downarrow$  context $\rightarrow$\\
\hline
mammals  \\
birds \\
plants  \\
airways \\
waterways \\
landways \\
    \end{tabular} & 

    \begin{tabular}{cccccc}
    \multicolumn{6}{c}\npph \\
    \midrule
  1& 2& 3& 4& 5& 6\\
  \hline
      638	& 4 &  2 & 	2 & 	3	& 3 \\
 7	& 321 & 	2	& 2 & 	3 & 	3 \\
 7	& 4 & 	154 & 	2 & 	3 & 	3 \\
 7	& 4 & 	2 & 	220 & 	3 & 	3 \\
 7	& 4 & 	2 & 	2 & 	266 & 	3 \\
 7	& 4	 & 2 & 	2 & 	3 & 	277 \\
\end{tabular} & 

\begin{tabular}{cccccc}
 \multicolumn{6}{c}\nppm \\
    \midrule
   1& 2& 3& 4& 5& 6\\
  
  \hline
     638	& 4 &  2 & 	2 & 	3	& 3 \\
 7	& 321 & 	2	& 2 & 	3 & 	3 \\
 7	& 4 & 	154 & 	2 & 	3 & 	3 \\
 7	& 4 & 	2 & 	220 & 	3 & 	3 \\
 7	& 4 & 	2 & 	2 & 	266 & 	3 \\
 7	& 4	 & 2 & 	2 & 	3 & 	277 \\
    \end{tabular} & 
\begin{tabular}{cccccc}
 \multicolumn{6}{c}\nppl \\
   \midrule
    1& 2& 3& 4& 5& 6\\
  \hline
      638	& 10 &  5 & 	7 & 	8	& 9 \\
 19	& 321 & 	5	& 7 & 	8 & 	9 \\
 19	& 10 & 	154 & 	7 & 	8 & 	9 \\
 19	& 10 & 	5 & 	220 & 	8 & 	9 \\
19	& 10 & 	5 & 	7 & 	266 & 	9 \\
19	& 10	 & 5 & 	7 & 	8 & 	277 \\
    \end{tabular} \\
      \bottomrule
\end{tabular}
}
    
\end{table*}

%% file: main_NPP_results_mAP_all.tex
\setlength{\tabcolsep}{3pt}
\renewcommand{\arraystretch}{1.0}
\begin{table*}[t!]
%\scriptsize
    \centering
    \caption{\textbf{Comparisons with prior art.} We compare 
    the clustering obtained via Step 3 of our proposed 
    methods  (see  \cref{fig:method})
    with the Domino and FACTS using averaged \patk scores with $K=10$.  For our method, we split the dataset into 5 ``easy'' and  5 ``hard'' clusters and use their union as final partitioning. On the top rows (below Random) we show results obtained with the partitioning 
    learnt on the \unbiased{} set and on the bottom  rows the results 
    were obtained when  the partitioning is learnt on the \biased{} set. Random corresponds to  assigning the test data randomly to  the partitions. 
    } \label{tab:NPP-mAP}
     %\resizebox{0.95\linewidth}{!}{
    \begin{tabular}{ll|cc|cc|cc}
    \toprule
    \textbf{Method} &
    %\multirow{2}{*}{\textbf{Method}}
    % \multirow{2}{*}{(Test set \rightarrow})
    & \multicolumn{2}{c|} \nppl
    & \multicolumn{2}{c|} \nppm
    & \multicolumn{2}{c} \npph \\
 \,\,\,\,\,\,\, Test set &  $\rightarrow$
     & \textbf{Unbiased} & \textbf{Biased} &
    \textbf{Unbiased} & \textbf{Biased} &
    \textbf{Unbiased} & \textbf{Biased} \\
\midrule
Random &   & $29.7$  & $12.6$  & $32.0$ & $14.0$ & $33.0$ & $14.6 $ \\
\midrule
Domino~\cite{EyubogluICLR22DominoDiscoveringSystematicErrors} & \multirow{3 }{*}{\rotatebox[origin=c]{90}{\small \textbf{Unbiased}}} & $44.4 $ & $12.1 $ & $47.0 $ & $12.3$ & $52.7 $ & $22.3$ \\
FACTS~\cite{YenamandraICCV23FACTS} &   & $\textbf{58.8}$ & $28.1 $ & $\textbf{63.6}$ & $30.0$ & $\textbf{68.9} $ & $39.9 $ \\ 
% Ours(1-5) &   & $46.2 $ & $42.9$ & $46.4 $ & $43.0 $ & $48.5$ & $53.9$ \\
% Ours(3-3) &   & $44.6 $ & $41.8$ & $43.7 $ & $46.8 $ & $43.0$ & $50.5$ \\
 \textbf{Ours}(5-5) &   & $53.6 $ & $\textbf{52.3}$ & $54.2 $ & $\textbf{58.4} $ & $55.3$ & $\textbf{66.1}$ \\ 
\midrule
Domino~\cite{EyubogluICLR22DominoDiscoveringSystematicErrors} & \multirow{3 }{*}{\rotatebox[origin=c]{90}{\small \textbf{Biased}}}  & $26.0$ & $17.1$ & $19.0 $ & $19.8$ & $17.4 $ & $25.6 $ \\
FACTS~\cite{YenamandraICCV23FACTS} &   & $\textbf{53.3}$ & $31.4$ & $\textbf{53.2}$ & $28.7$ & $\textbf{58.9}$ & $34.4$ \\
% Ours(1-5) &   & $38.6 $ & $37.2 $ & $35.3 $ & $35.8 $ & $27.3 $ & $31.3$ \\
% Ours(3-3) &   & $41.6 $ & $38.9 $ & $41.4 $ & $42.4 $ & $35.8 $ & $42.3$ \\
\textbf{Ours}(5-5) &   & $51.3 $ & $\textbf{52.2} $ & $46.8 $ & $\textbf{52.5} $ & $41.9 $ & $\textbf{46.5}$ \\
\bottomrule
 \end{tabular}
% }
 %\vspace{-0.3cm}
\end{table*}

%% file: vqa_gt_eval.tex
\begin{table}[t!]
\setlength{\tabcolsep}{3pt}
\centering
\caption{Comparison of pseudo-ground truth annotations created using LLaVA and OFA with manual ground truth annotations in WD2.}
\resizebox{0.95\linewidth}{!}{
\begin{tabular}{l|c|c|c|c|c}
\toprule
\textbf{Model} & \textbf{Accuracy$\uparrow$} & \textbf{TP$\uparrow$} & \textbf{TN$\uparrow$} & \textbf{FP$\downarrow$} & \textbf{FN$\downarrow$} \\
\midrule
OFA & 0.799 & 0.489 & 0.852 & 0.148 & 0.511 \\
LLaVA & 0.731 & 0.623 & 0.696 & 0.304 & 0.377 \\
OFA | LLaVA & 0.701 & \textbf{0.735} & \textbf{0.642} & 0.358 & \textbf{0.265} \\
OFA \& LLaVA & \textbf{0.829} & 0.377 & 0.905 & \textbf{0.095} & 0.623 \\
\bottomrule
\end{tabular}
}
\label{tab:vqa-gt-eval}
\end{table}

%% file: TPFPUrbanSegm.tex
\begin{table*}[hhh]
    \footnotesize
    \centering
    \caption{
          Example sentences from the User-defined sentence set $\calS$. For each sentence we show if it belongs to $\calS^*_{\beta}$ (obtained with $o=.2$) and to $\calR_\calS$ for \tops (\textbf{TS}), \pdiff (\textbf{PD}), \setdiff (\textbf{SD}) and  \fpdiff (\textbf{FP}) for the  datasets WD2, IDD and ACDC. For  $\calS^*_{\beta}$, ``\gcheckmark'' means $s_n \in \calS^*_{\beta}$ and  ``\redx'' means $s_n \notin \calS^*_{\beta}$---namely, for that dataset the hardness score of $s_n$  is below the required level.  For the methods, ``\gplus''/``\rplus'' indicate that the sentence is in $\calR_\calS$, where ``\gplus''  means  true positive (\ie, the sentence is also in $\calS^*_{\beta}$) while  ``\rplus''  means  false positive. An empty space indicates that the sentence is not in the corresponding $\calR_\calS$. Note that the table is not complete as we omitted to show a few sentences that were not in any $\calS^*_{\beta}$, semantically very similar to a  sentences in the table with similar behaviour or retained by a single method for one dataset, but gives a very good overview of the set based results.
    }
    \begin{tabular}{l|ccccc|ccccc|ccccc}
        \toprule
         & \multicolumn{5}{c}{WD2} &  \multicolumn{5}{c}{IDD} 
          & \multicolumn{5}{c}{ACDC}  \\
          \textbf{Sentences} (\textit{An image ...)} & $\calS^*$ & \textbf{TS} & \textbf{PD} & \textbf{SD} & \textbf{FP} & 
          $\calS^*$ & \textbf{TS} & \textbf{PD} & \textbf{SD} & \textbf{FP}  &
          $\calS^*$ & \textbf{TS} & \textbf{PD} & \textbf{SD} & \textbf{FP} \\
        \midrule
         ``taken at night'' &  \gcheckmark  &  \gplus & \gplus & \gplus  & \gplus & \gcheckmark & & \gplus & \gplus & \gplus &  \gcheckmark &  \gplus & & & \\
        ``taken in the evening'' &  \gcheckmark   &   & \gplus & \gplus  & \gplus &  \gcheckmark   &   &   \gplus &  & \gplus &  \gcheckmark &  \gplus & & &  \\
        ``taken at dusk'' & \gcheckmark   &  & \gplus & \gplus & \gplus &
        \gcheckmark   &  & \gplus & \gplus & \gplus & \gcheckmark & & & &   \\
         ``taken at dawn'' & \redx &  &  \rplus & & &  \gcheckmark & & \gplus & \gplus  &    & \redx & & & & \\
       
        ``taken in a stormy weather'' &  \gcheckmark  &  & \gplus & \gplus & \gplus & \redx & & & & & \redx & & & & \rplus\\
        ``taken in a foggy weather''  & \gcheckmark &  &   & \gplus  &   & \gcheckmark & & & & &  \redx & \rplus  & \rplus &  \rplus & \rplus
        \\
        ``taken in a rainy weather'' &   \gcheckmark & \gplus & \gplus &   & \gplus & \gcheckmark & & & & &  \redx & \rplus  & \rplus &   & \rplus\\
            ``taken in a snowy weather''  & \gcheckmark &  & \gplus & \gplus & \gplus & \redx & & & & & \redx &  \rplus  & & & \\
             ``taken in a windy weather'' &   \gcheckmark  & \gplus & \gplus & \gplus & \gplus & 
        \redx & & & &  & \redx & &  \rplus & &  \rplus\\
             ``taken in a dull weather''  & \gcheckmark  & \gplus &  \gplus & \gplus & \gplus & \redx & & & & &  \redx & \rplus  & \rplus &  \rplus & \rplus\\
        ``taken in winter'' &   \gcheckmark & & & \gplus & \gplus  & \redx & & & & &  \redx & & & & \rplus\\
         ``taken in autumn'' & \redx &  & \rplus &  &  & \gcheckmark & & & & &  \redx & & & & \\
      ``with reflection on the road'' &   \gcheckmark & \gplus &  \gplus & \gplus & \gplus &   \gcheckmark & & &   \gplus & & \gcheckmark & & & & \gplus \\
        ``with shadows on the road'' & \redx &  &  &  & \rplus   & \redx & & & \rplus &  & \gcheckmark & & & \gplus & \\
         ``with water on the road'' &  \gcheckmark &   \gplus &  & & & \redx & & & & & \redx & & & \rplus & \rplus\\
          ``with mud on the road'' & \redx &  &  & \rplus &  \rplus & \redx & &  \rplus &  \rplus & & \redx & & & & \\
        ``with branches on the road'' &  \gcheckmark &  &  &  &  &  \gcheckmark & & &  \gplus &  \gplus & \redx & & \rplus & \rplus & \rplus\\
        ``with traffic cone on the road'' &  \gcheckmark &  &  &  &  &  \gcheckmark & & & & &  \gcheckmark & &  \gplus   &  \gplus  &  \gplus  \\
        ``with obstacle on the road'' &  \gcheckmark  & \gplus   &  &  & \gplus   & \redx &  \rplus & & & & \redx & & & &\\
    ``with road barrier on the road'' & \redx &  &  & \rplus & & \redx & &  & \rplus &  \rplus  &  \gcheckmark  & \gplus   &  \gplus  & \gplus  & \gplus\\
     ``with rail track on the road'' & \redx &  &   &  &   & \gcheckmark & & &  \gplus &  \gplus & \gcheckmark & \gplus & \gplus & \gplus &\gplus \\
      ``with rocks on the road'' & \redx &  &  & \rplus & \rplus & \redx &  &  & &  & \redx &    & &  & \\
        ``showing a highway scene'' & \redx &  &  & \rplus &  & \redx & & & \rplus & \rplus & \redx & \rplus & \rplus & \rplus & \rplus \\
         ``showing an industrial scene'' &  \gcheckmark &  &  &  &  & \gcheckmark &  &  &  & &  \redx &    & &  &\\
         ``showing construction site'' & \redx & & & & & \gcheckmark  & & \gplus &  \gplus & \gplus &  \redx & &  \rplus &  \rplus\\
        ``showing sub-urban scene'' &  \redx  & \rplus &  & \rplus &  &  \redx & \rplus &    &  & \rplus & \gcheckmark & & &  \\
        ``showing agricultural field''  & \redx & & & & & \gcheckmark & & & & & \redx & & & & \\
         ``showing a round-about'' & \redx & & & & & \redx & & & & & \gcheckmark & & & & \\
        ``with rickshaw on the road'' &  \redx  & \rplus & \rplus &  \rplus  & \rplus & \redx & \rplus & \rplus &  & \rplus  & \gcheckmark & & \gplus & &  \\
        ``with motorbike on the road'' & \redx &  &  \rplus &  & \rplus & \redx & & & &  \rplus & \gcheckmark & & & & \gplus \\
        ``with bike on the road''  & \redx & & & & & \redx & & & \rplus &  \rplus &  \redx & & & \rplus & \\
        ``with vehicle on the road'' & \redx & \rplus &  &  &  & \redx & \rplus &  &  & &  \redx & \rplus &  \rplus  & \rplus & \rplus\\
         ``with bus on the road'' & \redx &  &  &  &  & \redx & \rplus &  \rplus &  & \rplus & \redx & & & \rplus& \\
        ``with tram on the road'' & \redx & & & &  & \redx &  &  & \rplus  &  & \redx & \rplus & \rplus & & \rplus \\
        ``with jeep on the road'' & \redx & \rplus &  & \rplus &  \rplus & \redx & & & & & \redx & & & &\\
         ``with animal on the road'' & \redx &  &   & \rplus &  \rplus  & \gcheckmark & & &  \gplus & \gplus & \redx & & & &\\
          ``with bird  on the road'' & \redx &  &   &  &  & \gcheckmark & & & & & \redx & & & &\\
          ``with people on the road'' & \redx & \rplus  &  & & \rplus & \redx & \rplus  &  & & \rplus  & \redx & & \rplus & &\\
            ``with crowded background'' & \redx & & & & & \gcheckmark & & & \gplus & & \redx & & & &\\
          ``with crowded foreground'' & \redx & & & & & \redx & & \rplus& \rplus & \rplus & \redx & & & &\\
          ``with motion blur'' &  \gcheckmark &  \gplus &  \gplus & & \gplus & \gcheckmark & & & \gplus & \gplus & \gcheckmark & & \gplus & &  \\
        ``with overexposure''  & \gcheckmark &  &  &  &  & \redx & & & & & \redx & & & &\\
         ``with underexposure''  & \redx  &  &  & \rplus & \rplus  & \redx & & & & & \gcheckmark & & & \gplus  &  \\
        ``with low contrast'' & \gcheckmark &  &  &  & & \redx & & & \rplus & \rplus  &  \redx & & & \rplus &\\
          ``of a tunnel'' &  \gcheckmark & \gplus & \gplus & \gplus & \gplus &  \gcheckmark & &  \gplus & & &  \gcheckmark & \gplus & \gplus & \gplus & \gplus \\
          ``of fences'' & \redx & & & &  & \redx & & \rplus  & &  &  \gcheckmark & & \gplus & & \\
           ``of guard-rail'' & \redx & & & &  & \redx & &   & &  &  \redx & & \rplus & \rplus & \rplus \\
         ``of a traffic jam'' &  \gcheckmark &  &  &  &  & \gcheckmark & \gplus & \gplus & & \gplus & \redx & &  & & \\
        ``of a traffic'' & \redx & \rplus &  &  &   & \redx & \rplus & \rplus &  & \rplus & \redx & & \rplus & &\\
         ``of traffic lights'' & \gcheckmark &  &  &  & &  \gcheckmark &  &  & &  & \redx & & \rplus & &\\
        ``of buildings'' & \redx &  & \rplus &  &  &  \redx & & \rplus & & & \redx & & \rplus & & \\
        ``of a parking'' & \redx &  &   \rplus &  &  & \gcheckmark & & \gplus & \gplus & \gplus & \redx & &  \rplus & &\\
        ``of a mountain'' & \redx &  \rplus & \rplus &  &  & \redx & & & & &  \redx & & & &\\
        \bottomrule
    \end{tabular}
    \label{tab:TPFP_UrbanSegm}
\end{table*}